%% file: paper.tex
\documentclass[runningheads]{llncs}
\usepackage{graphicx}
\usepackage{amsmath}
\usepackage{amssymb}
\usepackage{booktabs}

\usepackage{subcaption}
\usepackage{xcolor}
\usepackage{makecell}
\usepackage{comment}
\usepackage{placeins}
\usepackage{longtable}
\usepackage{algorithm}
\usepackage{algpseudocode}
\usepackage{listings}
\usepackage{xspace}

\input{images}

\input{tables}

\newcommand{\spaplus}{CleverLabel\xspace}

\newcommand{\one}{\mathbf{1^*}}
\newcommand{\onem}{$\mathbf{1^*}$\xspace}

\usepackage{hyperref}

\begin{document}
\title{
Label Smarter, Not Harder: CleverLabel for Faster Annotation of Ambiguous Image Classification with Higher Quality}
\titlerunning{Label Smarter, Not Harder: CleverLabel}
\author{Lars Schmarje\inst{1}\and
Vasco Grossmann\inst{1}\and
Tim Michels\inst{1}\and
Jakob Nazarenus\inst{1}\and
Monty Santarossa\inst{1}\and
Claudius Zelenka\inst{1}\and
Reinhard Koch\inst{1}}
\authorrunning{L. Author et al.}
\institute{ Kiel University \email{\{las,vgr,tmi,jna,msa,cze,rk\}@informatik.uni-kiel.de} }
\maketitle              %
\begin{abstract}

High-quality data is crucial for the success of machine learning, but labeling large datasets is often a time-consuming and costly process.
While semi-supervised learning can help mitigate the need for labeled data, label quality remains an open issue due to ambiguity and disagreement among annotators.
Thus, we use proposal-guided annotations as one option which leads to more consistency between annotators.
However, proposing a label increases the probability of the annotators deciding in favor of this specific label.
This introduces a bias which we can simulate and remove.
We propose a new method \spaplus for \textbf{C}ost-effective \textbf{L}ab\textbf{E}ling using \textbf{V}alidated proposal-guid\textbf{E}d annotations and \textbf{R}epaired \textbf{LABEL}s.
\spaplus can  reduce labeling costs by up to 30.0\%, while achieving a relative improvement in Kullback-Leibler divergence of up to 29.8\% compared to the previous state-of-the-art on a multi-domain real-world image classification benchmark.

\spaplus offers a novel solution to the challenge of efficiently labeling large datasets while also improving the label quality.
   
\keywords{Ambiguous \and data-centric \and data annotation}
\end{abstract}

\section{Introduction}
\label{sec:intro}

\picTwo{idea}{idea}{examples}{Illustration of distribution shift --
We are interested in the ground-truth label distribution (blue) which is costly to obtain due to multiple required annotations per image.
Thus, we propose to use proposals as guidance during the annotation to approximate the distribution more cost efficiently (red).
However, this distribution might be shifted toward the proposed class.
We provide with \spaplus (green) a method to improve the biased label distribution (red) to be closer to the original unbiased distribution (blue).
Additionally, we provide with SPA an algorithm to simulate and analyze the distribution shift.
The concrete effects are shown in the right example for the MiceBone dataset on a public benchmark~\cite{schmarje2022benchmark} with the proposal marked by x.
}

Labeled data is the fuel of modern deep learning. 
However, the time-consuming manual labeling process is one of the main limitations of machine learning~\cite{Sheng2008}.
Therefore, current research efforts try to mitigate this issue by using unlabeled data \cite{singh2008unlabeled,berthelot2019mixmatch,fixmatch} or forms of self-supervision \cite{misra2020self,hendrycks2019using,BIT,He2021}.
Following the data-centric paradigm, another approach focuses on improving data quality rather than quantity \cite{data-centric-appraoch,Northcutt2021,Gordon2021}.
This line of research concludes that one single annotation is not enough to capture ambiguous samples \cite{Collins2022,Davani2021BeyondMajority,Basile2021ConsiderDisagree,schmarje2022benchmark}, where different annotators will provide different annotations for the same image.
These cases are common in most real-world datasets~\cite{underwater_uncertainty,mammo-variability,foc,tailception}
and would require multiple annotations per image to accurately estimate its label distribution.
Yet, established benchmarks such as ImageNet or CIFAR \cite{imagenet,cifar} are currently not considering this issue which significantly limits their use in the development of methods that generalize well for ambiguous real-world data.

Acquiring multiple annotations per sample introduces an additional labeling effort, necessitating a trade-off between label quality and quantity.
While semi-supervised learning potentially reduces the amount of labeled data, the issue of label quality still arises for the remaining portion of labeled data  \cite{li2019safe}.
One possible solution for handling ambiguous data is using proposal guided annotations~\cite{papadopoulos2021scaling,desmond2021semi} which have been shown to lead to faster and more consistent annotations \cite{foc,morphocluster}.
However, this approach suffers from two potential issues:
(1) Humans tend towards deciding in favor of the provided proposal~\cite{jachimowicz2019and}.
This \textit{default effect} introduces a bias, since the proposed class will be annotated more often than it would have been without the proposal.
Thus, an average across multiple annotation results in a skewed distribution towards the proposed class as shown in \autoref{fig:idea}.
(2) Real human annotations are required during development which prevents rapid prototyping of proposal systems. %

We provide with \spaplus and SPA two methods to overcome these two issues.
Regarding issue (1), we propose \textbf{C}ost-effective \textbf{L}ab\textbf{E}ling using \textbf{V}alidated proposal-guid\textbf{E}d annotations and \textbf{R}epaired \textbf{LABEL}s (\spaplus) which uses a single class per image as proposal to speed-up the annotation process.
As noted above, this might skew the label distribution towards the proposed class which can be corrected with \spaplus.
We evaluate the data quality improvement achieved by training a network on labels generated by \spaplus by comparing the network's predicted label probability distribution to the ground truth label distribution, which is calculated by averaging labels across multiple annotations as in \cite{schmarje2022benchmark}.
Improved data quality is indicated by a reduction in the difference between the predicted distribution and the ground truth distribution.
In addition, based on a previously published user study \cite{dc3}, we empirically investigate the influence of proposals on the annotator's labeling decisions.
Regarding issue (2), we propose Simulated Proposal Acceptance (SPA), a mathematical model that mimics the human behavior during proposal-based labeling.
We evaluate \spaplus and SPA with respect to their technical feasibility and their benefit when applied to simulated and real-world proposal acceptance data.
Finally, we evaluate these methods on a real-world benchmark and we provide general guidelines on how to annotate ambiguous data based on the gained insights.

Overall, our contributions commit to three different areas of interest:
(1) For improving label quality, we provide the novel method \spaplus and show across multiple simulated and real world datasets a relative improvement of up to 29.8\% with 30.0\% reduced costs in comparison to the state of the art.
(2) For annotating real-world ambiguous data, we provide annotation guidelines based on our analysis, in which cases to use proposals during the annotation.
(3) For researching of countering the effect of proposals on human annotation behavior, we provide our simulation of proposal acceptance (SPA) as an analysis tool.
SPA is motivated by theory and shows similar behavior to human annotators on real-world tasks.
It is important to note that this research allowed us to achieve the previous contributions.
    We provide a theoretical justification for SPA and show that it behaves similarly to human annotators.

\subsection{Related work}

Data and especially high-quality labeled data is important for modern machine learning~\cite{relabelImagenet,Northcutt2021pervasiveErrors}.
Hence, the labeling process is most important in uncertain cases or in ambiguous cases as defined by~\cite{schmarje2022benchmark}.
However, labeling is also not easy in these cases as demonstrated by the difficulties of melanoma skin cancer classification~\cite{naeem2020malignant}. 
The issue of data ambiguity still remains even in large datasets like ImageNet~\cite{imagenet} despite heavy cleaning efforts~\cite{are_we_done,Vasudevan2022}.
The reasons for this issue can arise for example from image artifacts like low resolution~\cite{cifar10h}, inconsistent definitions~\cite{Uijlings2022}, uncertainty of the data or annotators~\cite{merIssue,deep_fish} or subjective interpretations~\cite{Schustek2018instance,Mazeika2022}.

It is important to look at data creation as part of the problem task because it can greatly impact the results.
Recent works have shown that differences can depend on the aggregation of labels between annotators~\cite{Wei2022,Collins2022}, the selection of image data sources on the web~\cite{Nguyen2022}, if soft or hard labels are used as label representation~\cite{Collins2022,Davani2021BeyondMajority,grossmann2022beyond,Basile2021ConsiderDisagree} or the usage of label smoothing~\cite{lukasik2020smoothing,Lukov2022,Muller2019smoothing}.
In this work we concentrate on the labeling step issues only.
Simply applying SSL only partially solves the problem as it tends to overfit~\cite{Arazo2020Confirmation}.
Hence labeling is necessary and the goal should be to label better and more.

A commonly used idea we want to focus on is proposal-based labeling.
It is also known as verification-based labeling~\cite{papadopoulos2021scaling}, label-spreading~\cite{desmond2021semi}, semi-automatic labeling~\cite{lopresti2012optimal}, or suggestion-based annotation~\cite{schulz-etal-2019-analysis}.
\cite{speedlabelaiassist} showed that  proposal-based data labeling increases both accuracy and speed for their user study (n=54) which is in agreement with proof-of-concepts by \cite{foc,dc3}. %
The annotation suggestions for the segmentation and classification in diagnostic reasoning texts had positive effects on label speed and performance without an introduction of a noteworthy bias~\cite{schulz-etal-2019-analysis}.
 We continue this research for the problem of image classification and show that a bias is introduced and how it can be modeled and reversed.

Acceptance or rejection of a proposal was previously modeled e.g. for the review process of scientific publications~\cite{cortes2021inconsistency}.
They applied a  Gaussian process model to simulate the impact of human bias on the acceptance of a paper, but rely on a per annotator knowledge. %
A simulation framework for instance-dependent noisy labels is presented in \cite{gu2022instance,Gao2022SyntheticAnnotators} by using a pseudo-labeling paradigm and  \cite{Jung_Park_Lease_2014} uses latent autoregressive time series model for label quality in crowd sourced labeling. %
Another aspect of labeling are annotation guidelines which can also have an impact on data quality as \cite{shah2018impact} demonstrate for app reviews. 
We do not consider guidelines as biases, instead they are a part of data semantics and use only real annotations per image.%
This has the benefit of avoiding unrealistic synthetic patterns as shown by~\cite{Wei2021cifar10n} and simplifies the required knowledge which makes the process more easily applicable.

Note that active learning~\cite{ren2021survey} is a very different approach, in which the model in the loop decides which data-point is annotated next and the model is incrementally retrained.
It is outside the scope of this article and it might not be suited for a low number of samples with high ambiguity as indicated by \cite{Tifrea2022}.
Consensus processes~\cite{merIssue,swarmv1} where a joined statement is reached manually or with technical support are also out of scope. %

\section{Methods}
\label{sec:methods}

\pic{dataset_bias_3}{
Average annotation probability of a proposed class with the proposal unknown (GT, Unbiased) and known (Biased) to the annotators in four evaluated datasets.
The proposal increases the probability in all observed cases, revealing a clear default effect in the investigated study. 
Its value is shown without any further processing (Biased) and with the contributed correction (\spaplus) which consistently reduces the difference to the unbiased probabilities.
}{0.99}

Previous research on proposal-based systems~\cite{papadopoulos2021scaling,lopresti2012optimal,schulz-etal-2019-analysis} suggests an influence of the default effect bias on the label distribution.
While its impact  is assessed as negligible in some cases, it circumvents the analysis of an unbiased annotation distribution~\cite{jachimowicz2019and} which can be desirable, e.g. in medical diagnostics.
As we can identify a significant bias in our own proposal-based annotation pipeline for several datasets (see Fig.~\ref{fig:dataset_bias_3}), two questions arise: how to mitigate the observed default effect and how it was introduced?

In this section, we provide methods to answer both questions. 
Before we can mitigate the observed default effect, we have to understand how it was introduced.
Thus, we introduce simulated proposal acceptance (SPA) with the goal of reproducing the human behavior for annotating images with the guidance of proposals.
SPA can be used to simulate the labeling process and allow experimental analysis and algorithm development before conducting large scale human annotations with proposals.
Building on this understanding, we  propose CleverLabel which uses two approaches for improving the biased label distribution to mitigate the default effect: 1. a heuristic approach of class distribution blending (CB) 2. a theoretically motivated bias correction (BC).
CleverLabel can be applied to biased distributions generated by humans or to simulated results of SPA.

For a problem with $K \in \mathbb{N}$ classes let $L^x$ and $L_b^x$ be random variables mapping an unbiased or biased annotation of an image $x$ to the selected class $k$.
Their probability distributions $P(L^x = k)$ and $P(L_b^x = k)$ describe the probability that image $x$ is of class $k$ according to a set of unbiased or biased annotations.
As discussed in the literature~\cite{lukasik2020smoothing,Lukov2022,Muller2019smoothing,Davani2021BeyondMajority}, we do not restrict the distribution of $L_x$ further e.g. to only hard labels and instead assume, that we can approximate it via the average of $N$ annotations by $P(L^x = k) \approx \sum_{i=0}^{N-1} \frac{a^x_{i,k}}{N}$ with $a^x_{i,k} \in \{0,1\}$ the i-th annotation for the class $k$ which is one if the class $k$ was selected by the i-th annotator or zero, otherwise.
The default effect can cause a bias, $P(L^x = k) \neq P(L_b^x = k)$ for at least one class $k$.
Especially, for the proposed class $\rho_x$ it can be expected that $P(L^x = \rho_x) < P(L_b^x = \rho_x)$.

\begin{algorithm}[tb]
\caption{Simulated Proposal Acceptance (SPA)}\label{alg:spa}
\begin{algorithmic}
\Require Proposal $\rho_x$; $a'^x_{i} \in \{0\}^{K} $
\State Calculate acceptance probability $A$
\State $r \gets $ random(0,1) 
\If{$r \leq A$} \Comment{Accept proposal}
    \State $a'^x_{i,\rho_x} \gets 1$
\Else \Comment{Sample from remaining classes}
    \State $k \gets $ sampled from $ P(L^x = k ~|~ \rho_x \neq k)$
    \State $a'^x_{i,k} \gets 1$
\EndIf
\end{algorithmic}
\end{algorithm}

\subsection{Simulated Proposal Acceptance}
 \label{subsec:spa}

Given both unbiased as well as biased annotations for the same datasets, we analyze the influence of proposals on an annotator's choice. 
We notice that a main characteristic is that the acceptance probability increases almost linearly with the ground truth probability of the proposal, $P(L^x = \rho_x)$, as shown in the main diagonal in \autoref{fig:micebone_pa}.
If a proposal was rejected, the annotation was mainly influenced by the ground truth probability of the remaining classes.
This observation leads to the following model:
For a given proposal $\rho_x$, we calculate the probability $A$ that it gets accepted by an annotator as
\begin{equation}\label{eq:accept}
A = \delta + (\one - \delta) P(L^x = \rho_x)
\end{equation}
with $\delta \in [0,1]$.
\onem is an upper-bound for the linear interpolation which should be close to one.
The offset parameter $\delta$ can be explained due to the most likely higher probability for the proposed class.
We also find that this parameter is dataset dependent because for example with a lower image quality the annotator is inclined to accept a more unlikely proposal.
In \autoref{subsec:details}, we provide more details on how to calculate these values.

With this acceptance probability we can now generate simulated annotations $a'^x_{i,k} \in \{0,1\}$ as in \autoref{alg:spa} and describe the biased distribution similar to the unbiased distribution via $P(L_b^x = k) \approx \sum_{i=0}^{N'-1} \frac{a'^x_{i,k}}{N'}$ with $N'$ describing the number of simulated annotations.
The full source-code is in the supplementary and describes all corner cases e.g. $P(L^x \rho_x) = 1$.
An experimental validation of this method can be found in \autoref{subsec:eval_spa}.

\subsection{\spaplus}
\label{subsec:improve}

\paragraph{Class distribution Blending (CB)}
A label of an image is in general sample dependent but \cite{correlated_label_noise} showed that certain classes are more likely to be confused than others.
Thus, we propose to  blend the estimated distribution $P(L_b^x =k)$ with a class dependent probability distribution $c(\hat{k},k)$ to include this information.
This class probability distribution describes how likely $\hat{k}$ can be confused with any other given class $k$.
These probabilities can either be given by domain experts or approximated on a small subset of the data as shown in \autoref{subsec:details}.
The blending can be calculated as $\mu P(L_b^x = k) + (1-\mu) c(\hat{k},k) $ with the most likely class $\hat{k} = \text{argmax}_{j \in \{1,..,K\}}P(L_b^x = j)$ and blending parameter $\mu \in [0,1]$.
This approach can be interpreted as a smoothing of the estimated distribution which is especially useful in cases with a small number of annotations.

\paragraph{Bias Correction (BC)}
In \autoref{subsec:spa}, we proposed a model to use the knowledge of the unbiased distribution $P(L^x = k)$ to simulate the biased distribution $P(L_b^x = k)$ under the influence of the proposals $\rho_x$.
In this section, we formulate the reverse direction for correcting the bias to a certain degree.

According to \autoref{eq:accept}, for $k=\rho_x$ we can approximate
\begin{equation*}
    B := P(L^x = \rho_x) =   \frac{A-\delta}{\one-\delta} \approx \frac{\frac{|M_{\rho_x}|}{N'} - \delta}{\one  - \delta}, \label{eq:A_approx}
\end{equation*}
with $M_{\rho_x} = \{ i  ~|~ i\in \mathbb{N}, i \leq N', a'_{i,\rho_x} = 1 \}$ the indices of the annotations with an accepted proposal.
Note that we have to clamp the results to the interval $[0,1]$ to receive valid probabilities for numerical reasons.

For $k\neq\rho_x$ we deduce the probability from the reject case of \autoref{alg:spa}
\begin{align*}
P(L^x = k~|~L^x\neq \rho_x)
&=P(L^x_b = k~|~L^x\neq \rho_x)\\
\Leftrightarrow \frac{P(L^x = k, L^x\neq\rho_x)}{P(L^x\neq\rho_x)} &= P(L^x_b = k~|~L^x\neq \rho_x)\\
     \Leftrightarrow P( L^x \neq\rho_x)  &= (1-B)P(L^x_b = k~|~L^x\neq \rho_x)\\
     & \approx(1-B)\cdot \sum_{i \not\in M_{\rho_x}} \frac{a'_{i,k}}{N'-|M_{\rho_x}|}.
\end{align*}
This results in a approximate formula for the original ground truth distribution which relies only on the annotations with proposals.
The joined distribution is deducted  in the  supplementary.
It is important to note that the quality of these approximations relies on a large enough number of annotations $N'$.

\subsection{Implementation details}
\label{subsec:details}

We use a small user study which was proposed in \cite{dc3} to develop / verify our proposal acceptance on different subsets. 
The original data consists of four dataset with multiple annotations per image. 
We focus on the no proposal and class label proposal annotation approaches but the results for e.g. specific DC3 cluster proposals are similar and can be found in the supplementary. 

We calculated the ground-truth dataset dependent offset $\delta$ with a light weight approximation described in the supplementary.
An overview about the calculated offsets is given in \autoref{tbl:datasetOffsets} in combination with the values of the user study where applicable. 
Due to the fact, that it can not be expected, that this parameter can approximated in reality with a high precision we use for all experiment except otherwise stated, a balancing threshold $\mu = 0.75$, \onem = 0.99 and $\delta = 0.1$.
More details about the selection of these parameters are given in the supplementary.

\tbldatasetOffsets

The class distributions used for blending are approximated on 100 random images with 10 annotations sampled from the ground truth distribution. 
For a better comparability, we do not investigate different amounts of images and annotations for different datasets but we believe a lower cost solution is possible especially on smaller datasets such as QualityMRI.
For this reason, we ignore this static cost in the following evaluations.
If not otherwise stated, we use the method DivideMix  \cite{divide-mix} and its implementation in \cite{schmarje2022benchmark} to generate the proposals. 
With other methods the results are very similar and thus are excluded because they do not add further insights.
We include the original labels which are used to train the method in the outputted label distribution by blending it with the output in proportion to the used number of annotations.
Please see the supplementary for more details about the reproducibility.

\picTwo{micebone_pa}{Real $M_r$}{Simulated $M_s$}{Visual comparison the uncertainty bins for real vs. simulated proposal acceptance on the MiceBone dataset, Normalized per row / per proposal uncertainty bin, meaning that e.g. in Real $M_r$ that if a class with soft ground truth probability 0.21-0.4 is proposed, in 0.10 of cases a class with ground truth probability .041-0.6 is annotated. Hence,  some cells are 0 by default.
}

\section{Evaluation}
\label{sec:eval}

We show that SPA and our label improvements can be used to create / reverse a biased distribution, respectively. 
In three subsections, we show that both directions are technically feasible and are beneficial in practical applications.
Each section initially gives a short motivation, describes the evaluation metrics and provides the actual results.

 \picTwoLarge{simulated}{Synthetic}{Real}{Evaluation of label improvement on synthetic data created with SPA and real user study across different amounts of annotations. Results are clamped for visualization to the range 0 to 1.
 }

\subsection{Simulated Proposal Acceptance}
\label{subsec:eval_spa}

We need to verify that the our proposed method SPA is a good approximation of the reality in comparison to other methods and that an implementation is technically feasible.

\paragraph{Metrics \& Comparisons} 
We know for the evaluation of every image $x$ the proposed class $\rho_x$, the annotated class $a_x$ and the soft ground truth distribution $P(L^x = k)$ for class $k$ in the study~\cite{dc3}.
For the evaluation , we calculate a matrix between the proposed class probability and the actually annotated class probability by aggregating the probabilities into uncertainty bins with a size of 0.2 and a special bin for 0.0, which results in total in 6 bins.
Normalized examples are in \autoref{fig:micebone_pa}.

Our proposed method is composed of two parts: \textit{ACCEPT} with offset $\delta$ the proposal, otherwise use the \textit{GT} distribution for selection as defined in \autoref{subsec:spa}.
Other possible components would be \textit{RANDOM}ly annotate a label,  use the most \textit{LIKELY} class label or any combination of the before.
We compare our proposed method (ACCEPT+GT) against 6 other methods: ACCEPT+LIKELY, 2*ACCEPT+GT, 2*ACCEPT+RANDOM, RANDOM, GT, LIKELY.
Two ACCEPTS mean that we use an offset acceptance of the proposal and then an offset acceptance of the most likely class if it was not the original proposal.
As metric, we use half of the sum of differences (SOD) between the real matrix $M_r$ and the simulated matrix $M_s$ 
or as formula $SOD(M_{r},M_{s}) = 0.5 * (\sum_{i,j} \text{abs}(M_{r_{i,j}}- M_{s_{i,j}})) $
which is the number of differently assigned images to all bins asides from duplicates.
We report the normalized SOD by the total number of entries in $M_r$ as the average with standard deviation across three repetitions and include more results e.g. proposals based on DC3 clusters in the supplementary.
The $\delta$ of the User Study was used for the simulation.
We developed our method and the comparisons on the datasets Turkey and Plankton and only verified on  MiceBone and CIFAR10H. %

\paragraph{Results}

A visual comparison of the real and simulated results for all uncertainty bins can be seen in \autoref{fig:micebone_pa}.
The main diagonal line contains the accepted proposals while the rest, especially the upper right corner are the rejected images.
We see that the presented matrices are very similar, even in overlapping regions between accepted and rejected proposals as for uncertainty bin 0.41-0.6 of the proposed and annotated classes.

\tblSPAResults

In \autoref{tbl:SPAComparison}, we compare the proposed method with six other possible algorithms.
We see that our proposed method is for all datasets one of the best methods. 
However, some methods e.g. 2*ACCEPT+GT can sometimes even be better.
This data allows two main conclusions.
SPA is clearly better than some naive approaches like RANDOM or GT.
SPA is not optimal.
It can neither reproduce the real results completely nor is the best method across all datasets.
However, it shows very strong performance and is less complex then e.g. 2*ACCEPT + RANDOM.
We conclude that SPA is at a sweet spot between simplicity and correctness.

 \subsection{Label Improvement}
 \label{subsec:eval_improve}

 We show that CB and BC lead to similar increased results on simulated and real biased distributions  while the similarity illustrates the practical benefit of SPA.
 
 \paragraph{Metrics \& Comparison}
 As a metric, we use the Kullback-Leibler divergence \cite{kullback} between the soft ground truth $P(L^x = k)$ and the estimated distribution.
 We generate the skewed distributions either by our method SPA or use real proposal acceptance data from \cite{dc3}.
The reported results are the median performance across different annotation offsets or datasets for the synthetic and real data, respectively.
For the real data, we used the calculated $\delta$ defined in \autoref{tbl:datasetOffsets} for the simulation but as stated above $\delta = 0.1$ for the correction in \spaplus.
The method \emph{GT} is the baseline and samples annotations directly from $P(L^x = k)$.
The full results are in the supplementary.

\paragraph{Results}

If we look at the results on the synthetic data created by SPA in \autoref{fig:simulated-0}, we see the expected trend of improved results with more annotations.
While using only CB is the best method with one annotation, the performance is surpassed by all other methods with enough annotations.
The baseline (GT) is especially with the combination of blending (+CB) the best method for most number of annotations.
Our label improvement (\spaplus) is in most cases the second best method and blending is a major component (+CB).
The bias correction (+BC) improves the results further for higher number of annotations at around 20+.
Using the correct offset (+ $\delta$ GT) during the correction which was used in the simulation of SPA, is of lower importance. 
When we look at the full results in the supplementary, wee see benefits of a better $\delta$ at an offset larger than 0.4 and more annotations than 5.
We conclude that label improvement is possible for synthetic and real data and that the combination of CB and BC with an offset of 0.1 is in most cases the strongest improvement.

The real results in \autoref{fig:simulated-1} show the similar trends as in the synthetic data.
However, the baseline method without blending performs stronger and some trends are not observable because we only have up to 12 annotations.
The correct value for the offsets is even less important in the real data, most likely because the effect is diminished by the difference of the simulation and reality.
It is important to note that we keep the same notation with \spaplus but in this case SPA was not used to generate the biased distribution but real annotations with proposals.
Overall, the results analysis on synthetic and real data is similar and thus SPA can be used as a valid tool during method development.

It should be pointed out that the cost of labeling is not equivalent to the number annotations as we can expect a speedup of annotations when using proposals.
For example, \spaplus often performs slightly worse than GT in \autoref{fig:simulated-1}.
Considering a speedup of 2, we actually have to compare \spaplus with 5 annotations to GT at around 3, as explained in the budget calculation in \autoref{subsec:eval_benchmark}.

\subsection{Benchmark evaluation}
\label{subsec:eval_benchmark}

We show the results for \spaplus on~\cite{schmarje2022benchmark}.
  
\paragraph{Metrics \& Comparison}

We compare against the top three benchmarked methods: \emph{Baseline}, \emph{DivideMix} and \emph{Pseudo v2 soft}.
\emph{Baseline} just samples from the ground-truth but still performed the best with a high number of annotations. 
\emph{DivideMix} was proposed by~\cite{divide-mix} and \emph{Pseudo v2 soft (Pseudo soft)} uses Pseudo-Labels~\cite{pseudolabel} of soft labels to improve the labels.
We evaluate the Kullback-Leibler divergence ($KL$) \cite{kullback} between the ground truth and the output of the second stage (the evaluation with a fixed model) and KL between ground truth and the input of the second stage ($\hat{KL}$).
We also provide an additional ablation where we replaced the fixed model in the second sage with a visual transformer \cite{Dosovitskiy2021Vit}.
The hyperparameters of the transformer were not tuned for each dataset but kept to common recommendations.%
The speedup $S$ which can be expected due to using proposals depends on the dataset and used approach. 
For this reason, we include this parameter in our comparison with the values of 1 (no speedup), 2.5 as in \cite{dc3} or 10 as in \cite{morphocluster}.
$S$ is used to calculate the \emph{budget} as \emph{initial supervision per image (in. sup.)} $ + ($ \emph{percentage annotated of } $ X \cdot $ \emph{number of annotations per image}  $) / S)$.
In. sup. describes the percentage of labeled data that is annotated once in phase one of the benchmark.
For the skewed distribution generation which is correct by \spaplus, we used SPA with the calculated $\delta$ in \autoref{tbl:datasetOffsets}.
For CleverLabel a heuristically chosen $\delta = 0.1$ was used if not otherwise stated ($+ GT \delta$).
The results are the median scores of all datasets of the averages across three folds.
Full results including mean scores are in the supplementary.

\begin{figure}[!tbp]
  \centering
  \begin{subfigure}[b]{0.46\textwidth}
    \includegraphics[width=\textwidth]{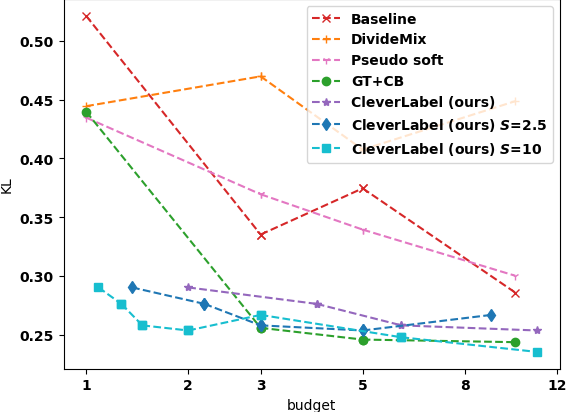}
    
    \caption{Comparisons of benchmark results with 100\% in. sup.}
    \label{fig:spa_supervised_median}
  \end{subfigure}
  \hfill
  \begin{subfigure}[b]{0.46\textwidth}
    \includegraphics[width=\textwidth]{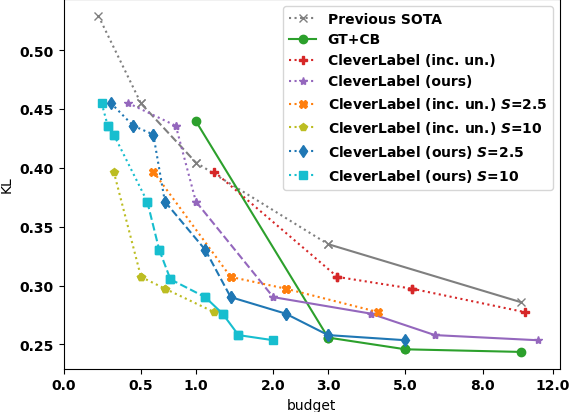}
    
    \caption{Pareto front visualization of benchmark results with 20,50, 100\% in. sup. %
}
\label{fig:spa_semi-supervised_median}
  \end{subfigure}

  \caption{Left: Compares previous state-of-the-art (first three), new baseline (GT+CB) and our method (\spaplus) including different speedups $S$. Right: The marker and color define the method, while the line style to the next node on the x-axis visualizes the initial supervision,  logarithmic scaled budgets}
\end{figure}

\paragraph{Results}

We present in \autoref{fig:spa_supervised_median} a comparison of our method \spaplus with previous state-of-the-art methods on the benchmark with an initial supervision of 100\%.
Even if we assume no speedup, we can achieve lower $KL$ scores than all previous methods, regardless of the used number of annotations.
Our proposed label improvement with class blending can also be applied to samples from the ground truth distribution (GT + CB) and achieves the best results in many cases.
Due to the fact that it does not leverage proposals it can not benefit from any speedups $S$.
If we take these speedups into consideration, \spaplus can achieve the best results across all budgets except for outliers.

We investigate lower budgets where the initial supervision could be below 100\% in \autoref{fig:spa_semi-supervised_median}.
The full results can be found in the supplementary.
If we compare our method to the combined Pareto front of all previous reported results, we see a clear improvement regardless of the expected speedup.
Two additional major interesting findings can be taken from our results.
Firstly, the \emph{percentage of labeled data} which is equal to the \emph{initial supervision} for \spaplus (violet,blue,lightblue)  is important as we see improved results from \emph{initial supervision} of 20 to 50 to 100\%.
This effect is mitigated with higher speedups because then \spaplus can achieve lower budget results not possible by other initial supervisions.
Secondly, we can improve the results further by using proposals also on the unlabeled data (inc. un., red,orange,yellow) after this initialization.
This increases the budget because the \emph{percentage of labeled data} is 100\% regardless of the \emph{initial supervision} but results in improved scores.
With $S=10$ we can even improve the previous state of the art (Pseudo soft, in. sup 20\%, 5 annotations) at the budget of 1.0 from 0.40/0.47 to 0.30/0.33 at a budget of 0.7 which is a relative improvement of 25\%/29.8\% with median/ mean aggregation.

In \autoref{tbl:BenchmarkAblationMedian}, we conduct several ablations to investigate the impact of individual parts of our method.
Comparing $KL$ and $\hat{KL}$ scores, we see similar trends between each other and to \autoref{subsec:eval_improve}.
Class blending (CB) is an important part of improved scores but the impact is stronger for $\hat{KL}$. 
A different blending threshold ($\mu = 0.25)$ which prefers the sample independent class distribution leads in most cases to similar or worse results than our selection of 0.75.
Bias Correction (BC) and the correct GT offset have a measurable impact on the $\hat{KL}$ while on $KL$ we almost no difference but a saturation at around 0.24 for all approaches most likely due the used network backbone.
With a different backbone e.g. a transformer~\cite{Dosovitskiy2021Vit} we can verify that BC positively impacts the results.

\section{Discussion}

\tblBenchmarkAblationMedian

In summary, we analyzed the introduced bias during the labeling process when using proposals by developing a simulation of this bias and provided two methods for reducing the proposal-introduced bias.
We could show that our methods outperform current state of the art methods on the same or even lower labeling budgets.
For low annotation budgets, we have even surpassed our newly proposed baseline of class blending in combination with annotation without proposals.
Cost is already a limiting factor when annotating data and thus only results with a better performance for a budget of less than one (which equals the current annotation of every image once) can be expected to be applied in real world applications.
We achieved this goal with \spaplus with speedups larger than 4 with is reasonable based on previously reported values~\cite{dc3}.

Based on our research, how should one annotate ambiguous image classification data?
While there currently is no strategy for every case, the problem can be broken down into the two major questions as depicted in \autoref{fig:guidelines}.
Firstly, is a bias in the data acceptable? 
Be aware that in \spaplus all labels are human validated and that many consensus process already use an agreement system~\cite{merIssue} with multiple reviewers. 
If a small bias is acceptable you can directly use proposals and an optional correction like \spaplus.
However, if a bias is not acceptable, the second major question is the expected speedup by using proposals for annotating your specific data. 
In case of a high expected speedup, the trade-off between the introduced bias and the ability to mitigate it with BC and CB favors \spaplus. 
For a low speedup, we recommend avoiding proposals and to rely on class blending which is applicable to any dataset if you can estimate the class transitions as described in \autoref{subsec:details}.
It is difficult to determine the exact trade-off point, because CB improves the results with fewer (10-) annotations, BC improves the results at above (20+) and both each other.
Based on this research, we recommend a rough speedup threshold of around three for the trade-off.

\pic{guidelines}{Flowchart about how to annotate ambiguous data based on the questions if an introduced bias is acceptable and if the expected speedup $S$ is high ($> 3$) }{0.85}

\subsection{Limitations}

We aim at a general approach for different datasets but this results in non-optimal solutions for individual datasets.
Multiple extensions for SPA like different kinds of simulated annotators would be possible but would require a larger user study for evaluation.
In \autoref{subsec:eval_improve}, we compared our simulation with real data on four datasets, but a larger comparison was not feasible.
It is important to note that SPA must not replace human evaluation but should be used for method development and hypothesis testing before an expensive human study which is needed to verify results. 
We gave a proof of concept about the benefit of bias correction with higher annotation counts with a stronger backbone like transformers. 
A full reevaluation of the benchmark was not feasible and it is questionable if it would lead to new insights because the scores might be lower but are expected to show similar relations.

\section{Conclusion}

Data quality is important but comes at a high cost.
Proposals can reduce this cost but introduce a bias.
We propose to mitigate this issue by simple heuristics and a theoretically motivated bias correction which makes them broader applicable and achieve up to 29.8\% relative better scores with reduced cost of 30\%.
This analysis is only possible due to our new proposed method SPA and results in general guidelines for how to annotate ambiguous data.

\FloatBarrier

\section*{Ethical Statement}

 While proposal-based labeling offers several benefits, it generally introduces a bias which might have ethical implications depending on the use case.
  We believe that it is important to take steps to improve proposal-guided annotations and introduce CleverLabel in order to enhance their quality by mitigating, however not eliminating, the effects of bias.
 Every operator must consciously decide whether the resulting reduced bias has a negative effect for their own application.
 CleverLabel aims at the utilization of all available data. As ambiguous labels can introduce additional data uncertainty into a model, while it is common to exclude these data from training.
 However, we expect that their consideration can also provide more nuanced information, potentially allowing for more accurate and fair decision-making. By this means, ambiguous labels may result in less overconfident models.

This work aims to facilitate and encourage further research in this field. While the contributions can be used to investigate a variety of research questions in order to improve the accuracy of predictions, we cannot identify any direct negative ethical impacts.

\bibliographystyle{splncs04}
\bibliography{lib}

\newpage
\appendix
\onecolumn

\section{Appendix}

\subsection{Theory for bias correction}

In the main paper, we calculated the approximated probabilities for the two cases in the if-clause in \autoref{alg:spa}.
These probabilities are
\begin{align}
\begin{split}
\label{eq:approx}
  B = P(L^x = k~|~ L^x = \rho_x)  &\approx \frac{\frac{|M_{\rho_x}|}{N'} - \delta}{0.99  - \delta}\\  
  C_k := P(L^x = k~|~ L^x \neq \rho_x) & \approx  \sum_{i \not\in M_{\rho_x}} \frac{a'_{i,k}}{N'-|M_{\rho_x}|}.
\end{split}
\end{align}
with $M_{\rho_x} = \{ i  ~|~ i\in \mathbb{N}, i \leq N', a'_{i,\rho_x} = 1 \}$ the indices of the $N'$ annotations with an accepted proposal for class $k$.

We can than estimate the complete distribution by
\begin{align}
\begin{split}
   & P(L^x = k) \\
   &= P(L^x = k, L^x = \rho_x) + P(L^x = k, L^x \neq \rho_x)  \\
    &= P(L^x = k ~|~  L^x = \rho_x)   P(L^x = \rho_x)  
     + P(L^x = k ~|~  L^x \neq \rho_x)  P(L^x \neq \rho_x)  \\
    &= \mathbf{1}_{\rho_x}(k) \cdot B + C_k \cdot  (1- B) \label{eq:split}
\end{split}
\end{align}
with indicator function $\mathbf{1}_{\rho_x}(k)$.

\subsection{The selection of parameters for SPA}

In \autoref{tbl:SPAparamters}, we provide an overview about other investigated values for the upper bound of \autoref{eq:accept}.
We call this upper bound \onem.
We did not investigate the value $\one = 1$ because we expect that proposals are always rejected by some annotators either due errors or different preferences.
This table also shows the results of the simulation for certain and ambiguous predictions of DC3 as defined in ~\cite{dc3}.
Please note that the provided results are from a previous version with slightly different implementation details and thus the results are not directly comparable to the ones in the main paper but quite close.
We see that a higher \onem is better across all datasets.
Moreover, DC3 predictions result in a similar score but predictions on certain data are a bit better while on ambiguous data they are marginally worse.
The parameters $\delta$ and $\mu$ were selected heuristically with the following motivations.
We normally want to correct only a bit with regard to $\delta$.
We weight the calculated distribution more than the class based distribution with $\mu$.
Both heuristically chosen values yielded robust results in preliminary studies.
Based on the ablation in the main paper, we see that a more realistic $\delta$ (with regard to the GT $\delta$) has a significant but small impact and a lower value of $\mu$ yields a inferior result.

\begin{table}[tb]
	\caption{
		Comparison of different values \onem for the proposal acceptance across normal and DC3 predicions.
	}
 \centering
	\label{tbl:SPAparamters}
		\begin{tabular}{l c c c c c c c c c c c c c c c c}
                     
Prediction &  \onem & CIFAR10H  & MiceBone &  Plankton & Turkey \\
 \midrule

Normal & 0.8 & 16.68  & 14.15 & 15.59 & 15.17\\
Normal & 0.9 & 9.16 & 7.72 & 6.77 & 8.19\\
Normal & 0.95 & 5.50  & 4.92 & 2.70 & 5.44\\
Normal & 0.99 & 2.43 & 5.09 & 4.26 & 4.24\\
DC3 Certain & 0.8 & 18.52 & 17.65 & 16.10 & 15.58\\
DC3 Certain & 0.9 & 9.83 & 9.14 & 7.67 & 8.94\\
DC3 Certain & 0.95 & 5.66 & 5.38 & 3.55 & 5.92\\
DC3 Certain & 0.99 & 1.95 & 2.52 & 3.21 & 4.77\\
DC3 Ambiguous & 0.8 & 16.99 & 12.64 & 16.51 & 13.46\\
DC3 Ambiguous & 0.9 & 10.83 & 8.15 & 9.28 & 7.33 \\
DC3 Ambiguous & 0.95 & 7.62 & 6.28 & 5.71 & 4.73\\
DC3 Ambiguous & 0.99 & 5.36 & 5.95 & 4.21 & 3.44 \\

 \end{tabular}
		
\end{table}

\subsection{Approximation of GT $\delta$ }

We estimate the dataset dependent offset $\delta$ based on 20 images with $0.2 < P(L^x = \rho_x) \leq 0.4$ for all datasets.
While this approach leads to the most robust results in comparison to the other reported methods below, we notice a constant underestimation of $\delta$ (see supplementary) and thus scale it with 1.3 to compensate for it.
We credit this issue to the fact that one author was the annotator and thus was unwillingly influenced by the known lower ground truth distribution.

Aside from the above presented method, we investigate two other non-working approaches for estimating $\delta$.
We report them here to let fellow researchers know what has not worked for us.

The first method is not using proposals with a ground truth probability of 0.2-0.4 but of 0.
In theory, this approach is equal to the proposed one in the paper. 
However, due to the fact that only low quality proposals are shown and the fact the annotator was aware of the effect, no annotations (except for Plankton) have been accepted. 
Thus, a calculation of $\delta$ was not possible.
This effect is still present in the data shown in the paper but does not decrease the result to zero which allows a rescaling as counter action.

The second method is based on the probability theory and our approximation of the introduced bias.
We can estimate for a second, different proposal $\rho'_x \neq \rho_x$ 
\begin{align}
\begin{split}
\label{eq:approx_delta}
  C'_k :=\: & P(L^x = k~|~  L^x \neq \rho'_x)  \approx  \sum_{i \not\in M_{\rho'_x}} \frac{a''_{i,k}}{N''-|M'_{\rho'_x}|}\\
  D_k :=\: & P(L^x = k, L^x = \rho'_x  ~|~ L^x \neq \rho_x) = \mathbf{1}_{\rho'_x}(k) \cdot  P(L^x = \rho'_x  ~|~ L^x \neq \rho_x)  \\ \approx & \mathbf{1}_{\rho'_x}(k) \cdot C_{\rho'_x} \\
  D'_k :=\: & P(L^x = k, L^x = \rho_x  ~|~ L^x \neq \rho'_x)  \approx \mathbf{1}_{\rho_x}(k) \cdot C'_{\rho_x} \\
  E_k :=\: & P(L^x = k~|~  L^x \neq \rho'_x,  L^x \neq \rho_x) \approx \sum_{a \in M_{\rho_x\rho'_x} } \frac{a_k}{|M''_{\rho_x\rho'_x}|} 
\end{split}
\end{align}
with $M'_{\rho'_x} = \{ i  ~|~ i\in \mathbb{N}, i \leq N'', a''_{i,\rho'_x} = 1 \}$ the indices of the $N''$ annotations ($a''$) with an accepted proposal and $M''_{\rho_x\rho'_x} = \{ a'_i  ~|~ i\in \mathbb{N}, i \leq N', a'_{i,\rho_x} \neq 1 \neq a'_{i,\rho'_x} \} \cup \{ a''_i  ~|~ i\in \mathbb{N}, i \leq N'', a''_{i,\rho_x} \neq 1 \neq a''_{i,\rho'_x}\}$ the  annotations not accepting one of the two proposals.

Firstly, we look at the corner cases $A \approx \frac{|M_{\rho_x}|}{N'}$ or $A' :=  \frac{|M'_{\rho'_x}|}{N''}$ are zero or one.
In these cases, we can approximate $\delta$ directly.
If $A \approx \frac{|M_{\rho_x}|}{N'} = 1$, we can make the simplified conclusion that $P(L^x = \rho_x) = 1$ and thus all other elements of the distribution are zero $P(L^x \neq \rho_x) = 0$.
Thus, we can approximate $\delta \approx A'$ because $P(L^x = \rho'_x) = 0$ and only $\delta$ can influence the result.
If $A \approx \frac{|M_{\rho_x}|}{N'} = 0$, we can conclude that $\delta = 0$ because $A$ should be greater or equal than $\delta$.
Analogously, we can conclude these results for $A'$. 

Assume none of the above cases are given and $P(L^x \neq \rho_x) = 0$, then $P(L^x = \rho_x) = 1$ and $A \approx 1$ which is aside from unlikely cases a contradiction to the cases above.
Thus, we can conclude most likely that $P(L^x \neq \rho_x) \neq 0$ and $P(L^x \neq \rho'_x) \neq 0$.

If none of the above cases applies and $E_k = 0$ for k, we can conclude that $M''_{\rho_x\rho'_x} = \{\}$ is empty and no annotations aside from $\rho_x$ and $\rho'_x$ exist and we can assume $P(L^x \not \in \{\rho_x,\rho'_x\}) = 0$.
We know from the definition $C'_{\rho_x} \approx P(L^x = \rho_x  ~|~ L^x \neq \rho'_x)$ and $C'_{\rho'_x} \approx P(L^x = \rho'_x  ~|~ L^x \neq \rho_x)$.
In combination with \autoref{eq:split}, we have two formulas for $P(L^x=k)$ which are conditioned both on the unknown $\delta$ which allows the calculation of it.

$\delta = 0$ because even classes $k$ with $P(L^x = k)$ should be annotated to some degree based if $\delta > 0$.

Based on the above conclusion, we can assume most likely for  $\rho_x \neq k \neq \rho'_x$, that $P(L^x \neq \rho_x) \neq 0$ , $P(L^x \neq \rho'_x) \neq 0$  and $E_k \neq 0$ or we could have already approximated $\delta$.

We know that for  $\rho_x \neq k \neq \rho'_x$
\begin{align}
\begin{split}
    &   P(L^x = k  ~|~ L^x \neq \rho_x) \cdot P( L^x \neq \rho_x) \\
    & =  P(L^x = k, L^x \neq \rho_x) \\
    & = P(L^x = k, L^x \neq \rho_x, L^x = \rho'_x) +  P(L^x = k, L^x \neq \rho_x, L^x \neq \rho'_x)\\
    & = P(L^x = k , L^x = \rho'_x ~|~ L^x \neq \rho_x) \cdot P(L^x \neq \rho_x)   \\
    & + P(L^x = k ~|~ L^x \neq \rho_x, L^x \neq \rho'_x) \cdot P( L^x \neq \rho_x, L^x \neq \rho'_x) \\
    & = P(L^x = k , L^x = \rho'_x ~|~ L^x \neq \rho_x) \cdot P(L^x \neq \rho_x)   \\
    & + P(L^x = k ~|~ L^x \neq \rho_x, L^x \neq \rho'_x) \cdot P( L^x \neq \rho'_x  ~|~  L^x \neq \rho_x) \cdot P( L^x \neq \rho_x) \\
\end{split}
\end{align}

Thus, by dividing by $P(L^x \neq \rho_x) \neq 0$, we get

\begin{align}
\begin{split}
    &   P(L^x = k  ~|~ L^x \neq \rho_x)  \\
    & = P(L^x = k , L^x = \rho'_x ~|~ L^x \neq \rho_x)   \\
    & + P(L^x = k ~|~ L^x \neq \rho_x, L^x \neq \rho'_x) \cdot P( L^x \neq \rho'_x  ~|~  L^x \neq \rho_x) \\
    & \Leftrightarrow C_k = D_k + E_k \cdot  P( L^x \neq \rho'_x  ~|~  L^x \neq \rho_x)
\end{split}
\end{align}
Analogously, we get also $C'_k = D'_k + E_k \cdot  P( L^x \neq \rho_x  ~|~  L^x \neq \rho'_x)$ for $P(L^x \neq \rho'_x) \neq 0$ 

We also know that 
\begin{align}
\begin{split}
    &   P( L^x \neq \rho_x, L^x \neq \rho'_x)  \\
    & = P( L^x \neq \rho'_x  ~|~  L^x \neq \rho_x) \cdot P( L^x \neq \rho_x)   \\
    & = P( L^x \neq \rho_x  ~|~  L^x \neq \rho'_x) \cdot P( L^x \neq \rho'_x) \\
    & \Leftrightarrow P( L^x \neq \rho_x) \\
    & = \frac{P( L^x \neq \rho'_x)}{P( L^x \neq \rho'_x  ~|~  L^x \neq \rho_x)} \cdot P( L^x \neq \rho_x  ~|~  L^x \neq \rho'_x)\\
    & \Leftrightarrow P( L^x \neq \rho'_x) \\
    & = \frac{P( L^x \neq \rho_x)}{P( L^x \neq \rho_x  ~|~  L^x \neq \rho'_x)} \cdot P( L^x \neq \rho'_x  ~|~  L^x \neq \rho_x)\\
\end{split}
\end{align}

If we assume that the fractures are about one, we get
\begin{align}
\begin{split}
    &   P( L^x \neq \rho_x) \approx P( L^x \neq \rho_x  ~|~  L^x \neq \rho'_x)\\
    &  P( L^x \neq \rho'_x) \approx  P( L^x \neq \rho'_x  ~|~  L^x \neq \rho_x).\\
\end{split}
\end{align}
This assumption is reasonable as a rough approximation because it is the same probability once with and once without the quite general inequality condition. 
Following the previous results we get 
\begin{align}
\begin{split}
    &   P( L^x \neq \rho_x) \approx \frac{C'_k - D'_k}{E_k}\\
    &  P( L^x \neq \rho'_x) \approx  \frac{C_k - D_k}{E_k}\\
\end{split}
\end{align}
for every $\rho_x \neq k \neq \rho'_x$.

We can estimate $\delta$ based on  $P( L^x = \rho_x) = 1-P( L^x \neq \rho_x)$  with 
\begin{align}
\begin{split}
    & A \approx \delta  + (0.99 - \delta) P( L^x = \rho_x)\\
    & \Leftrightarrow  A \approx \delta  + 0.99 \cdot  P( L^x = \rho_x) - \delta \cdot P( L^x = \rho_x) \\
    & \Leftrightarrow  A \approx \delta (1- P( L^x = \rho_x)) + 0.99 \cdot  P( L^x = \rho_x)  \\
    & \Leftrightarrow  \delta \approx \frac{A- 0.99 \cdot  P( L^x = \rho_x)}{P( L^x \neq \rho_x)}\\
\end{split}
\end{align}
Analogously, we can calculate $\delta$ for $P( L^x = \rho'_x) = 1-P( L^x \neq \rho'_x)$.

To summarize, we provided various approximations which can be used to estimate $\delta$ for every image $x$.
Based, on the assumption that $\delta$ is fixed, we have taken all calculated $\delta$ below a threshold of 0.8 and used the median to approximate $\delta$ for the complete dataset.
The threshold is reasonable to remove very unlikely candidates.

We can show that this median approximation generates reasonable values for $\delta$ on synthethic data. 
However, it is not robust and not reliably working on real data. 
We credit this issue to the fact that some of the above approximation are not exact enough and that the available real world data mostly include correlated proposals $\rho_x$ and $\rho'_x$.
This correlation is based on the fact that the most likely and the second most likely predictions of a network have been used.

\subsection{Implementation details of simulated proposal acceptance}

Below the main parts of the python source code are provided for SPA and BC. 

This code especially explains how we sample from the remaining classes in the cases $P(L_x = \rho_x) = 1$.
It is important to note that these cases can only happen in up to 1\% of the mentioned alternatives. 
Due the creation of the acceptance probability in SPA, the proposal $\rho_x$ is accepted in 99\% of the cases for  $P(L_x = \rho_x) = 1$.
We looked at our experiments and found at most several dozens of images where this issue occurred.
Based on the provided code, the first class is then simulated to be annotated.
In a reimplementation, it might be beneficial to change this behavior to a random class, but we do not estimate the major impact on the results based on the low number of samples.

\begin{lstlisting}[language=Python]
# soft_gt: vector for specific image of ground truth 
# probabilities
# proposal_acceptance_offset: offset for dataset,
# called \delta in paper
# simulation_repetitions: number of repetitions
# aka the number of simulated annotations
# proposed_class: the proposed class label 
# aka \rho_x in paper

# execute simualtion (multiple times)
simulated_label = np.zeros((len(soft_gt)))
for j in range(simulation_repetitions):
    # calculate proposal acceptance
    accept_rate = proposal_acceptance_offset +\
     ((0.99 - proposal_acceptance_offset)) *\
     soft_gt[proposed_class]

    simulated_class = -1
    # idea: accept rate increases with
    # raising soft gt value
    # random generated value between 0 and 1
    if random() <= accept_rate:
        simulated_class = proposed_class
    else:

        # idea 2: select based on soft gt
        # proposed can not be selected anymore
        max_value = 1 - soft_gt[proposed_class]  
        rand_select = random() * max_value

        sum_gt = 0
        # increase collective probability
        # until rand_select is smaller
        for k, g in enumerate(soft_gt):
            if proposed_class != k:
                # ignore proposed element
                # update collective probability
                sum_gt += g  
                if rand_select <= sum_gt:
                    simulated_class = k
                    break

        assert simulated_class != -1

    simulated_label[simulated_class] += 1

# check if the calculated distribution
# should be corrected with BC
if correct_with_offset < 0:

    # No corrections
    return simulated_label / simulation_repetitions
else:
    # correct with estimated offset

    # need to lower this score
    # based on confirmation offset
    pc = simulated_label[proposed_class] /
        simulation_repetitions
    corrected = (pc - correct_with_offset) /
        (0.99 - correct_with_offset)
    corrected = min(1, max(0, corrected))

    m = max(1, simulation_repetitions -
        simulated_label[proposed_class])
    p = simulated_label / m

    p *= (1 - corrected)  # rescale with leftovers
    p[proposed_class] = corrected

    return p
\end{lstlisting}

\subsection{Used transition matrices}

In this section, we report the used transition matrices $c$ for class distribution blending (see \autoref{subsec:improve}).
The matrices are given as a python dictionary to indicate which class belongs to which row.
Additionally we provide the expected Kullback-Leibler divergence when using only these blended probabilities as a ground truth distribution $P(L^x = k)$.

\begin{lstlisting}
# KL 0.4286257124664755
'Benthic': {
    'coral':        [0.814, 0.000, 0.000, 0.057, 
    0.114, 0.014, 0.000, 0.000],
    'crustacean':   [0.043, 0.843, 0.000, 0.000, 
    0.114, 0.000, 0.000, 0.000],
    'cucumber':     [0.000, 0.000, 0.900, 0.000,
    0.100, 0.000, 0.000, 0.000],
    'encrusting':   [0.024, 0.000, 0.000, 0.756, 
    0.040, 0.052, 0.000, 0.128],
    'other_fauna':  [0.021, 0.016, 0.000, 0.037,
    0.805, 0.042, 0.000, 0.079],
    'sponge':       [0.000, 0.019, 0.000, 0.062, 
    0.044, 0.844, 0.025, 0.006],
    'star':         [0.000, 0.000, 0.000, 0.000,
    0.030, 0.000, 0.970, 0.000],
    'worm':         [0.017, 0.000, 0.017, 0.075,
    0.058, 0.000, 0.000, 0.833]
},

# KL:  0.18304355047068008
'CIFAR10H':  {
    'airplane':     [0.950, 0.000, 0.000, 0.013,
    0.000, 0.000, 0.000, 0.000, 0.037, 0.000],
    'automobile':   [0.000, 0.978, 0.000, 0.000,
    0.000, 0.000, 0.000, 0.000, 0.000, 0.022],
    'bird':         [0.000, 0.000, 0.925, 0.008,
    0.033, 0.017, 0.000, 0.017, 0.000, 0.000],
    'cat':          [0.000, 0.000, 0.008, 0.875,
    0.017, 0.042, 0.042, 0.008, 0.000, 0.008],
    'deer':         [0.000, 0.000, 0.000, 0.000,
    0.929, 0.036, 0.000,
    0.036, 0.000, 0.000],
    'dog':          [0.000, 0.000, 0.000, 0.044,
    0.000, 0.956, 0.000, 0.000, 0.000, 0.000],
    'frog':         [0.000, 0.000, 0.017, 0.008, 
    0.000, 0.000, 0.975, 0.000, 0.000, 0.000],
    'horse':        [0.000, 0.000, 0.000, 0.000, 
    0.000, 0.000, 0.000, 1.000, 0.000, 0.000],
    'ship':         [0.008, 0.000, 0.000, 0.000,
    0.000, 0.000, 0.000, 0.000, 0.977, 0.015],
    'truck':        [0.014, 0.029, 0.000, 0.000,
    0.000, 0.000, 0.000, 0.000, 0.000, 0.957]},

# KL:  0.1822585877541493
'MiceBone': {
    'g':  [0.727, 0.180, 0.093],
    'nr': [0.033, 0.868, 0.099],
    'ug': [0.06, 0.167, 0.773]},

# KL:  0.28142176337024305
'Plankton': {
    'bubbles': [0.950, 0.000, 0.000, 0.000, 0.000,
    0.050, 0.000, 0.000, 0.000, 0.000],
    'collodaria_black': [0.000, 1.000, 0.000,
    0.000, 0.000, 0.000, 0.000, 0.000, 0.000, 0.000],
    'collodaria_globule': [0.000, 0.033, 0.900,
    0.000, 0.000, 0.067, 0.000, 0.000, 0.000, 0.000],
    'cop': [0.000, 0.000, 0.000, 0.911, 0.000, 0.033,
    0.000, 0.000, 0.000, 0.056],
    'det':  [0.033, 0.011, 0.000, 0.000, 0.800, 0.111,
    0.000, 0.044, 0.000, 0.000],
    'no_fit': [0.000, 0.000, 0.000, 0.015, 0.021,
    0.941, 0.000, 0.003, 0.009, 0.012],
    'phyto_puff': [0.000, 0.000, 0.000, 0.000, 0.025,
    0.000, 0.900, 0.075, 0.000, 0.000],
    'phyto_tuft': [0.013, 0.000, 0.000, 0.000, 0.031,
    0.000,  0.006, 0.950, 0.000, 0.000],
    'pro_rhizaria_phaeodaria': [0.000, 0.025, 0.008,
    0.000, 0.008, 0.033, 0.000, 0.000, 0.925, 0.000],
    'shrimp': [0.000, 0.000, 0.000, 0.000,
    0.017, 0.100, 0.000, 0.000, 0.000, 0.883]},

# KL: 0.3932698040768524
'Synthetic': {
    'bc': [0.845, 0.075, 0.025, 0.025, 0.030, 0.000],
    'be': [0.100, 0.767, 0.017, 0.050, 0.000, 0.067],
    'gc': [0.052, 0.012, 0.756, 0.104, 0.052, 0.024],
    'ge': [0.011, 0.061, 0.106, 0.761, 0.011, 0.050],
    'rc': [0.018, 0.018, 0.065, 0.006, 0.788, 0.106],
    're': [0.036, 0.021, 0.029, 0.029, 0.079, 0.807]},

# KL: 0.2856125819585248
'Pig':  {
    '1_intact':         [0.671, 0.160, 0.097, 0.071],
    '2_short':          [0.123, 0.650, 0.131, 0.096],
    '3_fresh':          [0.067, 0.153, 0.683, 0.097],
    '4_notVisible':     [0.067, 0.156, 0.156, 0.622]
},

# KL: 0.29562331399407754
'Treeversity#1':  {
    'bark':         [0.930, 0.000, 0.000, 0.000,
    0.060, 0.010],
    'bud':          [0.005, 0.819, 0.067, 0.067,
    0.038, 0.005],
    'flower':       [0.000, 0.034, 0.883, 0.046,
    0.031, 0.006],
    'fruit':        [0.000, 0.050, 0.036, 0.836,
    0.064, 0.014],
    'leaf':         [0.000, 0.033, 0.000, 0.033,
    0.889, 0.044],
    'whole_plant':  [0.009, 0.009, 0.018, 0.000,
    0.018, 0.945]
},

# KL: 0.34350036180164795
'Treeversity#6':  {
    'bark':         [0.812, 0.025, 0.025, 0.000,
    0.087, 0.050],
    'bud':          [0.130, 0.610, 0.075, 0.035,
    0.150, 0.000],
    'flower':       [0.024, 0.110, 0.741, 0.014,
    0.093, 0.017],
    'fruit':        [0.033, 0.042, 0.050, 0.700,
    0.175, 0.000],
    'leaf':         [0.040, 0.104, 0.084, 0.060,
    0.688, 0.024],
    'whole_plant':  [0.100, 0.000, 0.050, 0.017,
    0.133, 0.700]
},

# KL: 0.04067997329557361
'QualityMRI': {
    '0': [0.696, 0.304],
    '1': [0.240, 0.760]},

# KL: 0.13049150048780664
'Turkey': {
    'head_injury':      [0.833, 0.047, 0.120],
    'not_injured':      [0.060, 0.780, 0.160],
    'plumage_injury':   [0.012, 0.039, 0.949]},
\end{lstlisting}

\subsection{Full result tables}

This subsection provides all mentioned extended and full results of the paper.
They are best viewed digitally.

\subsubsection{Label improvement}

Full results from \autoref{subsec:eval_improve} in \autoref{tbl:SimulatedSPASynthetic} and \autoref{tbl:SimulatedSPAReal}.

  \tblSimulatedSPASynthetic
 \tblSimulatedSPAReal

\subsubsection{Benchmark}

Full results from \autoref{subsec:eval_benchmark} in \autoref{tbl:BenchmarkSupervised}, \autoref{tbl:BenchmarkSupervisedMedian} and \autoref{tbl:BenchmarkSupervisedMean} for \autoref{fig:spa_supervised_median}; in \autoref{tbl:BenchmarkSemiSupervised}, \autoref{tbl:BenchmarkSemiSupervisedMedian} and \autoref{tbl:BenchmarkSemiSupervisedMean} for \autoref{fig:spa_semi-supervised_median}; in \autoref{tbl:BenchmarkAblationMean},\autoref{tbl:BenchmarkAblationFirstPart},  \autoref{tbl:BenchmarkAblationSecondPart},  \autoref{tbl:BenchmarkAblationThirdPart} for \autoref{tbl:BenchmarkAblationMedian}.

\tblBenchmarkSupervisedMedian
\tblBenchmarkSupervisedMean

\tblBenchmarkSemiSupervisedMedian
\tblBenchmarkSemiSupervisedMean

\tblBenchmarkAblationMean

\tblBenchmarkSupervised
\tblBenchmarkSemiSupervised

\tblBenchmarkAblationFirstPart
\tblBenchmarkAblationSecondPart
\tblBenchmarkAblationThirdPart

\end{document}

%% file: images.tex
\graphicspath{{images/}}

\usepackage{caption}
\usepackage{subcaption}

\newcommand{\pic}[3]{
	\begin{figure}[tb] 
		\centering
		\includegraphics[width=#3\linewidth]{#1}
		\caption{#2}
		\label{fig:#1}
	\end{figure}
}

\newcommand{\picLarge}[3]{
	\begin{figure}[tb] 
  \begin{subfigure}[c]{0.98\linewidth}	
			\centering	
			\includegraphics[width=0.9\textwidth]{#1-legend}
		\end{subfigure}
  \begin{subfigure}[c]{0.98\linewidth}	
			\centering	
			\includegraphics[width=#3\textwidth]{#1}
   \label{fig:#1}
		\end{subfigure}
  \caption{#2}
	\end{figure*}
}

\newcommand{\picTwoLarge}[4]{
	\begin{figure*}[tb]
		\centering
        \begin{subfigure}[c]{0.95\linewidth}	
			\centering	
			\includegraphics[width=0.9\textwidth]{#1-legend}
		\end{subfigure}
		\begin{subfigure}[c]{0.45\linewidth}
			\centering		
			\includegraphics[width=0.9\textwidth]{#1-0}
			\subcaption{#2}	
			\label{fig:#1-0}	
		\end{subfigure}
		\begin{subfigure}[c]{0.45\linewidth}	
			\centering	
			\includegraphics[width=0.9\textwidth]{#1-1}
			\subcaption{#3}
			\label{fig:#1-1}	
		\end{subfigure}
		\caption{#4}
		\label{fig:#1}
	\end{figure*}
}

\newcommand{\picTwo}[4]{
	\begin{figure*}[tb]
		\centering
		\begin{subfigure}[c]{0.45\linewidth}
			\centering		
			\includegraphics[width=0.9\linewidth]{#1-0}
			\subcaption{#2}	
			\label{fig:#1-0}	
		\end{subfigure}
		\begin{subfigure}[c]{0.45\linewidth}	
			\centering	
			\includegraphics[width=0.9\linewidth]{#1-1}
			\subcaption{#3}
			\label{fig:#1-1}	
		\end{subfigure}
		\caption{#4}
		\label{fig:#1}
	\end{figure*}
}

%% file: tables.tex
\newcommand{\tbldatasetOffsets}{
\begin{table}[tb]
	\caption{
		Used offsets ($\delta$) for proposal  acceptance
	}
	\label{tbl:datasetOffsets}
	\resizebox{\linewidth}{!}{
		\begin{tabular}{p{2.2cm} p{2.2cm} p{2.2cm} p{2.2cm} p{2.2cm} p{2.2cm}}
          
          \toprule           
dataset  & Benthic &  CIFAR10H & MiceBone & Pig & Plankton  \\
 \midrule
User Study & N/A & 9.73\% &  36.36\% & N/A &  57.84\%  \\
Calculated & 40.17\% & 0.00\% & 41.03\% & 25.72\% & 64.81\% \\

\midrule
dataset  &  QualityMRI & Synthethic &  Treeversity\#1 & Treeversity\#6 & Turkey \\
 \midrule
User Study & N/A & N/A & N/A & N/A &  21.64\% \\
Calculated  & 0.00\% & 26.08\% & 26.08\% & 20.67\% & 14.17\% 
			
		\end{tabular}
	}
		
\end{table}
}

\newcommand{\tblSPAResults}{
\begin{table}[tb]
	\caption{
		Comparison of investigated methods for the proposal acceptance, 
        results within a one percent boundary of the best results are marked bold
		$^\dagger$ datasets used only for validation, 
	}
	\label{tbl:SPAComparison}
	\resizebox{\linewidth}{!}{
		\begin{tabular}{p{3.8cm} p{2.2cm} p{2.2cm} p{2.2cm} p{2.2cm}}
                     
 $SOD$ in [\%]  & CIFAR10H $^\dagger$ & MiceBone $^\dagger$ &  Plankton & Turkey \\
 \midrule
ACCEPT+GT (Ours) & \textbf{1.97 $\pm$ 0.06} & \textbf{2.57 $\pm$ 0.16} & \textbf{3.62 $\pm$ 0.05} & \textbf{4.77 $\pm$ 0.13} \\
 ACCEPT+LIKELY & \textbf{1.20 $\pm$ 0.25} & 7.88 $\pm$ 0.06 & 5.50 $\pm$ 0.15 & 7.21  $\pm$ 0.32\\    
2*ACCEPT+GT & 2.79 $\pm$ 0.22 & \textbf{3.38 $\pm$ 0.26} & \textbf{3.25 $\pm$ 0.13} & \textbf{3.78  $\pm$ 0.28} \\
 2*ACCEPT+RANDOM & 3.03 $\pm$ 0.11 & 4.76 $\pm$ 0.11 & \textbf{4.00 $\pm$ 0.31} & \textbf{4.64  $\pm$ 0.28}\\
 RANDOM & 86.37 $\pm$ 0.13 & 50.92 $\pm$ 0.64  & 81.51 $\pm$ 0.18 & 54.83  $\pm$ 0.93\\
 GT & 5.00 $\pm$ 0.03 & 10.34  $\pm$ 0.19 & 10.59 $\pm$ 0.11 & 5.69  $\pm$ 0.30\\
 LIKELY & 4.08 $\pm$ 0.00 & 16.41 $\pm$ 0.00 &  11.85 $\pm$ 0.00 & 11.47  $\pm$ 0.00 
 \end{tabular}
	}
		
\end{table}
}

\newcommand{\tblSimulatedSPAReal}{
\begin{table}[tb]
	\caption{
		Simulation of SPA on Real data, full results for visualized median aggregation in \autoref{fig:simulated-1}
	}
 \centering
	\label{tbl:SimulatedSPAReal}
		\begin{tabular}{l c c c c c c c c c c c c c c c c}
			\toprule

 dataset & method & 1 & 3 & 5 & 12 \\
 \midrule
CIFAR10H & GT & 0.71 & 0.25 & 0.21 & 0.13 \\
CIFAR10H & GT+CB & 0.42 & 0.11 & 0.11 & 0.07 \\
CIFAR10H & SPA & 0.81 & 0.53 & 0.37 & 0.30 \\
CIFAR10H & Only BC & 0.81 & 0.53 & 0.37 & 0.30 \\
CIFAR10H & Only CB & 0.42 & 0.34 & 0.28 & 0.20 \\
CIFAR10H & \spaplus (+ GT $\delta$) & 0.47 & 0.26 & 0.20 & 0.16 \\
CIFAR10H & \spaplus & 0.47 & 0.26 & 0.19 & 0.16 \\
Turkey & GT & 1.37 & 0.51 & 0.27 & 0.11 \\
Turkey & GT+CB & 0.35 & 0.14 & 0.08 & 0.04 \\
Turkey & SPA & 2.64 & 1.87 & 0.66 & 0.22 \\
Turkey & Only BC & 2.64 & 1.86 & 0.65 & 0.22 \\
Turkey & Only CB & 0.49 & 0.47 & 0.38 & 0.15 \\
Turkey & \spaplus (+ GT $\delta$) & 0.77 & 0.61 & 0.25 & 0.08 \\
Turkey & \spaplus & 0.77 & 0.61 & 0.26 & 0.08 \\
MiceBone & GT & 2.36 & 0.83 & 0.51 & 0.23 \\
MiceBone & GT+CB & 0.48 & 0.17 & 0.12 & 0.06 \\
MiceBone & SPA & 2.58 & 2.36 & 1.08 & 0.57 \\
MiceBone & Only BC & 2.58 & 2.36 & 1.17 & 0.61 \\
MiceBone & Only CB & 0.31 & 0.32 & 0.29 & 0.21 \\
MiceBone & \spaplus (+ GT $\delta$) & 0.56 & 0.54 & 0.29 & 0.15 \\
MiceBone & \spaplus & 0.56 & 0.54 & 0.26 & 0.16 \\
Plankton & GT & 0.91 & 0.35 & 0.20 & 0.07 \\
Plankton & GT+CB & 0.54 & 0.20 & 0.12 & 0.04 \\
Plankton & SPA & 1.71 & 1.39 & 1.00 & 0.59 \\
Plankton & Only BC & 1.71 & 1.65 & 1.28 & 0.77 \\
Plankton & Only CB & 1.00 & 1.75 & 1.74 & 0.77 \\
Plankton & \spaplus (+ GT $\delta$) & 1.10 & 1.12 & 0.85 & 0.50 \\
Plankton & \spaplus & 1.10 & 0.84 & 0.58 & 0.39 \\

		\end{tabular}
		
\end{table}
}

\newcommand{\tblSimulatedSPASynthetic}{
\begin{table}[tb]
	\caption{
		Simulation of SPA on Synthetic data, full results for visualized median aggregation in \autoref{fig:simulated-0}
	}
 \centering
	\label{tbl:SimulatedSPASynthetic}
		\begin{tabular}{l c c c c c c c c c c c c c c c c}
			\toprule

 offset & method & 1 & 3 & 5 & 10 & 20 & 50 & 100 \\
 \midrule
0 & GT & 3.69 & 1.47 & 0.89 & 0.43 & 0.21 & 0.08 & 0.04 \\
0 & GT+CB & 0.91 & 0.35 & 0.21 & 0.12 & 0.07 & 0.05 & 0.04 \\
0 & SPA & 3.67 & 1.46 & 0.90 & 0.43 & 0.20 & 0.08 & 0.03 \\
0 & Only BC & 3.72 & 1.49 & 0.90 & 0.42 & 0.20 & 0.07 & 0.04 \\
0 & Only CB & 0.56 & 0.46 & 0.39 & 0.37 & 0.33 & 0.34 & 0.32 \\
0 & \spaplus (+ GT $\delta$) & 0.90 & 0.34 & 0.21 & 0.12 & 0.07 & 0.05 & 0.04 \\
0 & \spaplus & 0.93 & 0.36 & 0.23 & 0.14 & 0.09 & 0.06 & 0.06 \\
0.1 & GT & 3.67 & 1.46 & 0.89 & 0.44 & 0.21 & 0.08 & 0.04 \\
0.1 & GT+CB & 0.86 & 0.32 & 0.21 & 0.12 & 0.08 & 0.05 & 0.04 \\
0.1 & SPA & 3.95 & 1.58 & 0.94 & 0.46 & 0.22 & 0.10 & 0.06 \\
0.1 & Only BC & 3.92 & 1.56 & 0.96 & 0.51 & 0.25 & 0.10 & 0.05 \\
0.1 & Only CB & 0.58 & 0.47 & 0.41 & 0.37 & 0.35 & 0.33 & 0.32 \\
0.1 & \spaplus (+ GT $\delta$) & 0.94 & 0.36 & 0.24 & 0.14 & 0.09 & 0.06 & 0.04 \\
0.1 & \spaplus & 0.95 & 0.36 & 0.24 & 0.14 & 0.09 & 0.06 & 0.05 \\
0.4 & GT & 3.71 & 1.48 & 0.89 & 0.45 & 0.20 & 0.08 & 0.04 \\
0.4 & GT+CB & 0.87 & 0.34 & 0.22 & 0.12 & 0.08 & 0.05 & 0.04 \\
0.4 & SPA & 4.75 & 2.22 & 1.44 & 0.82 & 0.48 & 0.31 & 0.25 \\
0.4 & Only BC & 4.64 & 2.56 & 1.69 & 0.86 & 0.42 & 0.18 & 0.09 \\
0.4 & Only CB & 0.75 & 0.61 & 0.54 & 0.45 & 0.39 & 0.33 & 0.34 \\
0.4 & \spaplus (+ GT $\delta$) & 1.20 & 0.63 & 0.44 & 0.24 & 0.14 & 0.08 & 0.06 \\
0.4 & \spaplus & 1.21 & 0.62 & 0.46 & 0.31 & 0.26 & 0.21 & 0.19 \\
0.6 & GT & 3.75 & 1.47 & 0.90 & 0.43 & 0.21 & 0.08 & 0.04 \\
0.6 & GT+CB & 0.96 & 0.35 & 0.21 & 0.12 & 0.08 & 0.05 & 0.04 \\
0.6 & SPA & 5.18 & 3.08 & 2.24 & 1.33 & 0.83 & 0.56 & 0.48 \\
0.6 & Only BC & 5.19 & 3.25 & 2.63 & 1.37 & 0.67 & 0.28 & 0.15 \\
0.6 & Only CB & 0.89 & 0.71 & 0.65 & 0.55 & 0.47 & 0.36 & 0.34 \\
0.6 & \spaplus (+ GT $\delta$) & 1.47 & 0.92 & 0.64 & 0.37 & 0.22 & 0.11 & 0.07 \\
0.6 & \spaplus & 1.47 & 0.91 & 0.71 & 0.56 & 0.50 & 0.46 & 0.46 \\

		\end{tabular}
		
\end{table}
}

\newcommand{\tblBenchmarkSupervisedMedian}{
\begin{table}[tb]
	\caption{
		Benchmark results  with 100\% in. sup. with median aggregation, metric: KL $\pm$ STD (budget), columns are the number of annotations used, $S$ describes the expected speedup
  }
	\label{tbl:BenchmarkSupervisedMedian}
 \centering
	\resizebox{\linewidth}{!}{
		\begin{tabular}{l c c c c c c c c c c c c c c c c}
			\toprule
	
	method & 1 & 3 & 5 & 10& 20 & 50 & 100 \\
\midrule		
Baseline &0.52 $\pm$ 0.04( 1.00) & 0.34 $\pm$ 0.02( 3.00) & 0.37 $\pm$ 0.03( 5.00) & 0.29 $\pm$ 0.02( 10.00) &   -   &   -   &   -   \\
 DivideMix &0.44 $\pm$ 0.05( 1.00) & 0.47 $\pm$ 0.03( 3.00) & 0.41 $\pm$ 0.04( 5.00) & 0.45 $\pm$ 0.04( 10.00) &   -   &   -   &   -   \\
 Pseudo soft &0.43 $\pm$ 0.07( 1.00) & 0.37 $\pm$ 0.03( 3.00) & 0.34 $\pm$ 0.02( 5.00) & 0.30 $\pm$ 0.02( 10.00) &   -   &   -   &   -   \\
 GT+CB &0.44 $\pm$ 0.03( 1.00) & 0.26 $\pm$ 0.01( 3.00) & 0.25 $\pm$ 0.01( 5.00) & 0.24 $\pm$ 0.01( 10.00) &   -   &   -   &   -   \\
 \spaplus (ours) &0.29 $\pm$ 0.02( 2.00) & 0.28 $\pm$ 0.01( 4.00) & 0.26 $\pm$ 0.02( 6.00) & 0.25 $\pm$ 0.01( 11.00) & 0.27 $\pm$ 0.02( 21.00) & 0.25 $\pm$ 0.02( 51.00) & 0.24 $\pm$ 0.01( 101.00) \\
 \spaplus (ours) $S$=2.5 &0.29 $\pm$ 0.02( 1.40) & 0.28 $\pm$ 0.01( 2.20) & 0.26 $\pm$ 0.02( 3.00) & 0.25 $\pm$ 0.01( 5.00) & 0.27 $\pm$ 0.02( 9.00) & 0.25 $\pm$ 0.02( 21.00) & 0.24 $\pm$ 0.01( 41.00) \\
 \spaplus (ours) $S$=10 &0.29 $\pm$ 0.02( 1.10) & 0.28 $\pm$ 0.01( 1.30) & 0.26 $\pm$ 0.02( 1.50) & 0.25 $\pm$ 0.01( 2.00) & 0.27 $\pm$ 0.02( 3.00) & 0.25 $\pm$ 0.02( 6.00) & 0.24 $\pm$ 0.01( 11.00) \\

		\end{tabular}
	}
		
\end{table}
}

\newcommand{\tblBenchmarkSupervisedMean}{
\begin{table}[tb]
	\caption{
		Benchmark results  with 100\% in. sup. with mean aggregation, metric: KL $\pm$ STD (budget), columns are the number of annotations used, $S$ describes the expected speedup
  }
	\label{tbl:BenchmarkSupervisedMean}
 \centering
	\resizebox{\linewidth}{!}{
		\begin{tabular}{l c c c c c c c c c c c c c c c c}
			\toprule
	
	method & 1 & 3 & 5 & 10 & 20 & 50 & 100\\
\midrule		
Baseline &0.69 $\pm$ 0.08( 1.00) & 0.37 $\pm$ 0.04( 3.00) & 0.39 $\pm$ 0.07( 5.00) & 0.33 $\pm$ 0.04( 10.00) &   -   &   -   &   -   \\
 DivideMix &0.52 $\pm$ 0.06( 1.00) & 0.56 $\pm$ 0.11( 3.00) & 0.54 $\pm$ 0.05( 5.00) & 0.64 $\pm$ 0.14( 10.00) &   -   &   -   &   -   \\
 Pseudo soft &0.52 $\pm$ 0.07( 1.00) & 0.39 $\pm$ 0.04( 3.00) & 0.36 $\pm$ 0.03( 5.00) & 0.33 $\pm$ 0.02( 10.00) &   -   &   -   &   -   \\
 GT+CB &0.60 $\pm$ 0.29( 1.00) & 0.32 $\pm$ 0.03( 3.00) & 0.30 $\pm$ 0.01( 5.00) & 0.30 $\pm$ 0.02( 10.00) &   -   &   -   &   -   \\
 \spaplus (ours) &0.37 $\pm$ 0.02( 2.00) & 0.33 $\pm$ 0.02( 4.00) & 0.32 $\pm$ 0.03( 6.00) & 0.30 $\pm$ 0.02( 11.00) & 0.31 $\pm$ 0.03( 21.00) & 0.29 $\pm$ 0.02( 51.00) & 0.29 $\pm$ 0.02( 101.00) \\
 \spaplus (ours) $S$=2.5 &0.37 $\pm$ 0.02( 1.40) & 0.33 $\pm$ 0.02( 2.20) & 0.32 $\pm$ 0.03( 3.00) & 0.30 $\pm$ 0.02( 5.00) & 0.31 $\pm$ 0.03( 9.00) & 0.29 $\pm$ 0.02( 21.00) & 0.29 $\pm$ 0.02( 41.00) \\
 \spaplus (ours) $S$=10 &0.37 $\pm$ 0.02( 1.10) & 0.33 $\pm$ 0.02( 1.30) & 0.32 $\pm$ 0.03( 1.50) & 0.30 $\pm$ 0.02( 2.00) & 0.31 $\pm$ 0.03( 3.00) & 0.29 $\pm$ 0.02( 6.00) & 0.29 $\pm$ 0.02( 11.00) \\

		\end{tabular}
	}
		
\end{table}
}

\newcommand{\tblBenchmarkSupervised}{
\begin{table*}[tb]
	\caption{
		Benchmark results  with 100\% in. sup. without aggregation, metric: KL $\pm$ STD (budget), columns are the number of annotations used, $S$ describes the expected speedup
	}
	\label{tbl:BenchmarkSupervised}
	\resizebox{\linewidth}{!}{
		\begin{tabular}{l c c c c c c c c c c c c c c c c}
			\toprule
			
	 dataset & method & 1 & 3 & 5 & 10 & 20 & 50 & 100 \\
	 \midrule
 Benthic & Baseline &1.17 $\pm$ 0.04( 1.00) & 0.81 $\pm$ 0.04( 3.00) & 0.76 $\pm$ 0.04( 5.00) & 0.75 $\pm$ 0.03( 10.00) &   -   &   -   &   -   \\
 Benthic & DivideMix &0.87 $\pm$ 0.07( 1.00) & 0.82 $\pm$ 0.00( 3.00) & 0.84 $\pm$ 0.02( 5.00) & 0.83 $\pm$ 0.05( 10.00) &   -   &   -   &   -   \\
 Benthic & Pseudo soft &1.00 $\pm$ 0.08( 1.00) & 0.76 $\pm$ 0.06( 3.00) & 0.71 $\pm$ 0.05( 5.00) & 0.70 $\pm$ 0.02( 10.00) &   -   &   -   &   -   \\
 Benthic & GT+CB &0.81 $\pm$ 0.04( 1.00) & 0.68 $\pm$ 0.02( 3.00) & 0.67 $\pm$ 0.00( 5.00) & 0.65 $\pm$ 0.01( 10.00) &   -   &   -   &   -   \\
 Benthic & \spaplus (ours) &0.78 $\pm$ 0.03( 2.00) & 0.76 $\pm$ 0.06( 4.00) & 0.71 $\pm$ 0.05( 6.00) & 0.67 $\pm$ 0.02( 11.00) & 0.68 $\pm$ 0.02( 21.00) & 0.67 $\pm$ 0.06( 51.00) & 0.65 $\pm$ 0.03( 101.00) \\
 Benthic & \spaplus (ours) $S$=2.5 &0.78 $\pm$ 0.03( 1.40) & 0.76 $\pm$ 0.06( 2.20) & 0.71 $\pm$ 0.05( 3.00) & 0.67 $\pm$ 0.02( 5.00) & 0.68 $\pm$ 0.02( 9.00) & 0.67 $\pm$ 0.06( 21.00) & 0.65 $\pm$ 0.03( 41.00) \\
 Benthic & \spaplus (ours) $S$=10 &0.78 $\pm$ 0.03( 1.10) & 0.76 $\pm$ 0.06( 1.30) & 0.71 $\pm$ 0.05( 1.50) & 0.67 $\pm$ 0.02( 2.00) & 0.68 $\pm$ 0.02( 3.00) & 0.67 $\pm$ 0.06( 6.00) & 0.65 $\pm$ 0.03( 11.00) \\
 CIFAR10H & Baseline &0.41 $\pm$ 0.02( 1.00) & 0.30 $\pm$ 0.02( 3.00) & 0.27 $\pm$ 0.03( 5.00) & 0.26 $\pm$ 0.02( 10.00) &   -   &   -   &   -   \\
 CIFAR10H & DivideMix &0.36 $\pm$ 0.02( 1.00) & 0.35 $\pm$ 0.01( 3.00) & 0.35 $\pm$ 0.05( 5.00) & 0.37 $\pm$ 0.03( 10.00) &   -   &   -   &   -   \\
 CIFAR10H & Pseudo soft &0.41 $\pm$ 0.02( 1.00) & 0.41 $\pm$ 0.02( 3.00) & 0.42 $\pm$ 0.01( 5.00) & 0.43 $\pm$ 0.02( 10.00) &   -   &   -   &   -   \\
 CIFAR10H & GT+CB &0.27 $\pm$ 0.01( 1.00) & 0.25 $\pm$ 0.01( 3.00) & 0.24 $\pm$ 0.02( 5.00) & 0.25 $\pm$ 0.01( 10.00) &   -   &   -   &   -   \\
 CIFAR10H & \spaplus (ours) &0.25 $\pm$ 0.01( 2.00) & 0.26 $\pm$ 0.01( 4.00) & 0.24 $\pm$ 0.01( 6.00) & 0.24 $\pm$ 0.01( 11.00) & 0.24 $\pm$ 0.01( 21.00) & 0.25 $\pm$ 0.02( 51.00) & 0.24 $\pm$ 0.01( 101.00) \\
 CIFAR10H & \spaplus (ours) $S$=2.5 &0.25 $\pm$ 0.01( 1.40) & 0.26 $\pm$ 0.01( 2.20) & 0.24 $\pm$ 0.01( 3.00) & 0.24 $\pm$ 0.01( 5.00) & 0.24 $\pm$ 0.01( 9.00) & 0.25 $\pm$ 0.02( 21.00) & 0.24 $\pm$ 0.01( 41.00) \\
 CIFAR10H & \spaplus (ours) $S$=10 &0.25 $\pm$ 0.01( 1.10) & 0.26 $\pm$ 0.01( 1.30) & 0.24 $\pm$ 0.01( 1.50) & 0.24 $\pm$ 0.01( 2.00) & 0.24 $\pm$ 0.01( 3.00) & 0.25 $\pm$ 0.02( 6.00) & 0.24 $\pm$ 0.01( 11.00) \\
 MiceBone & Baseline &0.55 $\pm$ 0.06( 1.00) & 0.27 $\pm$ 0.03( 3.00) & 0.27 $\pm$ 0.02( 5.00) & 0.23 $\pm$ 0.02( 10.00) &   -   &   -   &   -   \\
 MiceBone & DivideMix &0.38 $\pm$ 0.05( 1.00) & 0.30 $\pm$ 0.04( 3.00) & 0.34 $\pm$ 0.02( 5.00) & 0.35 $\pm$ 0.05( 10.00) &   -   &   -   &   -   \\
 MiceBone & Pseudo soft &0.40 $\pm$ 0.08( 1.00) & 0.29 $\pm$ 0.03( 3.00) & 0.27 $\pm$ 0.04( 5.00) & 0.22 $\pm$ 0.02( 10.00) &   -   &   -   &   -   \\
 MiceBone & GT+CB &0.35 $\pm$ 0.05( 1.00) & 0.24 $\pm$ 0.01( 3.00) & 0.25 $\pm$ 0.01( 5.00) & 0.23 $\pm$ 0.02( 10.00) &   -   &   -   &   -   \\
 MiceBone & \spaplus (ours) &0.31 $\pm$ 0.03( 2.00) & 0.30 $\pm$ 0.00( 4.00) & 0.27 $\pm$ 0.01( 6.00) & 0.26 $\pm$ 0.04( 11.00) & 0.25 $\pm$ 0.04( 21.00) & 0.25 $\pm$ 0.02( 51.00) & 0.23 $\pm$ 0.03( 101.00) \\
 MiceBone & \spaplus (ours) $S$=2.5 &0.31 $\pm$ 0.03( 1.40) & 0.30 $\pm$ 0.00( 2.20) & 0.27 $\pm$ 0.01( 3.00) & 0.26 $\pm$ 0.04( 5.00) & 0.25 $\pm$ 0.04( 9.00) & 0.25 $\pm$ 0.02( 21.00) & 0.23 $\pm$ 0.03( 41.00) \\
 MiceBone & \spaplus (ours) $S$=10 &0.31 $\pm$ 0.03( 1.10) & 0.30 $\pm$ 0.00( 1.30) & 0.27 $\pm$ 0.01( 1.50) & 0.26 $\pm$ 0.04( 2.00) & 0.25 $\pm$ 0.04( 3.00) & 0.25 $\pm$ 0.02( 6.00) & 0.23 $\pm$ 0.03( 11.00) \\
 Pig & Baseline &0.75 $\pm$ 0.05( 1.00) & 0.56 $\pm$ 0.02( 3.00) & 0.62 $\pm$ 0.04( 5.00) & 0.54 $\pm$ 0.04( 10.00) &   -   &   -   &   -   \\
 Pig & DivideMix &0.95 $\pm$ 0.04( 1.00) & 1.11 $\pm$ 0.32( 3.00) & 1.36 $\pm$ 0.19( 5.00) & 1.04 $\pm$ 0.25( 10.00) &   -   &   -   &   -   \\
 Pig & Pseudo soft &0.70 $\pm$ 0.06( 1.00) & 0.63 $\pm$ 0.03( 3.00) & 0.59 $\pm$ 0.04( 5.00) & 0.61 $\pm$ 0.07( 10.00) &   -   &   -   &   -   \\
 Pig & GT+CB &0.67 $\pm$ 0.08( 1.00) & 0.55 $\pm$ 0.01( 3.00) & 0.51 $\pm$ 0.01( 5.00) & 0.51 $\pm$ 0.04( 10.00) &   -   &   -   &   -   \\
 Pig & \spaplus (ours) &0.61 $\pm$ 0.02( 2.00) & 0.54 $\pm$ 0.02( 4.00) & 0.62 $\pm$ 0.14( 6.00) & 0.51 $\pm$ 0.00( 11.00) & 0.58 $\pm$ 0.08( 21.00) & 0.50 $\pm$ 0.02( 51.00) & 0.50 $\pm$ 0.02( 101.00) \\
 Pig & \spaplus (ours) $S$=2.5 &0.61 $\pm$ 0.02( 1.40) & 0.54 $\pm$ 0.02( 2.20) & 0.62 $\pm$ 0.14( 3.00) & 0.51 $\pm$ 0.00( 5.00) & 0.58 $\pm$ 0.08( 9.00) & 0.50 $\pm$ 0.02( 21.00) & 0.50 $\pm$ 0.02( 41.00) \\
 Pig & \spaplus (ours) $S$=10 &0.61 $\pm$ 0.02( 1.10) & 0.54 $\pm$ 0.02( 1.30) & 0.62 $\pm$ 0.14( 1.50) & 0.51 $\pm$ 0.00( 2.00) & 0.58 $\pm$ 0.08( 3.00) & 0.50 $\pm$ 0.02( 6.00) & 0.50 $\pm$ 0.02( 11.00) \\
 Plankton & Baseline &0.34 $\pm$ 0.02( 1.00) & 0.23 $\pm$ 0.02( 3.00) & 0.22 $\pm$ 0.01( 5.00) & 0.20 $\pm$ 0.02( 10.00) &   -   &   -   &   -   \\
 Plankton & DivideMix &0.34 $\pm$ 0.01( 1.00) & 0.34 $\pm$ 0.01( 3.00) & 0.34 $\pm$ 0.01( 5.00) & 0.34 $\pm$ 0.01( 10.00) &   -   &   -   &   -   \\
 Plankton & Pseudo soft &0.32 $\pm$ 0.06( 1.00) & 0.33 $\pm$ 0.08( 3.00) & 0.24 $\pm$ 0.03( 5.00) & 0.25 $\pm$ 0.01( 10.00) &   -   &   -   &   -   \\
 Plankton & GT+CB &0.25 $\pm$ 0.01( 1.00) & 0.21 $\pm$ 0.02( 3.00) & 0.20 $\pm$ 0.01( 5.00) & 0.20 $\pm$ 0.01( 10.00) &   -   &   -   &   -   \\
 Plankton & \spaplus (ours) &0.26 $\pm$ 0.01( 2.00) & 0.23 $\pm$ 0.01( 4.00) & 0.23 $\pm$ 0.01( 6.00) & 0.22 $\pm$ 0.01( 11.00) & 0.22 $\pm$ 0.01( 21.00) & 0.22 $\pm$ 0.01( 51.00) & 0.22 $\pm$ 0.01( 101.00) \\
 Plankton & \spaplus (ours) $S$=2.5 &0.26 $\pm$ 0.01( 1.40) & 0.23 $\pm$ 0.01( 2.20) & 0.23 $\pm$ 0.01( 3.00) & 0.22 $\pm$ 0.01( 5.00) & 0.22 $\pm$ 0.01( 9.00) & 0.22 $\pm$ 0.01( 21.00) & 0.22 $\pm$ 0.01( 41.00) \\
 Plankton & \spaplus (ours) $S$=10 &0.26 $\pm$ 0.01( 1.10) & 0.23 $\pm$ 0.01( 1.30) & 0.23 $\pm$ 0.01( 1.50) & 0.22 $\pm$ 0.01( 2.00) & 0.22 $\pm$ 0.01( 3.00) & 0.22 $\pm$ 0.01( 6.00) & 0.22 $\pm$ 0.01( 11.00) \\
 QualityMRI & Baseline &1.73 $\pm$ 0.48( 1.00) & 0.39 $\pm$ 0.19( 3.00) & 0.52 $\pm$ 0.28( 5.00) & 0.31 $\pm$ 0.14( 10.00) &   -   &   -   &   -   \\
 QualityMRI & DivideMix &0.46 $\pm$ 0.26( 1.00) & 0.60 $\pm$ 0.58( 3.00) & 0.17 $\pm$ 0.05( 5.00) & 1.34 $\pm$ 0.84( 10.00) &   -   &   -   &   -   \\
 QualityMRI & Pseudo soft &0.62 $\pm$ 0.16( 1.00) & 0.21 $\pm$ 0.10( 3.00) & 0.28 $\pm$ 0.03( 5.00) & 0.12 $\pm$ 0.04( 10.00) &   -   &   -   &   -   \\
 QualityMRI & GT+CB &0.47 $\pm$ 0.10( 1.00) & 0.22 $\pm$ 0.17( 3.00) & 0.15 $\pm$ 0.01( 5.00) & 0.24 $\pm$ 0.13( 10.00) &   -   &   -   &   -   \\
 QualityMRI & \spaplus (ours) &0.24 $\pm$ 0.09( 2.00) & 0.12 $\pm$ 0.03( 4.00) & 0.11 $\pm$ 0.07( 6.00) & 0.11 $\pm$ 0.05( 11.00) & 0.10 $\pm$ 0.04( 21.00) & 0.12 $\pm$ 0.04( 51.00) & 0.16 $\pm$ 0.04( 101.00) \\
 QualityMRI & \spaplus (ours) $S$=2.5 &0.24 $\pm$ 0.09( 1.40) & 0.12 $\pm$ 0.03( 2.20) & 0.11 $\pm$ 0.07( 3.00) & 0.11 $\pm$ 0.05( 5.00) & 0.10 $\pm$ 0.04( 9.00) & 0.12 $\pm$ 0.04( 21.00) & 0.16 $\pm$ 0.04( 41.00) \\
 QualityMRI & \spaplus (ours) $S$=10 &0.24 $\pm$ 0.09( 1.10) & 0.12 $\pm$ 0.03( 1.30) & 0.11 $\pm$ 0.07( 1.50) & 0.11 $\pm$ 0.05( 2.00) & 0.10 $\pm$ 0.04( 3.00) & 0.12 $\pm$ 0.04( 6.00) & 0.16 $\pm$ 0.04( 11.00) \\
 Synthetic & Baseline &0.08 $\pm$ 0.01( 1.00) & 0.05 $\pm$ 0.00( 3.00) & 0.06 $\pm$ 0.01( 5.00) & 0.04 $\pm$ 0.00( 10.00) &   -   &   -   &   -   \\
 Synthetic & DivideMix &0.33 $\pm$ 0.00( 1.00) & 0.48 $\pm$ 0.06( 3.00) & 0.42 $\pm$ 0.07( 5.00) & 0.43 $\pm$ 0.04( 10.00) &   -   &   -   &   -   \\
 Synthetic & Pseudo soft &0.10 $\pm$ 0.01( 1.00) & 0.07 $\pm$ 0.00( 3.00) & 0.08 $\pm$ 0.01( 5.00) & 0.06 $\pm$ 0.01( 10.00) &   -   &   -   &   -   \\
 Synthetic & GT+CB &0.10 $\pm$ 0.00( 1.00) & 0.08 $\pm$ 0.00( 3.00) & 0.07 $\pm$ 0.01( 5.00) & 0.07 $\pm$ 0.00( 10.00) &   -   &   -   &   -   \\
 Synthetic & \spaplus (ours) &0.08 $\pm$ 0.01( 2.00) & 0.07 $\pm$ 0.01( 4.00) & 0.08 $\pm$ 0.00( 6.00) & 0.08 $\pm$ 0.00( 11.00) & 0.08 $\pm$ 0.00( 21.00) & 0.07 $\pm$ 0.01( 51.00) & 0.07 $\pm$ 0.00( 101.00) \\
 Synthetic & \spaplus (ours) $S$=2.5 &0.08 $\pm$ 0.01( 1.40) & 0.07 $\pm$ 0.01( 2.20) & 0.08 $\pm$ 0.00( 3.00) & 0.08 $\pm$ 0.00( 5.00) & 0.08 $\pm$ 0.00( 9.00) & 0.07 $\pm$ 0.01( 21.00) & 0.07 $\pm$ 0.00( 41.00) \\
 Synthetic & \spaplus (ours) $S$=10 &0.08 $\pm$ 0.01( 1.10) & 0.07 $\pm$ 0.01( 1.30) & 0.08 $\pm$ 0.00( 1.50) & 0.08 $\pm$ 0.00( 2.00) & 0.08 $\pm$ 0.00( 3.00) & 0.07 $\pm$ 0.01( 6.00) & 0.07 $\pm$ 0.00( 11.00) \\
 Treeversity\#1 & Baseline &0.49 $\pm$ 0.04( 1.00) & 0.37 $\pm$ 0.01( 3.00) & 0.37 $\pm$ 0.00( 5.00) & 0.36 $\pm$ 0.02( 10.00) &   -   &   -   &   -   \\
 Treeversity\#1 & DivideMix &0.47 $\pm$ 0.01( 1.00) & 0.46 $\pm$ 0.01( 3.00) & 0.46 $\pm$ 0.01( 5.00) & 0.46 $\pm$ 0.01( 10.00) &   -   &   -   &   -   \\
 Treeversity\#1 & Pseudo soft &0.46 $\pm$ 0.01( 1.00) & 0.41 $\pm$ 0.02( 3.00) & 0.39 $\pm$ 0.02( 5.00) & 0.37 $\pm$ 0.02( 10.00) &   -   &   -   &   -   \\
 Treeversity\#1 & GT+CB &0.41 $\pm$ 0.01( 1.00) & 0.37 $\pm$ 0.02( 3.00) & 0.35 $\pm$ 0.02( 5.00) & 0.36 $\pm$ 0.02( 10.00) &   -   &   -   &   -   \\
 Treeversity\#1 & \spaplus (ours) &0.39 $\pm$ 0.02( 2.00) & 0.37 $\pm$ 0.01( 4.00) & 0.37 $\pm$ 0.02( 6.00) & 0.35 $\pm$ 0.01( 11.00) & 0.35 $\pm$ 0.01( 21.00) & 0.35 $\pm$ 0.02( 51.00) & 0.35 $\pm$ 0.01( 101.00) \\
 Treeversity\#1 & \spaplus (ours) $S$=2.5 &0.39 $\pm$ 0.02( 1.40) & 0.37 $\pm$ 0.01( 2.20) & 0.37 $\pm$ 0.02( 3.00) & 0.35 $\pm$ 0.01( 5.00) & 0.35 $\pm$ 0.01( 9.00) & 0.35 $\pm$ 0.02( 21.00) & 0.35 $\pm$ 0.01( 41.00) \\
 Treeversity\#1 & \spaplus (ours) $S$=10 &0.39 $\pm$ 0.02( 1.10) & 0.37 $\pm$ 0.01( 1.30) & 0.37 $\pm$ 0.02( 1.50) & 0.35 $\pm$ 0.01( 2.00) & 0.35 $\pm$ 0.01( 3.00) & 0.35 $\pm$ 0.02( 6.00) & 0.35 $\pm$ 0.01( 11.00) \\
 Treeversity\#6 & Baseline &1.02 $\pm$ 0.03( 1.00) & 0.50 $\pm$ 0.02( 3.00) & 0.38 $\pm$ 0.00( 5.00) & 0.33 $\pm$ 0.01( 10.00) &   -   &   -   &   -   \\
 Treeversity\#6 & DivideMix &0.62 $\pm$ 0.05( 1.00) & 0.71 $\pm$ 0.02( 3.00) & 0.70 $\pm$ 0.04( 5.00) & 0.77 $\pm$ 0.03( 10.00) &   -   &   -   &   -   \\
 Treeversity\#6 & Pseudo soft &0.83 $\pm$ 0.13( 1.00) & 0.47 $\pm$ 0.02( 3.00) & 0.40 $\pm$ 0.02( 5.00) & 0.35 $\pm$ 0.01( 10.00) &   -   &   -   &   -   \\
 Treeversity\#6 & GT+CB &0.61 $\pm$ 0.03( 1.00) & 0.37 $\pm$ 0.00( 3.00) & 0.33 $\pm$ 0.01( 5.00) & 0.31 $\pm$ 0.00( 10.00) &   -   &   -   &   -   \\
 Treeversity\#6 & \spaplus (ours) &0.47 $\pm$ 0.01( 2.00) & 0.37 $\pm$ 0.02( 4.00) & 0.34 $\pm$ 0.02( 6.00) & 0.33 $\pm$ 0.01( 11.00) & 0.30 $\pm$ 0.01( 21.00) & 0.30 $\pm$ 0.01( 51.00) & 0.28 $\pm$ 0.00( 101.00) \\
 Treeversity\#6 & \spaplus (ours) $S$=2.5 &0.47 $\pm$ 0.01( 1.40) & 0.37 $\pm$ 0.02( 2.20) & 0.34 $\pm$ 0.02( 3.00) & 0.33 $\pm$ 0.01( 5.00) & 0.30 $\pm$ 0.01( 9.00) & 0.30 $\pm$ 0.01( 21.00) & 0.28 $\pm$ 0.00( 41.00) \\
 Treeversity\#6 & \spaplus (ours) $S$=10 &0.47 $\pm$ 0.01( 1.10) & 0.37 $\pm$ 0.02( 1.30) & 0.34 $\pm$ 0.02( 1.50) & 0.33 $\pm$ 0.01( 2.00) & 0.30 $\pm$ 0.01( 3.00) & 0.30 $\pm$ 0.01( 6.00) & 0.28 $\pm$ 0.00( 11.00) \\
 Turkey & Baseline &0.40 $\pm$ 0.08( 1.00) & 0.27 $\pm$ 0.01( 3.00) & 0.38 $\pm$ 0.24( 5.00) & 0.24 $\pm$ 0.06( 10.00) &   -   &   -   &   -   \\
 Turkey & DivideMix &0.43 $\pm$ 0.08( 1.00) & 0.46 $\pm$ 0.04( 3.00) & 0.39 $\pm$ 0.06( 5.00) & 0.41 $\pm$ 0.07( 10.00) &   -   &   -   &   -   \\
 Turkey & Pseudo soft &0.37 $\pm$ 0.08( 1.00) & 0.28 $\pm$ 0.05( 3.00) & 0.22 $\pm$ 0.02( 5.00) & 0.21 $\pm$ 0.02( 10.00) &   -   &   -   &   -   \\
 Turkey & GT+CB &2.09 $\pm$ 2.55( 1.00) & 0.26 $\pm$ 0.04( 3.00) & 0.22 $\pm$ 0.01( 5.00) & 0.19 $\pm$ 0.01( 10.00) &   -   &   -   &   -   \\
 Turkey & \spaplus (ours) &0.27 $\pm$ 0.00( 2.00) & 0.25 $\pm$ 0.01( 4.00) & 0.23 $\pm$ 0.02( 6.00) & 0.22 $\pm$ 0.02( 11.00) & 0.28 $\pm$ 0.10( 21.00) & 0.20 $\pm$ 0.02( 51.00) & 0.19 $\pm$ 0.01( 101.00) \\
 Turkey & \spaplus (ours) $S$=2.5 &0.27 $\pm$ 0.00( 1.40) & 0.25 $\pm$ 0.01( 2.20) & 0.23 $\pm$ 0.02( 3.00) & 0.22 $\pm$ 0.02( 5.00) & 0.28 $\pm$ 0.10( 9.00) & 0.20 $\pm$ 0.02( 21.00) & 0.19 $\pm$ 0.01( 41.00) \\
 Turkey & \spaplus (ours) $S$=10 &0.27 $\pm$ 0.00( 1.10) & 0.25 $\pm$ 0.01( 1.30) & 0.23 $\pm$ 0.02( 1.50) & 0.22 $\pm$ 0.02( 2.00) & 0.28 $\pm$ 0.10( 3.00) & 0.20 $\pm$ 0.02( 6.00) & 0.19 $\pm$ 0.01( 11.00) \\

		\end{tabular}
	}
		
\end{table*}
}

\newcommand{\tblBenchmarkSemiSupervised}{
\begin{table*}[tb]
	\caption{
		Benchmark results  with 20, 50 and 100\% in. sup. without aggregation, metric: KL $\pm$ STD (budget), columns are the number of annotations used, $S$ describes the expected speedup, best viewed in digital form
	}
	\label{tbl:BenchmarkSemiSupervised}
	\resizebox{\linewidth}{!}{
		\begin{tabular}{l c | c c c c | c c c c c  | c c c c c c}
			\toprule

   in. sup. & & 20\% & & & & 50 \% & & & & & 100\% \\
			
		dataset & method & 1 & 3 & 5 & 10 & 1 & 2 & 3 & 5 & 10 & 1 & 3 & 5 & 10 \\
  \midrule 
Benthic & Baseline &1.55 $\pm$ 0.08( 0.20) &   -   & 1.09 $\pm$ 0.03( 1.00) &   -   & 1.34 $\pm$ 0.08( 0.50) & 1.08 $\pm$ 0.10( 1.00) &   -   &   -   &   -   & 1.17 $\pm$ 0.04( 1.00) & 0.81 $\pm$ 0.04( 3.00) & 0.76 $\pm$ 0.04( 5.00) & 0.75 $\pm$ 0.03( 10.00) \\
 Benthic & DivideMix &0.99 $\pm$ 0.07( 0.20) &   -   & 1.06 $\pm$ 0.08( 1.00) &   -   & 0.83 $\pm$ 0.03( 0.50) & 1.06 $\pm$ 0.03( 1.00) &   -   &   -   &   -   & 0.87 $\pm$ 0.07( 1.00) & 0.82 $\pm$ 0.00( 3.00) & 0.84 $\pm$ 0.02( 5.00) & 0.83 $\pm$ 0.05( 10.00) \\
 Benthic & Pseudo soft &1.08 $\pm$ 0.05( 0.20) &   -   & 0.81 $\pm$ 0.02( 1.00) &   -   & 1.01 $\pm$ 0.02( 0.50) & 0.83 $\pm$ 0.05( 1.00) &   -   &   -   &   -   & 1.00 $\pm$ 0.08( 1.00) & 0.76 $\pm$ 0.06( 3.00) & 0.71 $\pm$ 0.05( 5.00) & 0.70 $\pm$ 0.02( 10.00) \\
 Benthic & GT+CB &  -   &   -   &   -   &   -   &   -   &   -   &   -   &   -   &   -   & 0.81 $\pm$ 0.04( 1.00) & 0.68 $\pm$ 0.02( 3.00) & 0.67 $\pm$ 0.00( 5.00) & 0.65 $\pm$ 0.01( 10.00) \\
 Benthic & \spaplus (inc. un.) &1.00 $\pm$ 0.12( 1.20) & 0.77 $\pm$ 0.04( 3.20) & 0.74 $\pm$ 0.05( 5.20) & 0.69 $\pm$ 0.05( 10.20) &   -   &   -   &   -   &   -   &   -   &   -   &   -   &   -   &   -   \\
 Benthic & \spaplus (ours) &0.92 $\pm$ 0.06( 0.40) & 0.95 $\pm$ 0.06( 0.80) & 0.88 $\pm$ 0.05( 1.20) & 0.91 $\pm$ 0.05( 2.20) & 0.79 $\pm$ 0.02( 1.00) &   -   & 0.90 $\pm$ 0.21( 2.00) & 0.72 $\pm$ 0.03( 3.00) & 0.74 $\pm$ 0.03( 5.50) & 0.78 $\pm$ 0.03( 2.00) & 0.76 $\pm$ 0.06( 4.00) & 0.71 $\pm$ 0.05( 6.00) & 0.67 $\pm$ 0.02( 11.00) \\
 Benthic & \spaplus (inc. un.) $S$=2.5 &1.00 $\pm$ 0.12( 0.60) & 0.77 $\pm$ 0.04( 1.40) & 0.74 $\pm$ 0.05( 2.20) & 0.69 $\pm$ 0.05( 4.20) &   -   &   -   &   -   &   -   &   -   &   -   &   -   &   -   &   -   \\
 Benthic & \spaplus (inc. un.) $S$=10 &1.00 $\pm$ 0.12( 0.30) & 0.77 $\pm$ 0.04( 0.50) & 0.74 $\pm$ 0.05( 0.70) & 0.69 $\pm$ 0.05( 1.20) &   -   &   -   &   -   &   -   &   -   &   -   &   -   &   -   &   -   \\
 Benthic & \spaplus (ours) $S$=2.5 &0.92 $\pm$ 0.06( 0.28) & 0.95 $\pm$ 0.06( 0.44) & 0.88 $\pm$ 0.05( 0.60) & 0.91 $\pm$ 0.05( 1.00) & 0.79 $\pm$ 0.02( 0.70) &   -   & 0.90 $\pm$ 0.21( 1.10) & 0.72 $\pm$ 0.03( 1.50) & 0.74 $\pm$ 0.03( 2.50) & 0.78 $\pm$ 0.03( 1.40) & 0.76 $\pm$ 0.06( 2.20) & 0.71 $\pm$ 0.05( 3.00) & 0.67 $\pm$ 0.02( 5.00) \\
 Benthic & \spaplus (ours) $S$=10 &0.92 $\pm$ 0.06( 0.22) & 0.95 $\pm$ 0.06( 0.26) & 0.88 $\pm$ 0.05( 0.30) & 0.91 $\pm$ 0.05( 0.40) & 0.79 $\pm$ 0.02( 0.55) &   -   & 0.90 $\pm$ 0.21( 0.65) & 0.72 $\pm$ 0.03( 0.75) & 0.74 $\pm$ 0.03( 1.00) & 0.78 $\pm$ 0.03( 1.10) & 0.76 $\pm$ 0.06( 1.30) & 0.71 $\pm$ 0.05( 1.50) & 0.67 $\pm$ 0.02( 2.00) \\
 CIFAR10H & Baseline &0.63 $\pm$ 0.01( 0.20) &   -   & 0.55 $\pm$ 0.01( 1.00) &   -   & 0.47 $\pm$ 0.02( 0.50) & 0.38 $\pm$ 0.01( 1.00) &   -   &   -   &   -   & 0.41 $\pm$ 0.02( 1.00) & 0.30 $\pm$ 0.02( 3.00) & 0.27 $\pm$ 0.03( 5.00) & 0.26 $\pm$ 0.02( 10.00) \\
 CIFAR10H & DivideMix &0.67 $\pm$ 0.02( 0.20) &   -   & 0.65 $\pm$ 0.13( 1.00) &   -   & 0.44 $\pm$ 0.03( 0.50) & 0.36 $\pm$ 0.01( 1.00) &   -   &   -   &   -   & 0.36 $\pm$ 0.02( 1.00) & 0.35 $\pm$ 0.01( 3.00) & 0.35 $\pm$ 0.05( 5.00) & 0.37 $\pm$ 0.03( 10.00) \\
 CIFAR10H & Pseudo soft &1.12 $\pm$ 0.03( 0.20) &   -   & 1.10 $\pm$ 0.03( 1.00) &   -   & 0.69 $\pm$ 0.02( 0.50) & 0.71 $\pm$ 0.07( 1.00) &   -   &   -   &   -   & 0.41 $\pm$ 0.02( 1.00) & 0.41 $\pm$ 0.02( 3.00) & 0.42 $\pm$ 0.01( 5.00) & 0.43 $\pm$ 0.02( 10.00) \\
 CIFAR10H & GT+CB &  -   &   -   &   -   &   -   &   -   &   -   &   -   &   -   &   -   & 0.27 $\pm$ 0.01( 1.00) & 0.25 $\pm$ 0.01( 3.00) & 0.24 $\pm$ 0.02( 5.00) & 0.25 $\pm$ 0.01( 10.00) \\
 CIFAR10H & \spaplus (inc. un.) &0.29 $\pm$ 0.01( 1.20) & 0.26 $\pm$ 0.01( 3.20) & 0.25 $\pm$ 0.01( 5.20) & 0.24 $\pm$ 0.02( 10.20) &   -   &   -   &   -   &   -   &   -   &   -   &   -   &   -   &   -   \\
 CIFAR10H & \spaplus (ours) &0.55 $\pm$ 0.01( 0.40) & 0.56 $\pm$ 0.01( 0.80) & 0.56 $\pm$ 0.04( 1.20) & 0.56 $\pm$ 0.02( 2.20) & 0.34 $\pm$ 0.02( 1.00) &   -   & 0.35 $\pm$ 0.03( 2.00) & 0.32 $\pm$ 0.01( 3.00) & 0.35 $\pm$ 0.01( 5.50) & 0.25 $\pm$ 0.01( 2.00) & 0.26 $\pm$ 0.01( 4.00) & 0.24 $\pm$ 0.01( 6.00) & 0.24 $\pm$ 0.01( 11.00) \\
 CIFAR10H & \spaplus (inc. un.) $S$=2.5 &0.29 $\pm$ 0.01( 0.60) & 0.26 $\pm$ 0.01( 1.40) & 0.25 $\pm$ 0.01( 2.20) & 0.24 $\pm$ 0.02( 4.20) &   -   &   -   &   -   &   -   &   -   &   -   &   -   &   -   &   -   \\
 CIFAR10H & \spaplus (inc. un.) $S$=10 &0.29 $\pm$ 0.01( 0.30) & 0.26 $\pm$ 0.01( 0.50) & 0.25 $\pm$ 0.01( 0.70) & 0.24 $\pm$ 0.02( 1.20) &   -   &   -   &   -   &   -   &   -   &   -   &   -   &   -   &   -   \\
 CIFAR10H & \spaplus (ours) $S$=2.5 &0.55 $\pm$ 0.01( 0.28) & 0.56 $\pm$ 0.01( 0.44) & 0.56 $\pm$ 0.04( 0.60) & 0.56 $\pm$ 0.02( 1.00) & 0.34 $\pm$ 0.02( 0.70) &   -   & 0.35 $\pm$ 0.03( 1.10) & 0.32 $\pm$ 0.01( 1.50) & 0.35 $\pm$ 0.01( 2.50) & 0.25 $\pm$ 0.01( 1.40) & 0.26 $\pm$ 0.01( 2.20) & 0.24 $\pm$ 0.01( 3.00) & 0.24 $\pm$ 0.01( 5.00) \\
 CIFAR10H & \spaplus (ours) $S$=10 &0.55 $\pm$ 0.01( 0.22) & 0.56 $\pm$ 0.01( 0.26) & 0.56 $\pm$ 0.04( 0.30) & 0.56 $\pm$ 0.02( 0.40) & 0.34 $\pm$ 0.02( 0.55) &   -   & 0.35 $\pm$ 0.03( 0.65) & 0.32 $\pm$ 0.01( 0.75) & 0.35 $\pm$ 0.01( 1.00) & 0.25 $\pm$ 0.01( 1.10) & 0.26 $\pm$ 0.01( 1.30) & 0.24 $\pm$ 0.01( 1.50) & 0.24 $\pm$ 0.01( 2.00) \\
 MiceBone & Baseline &0.56 $\pm$ 0.06( 0.20) &   -   & 0.34 $\pm$ 0.03( 1.00) &   -   & 0.53 $\pm$ 0.07( 0.50) & 0.39 $\pm$ 0.01( 1.00) &   -   &   -   &   -   & 0.55 $\pm$ 0.06( 1.00) & 0.27 $\pm$ 0.03( 3.00) & 0.27 $\pm$ 0.02( 5.00) & 0.23 $\pm$ 0.02( 10.00) \\
 MiceBone & DivideMix &0.42 $\pm$ 0.06( 0.20) &   -   & 0.47 $\pm$ 0.10( 1.00) &   -   & 0.38 $\pm$ 0.04( 0.50) & 0.51 $\pm$ 0.12( 1.00) &   -   &   -   &   -   & 0.38 $\pm$ 0.05( 1.00) & 0.30 $\pm$ 0.04( 3.00) & 0.34 $\pm$ 0.02( 5.00) & 0.35 $\pm$ 0.05( 10.00) \\
 MiceBone & Pseudo soft &0.46 $\pm$ 0.05( 0.20) &   -   & 0.28 $\pm$ 0.03( 1.00) &   -   & 0.48 $\pm$ 0.07( 0.50) & 0.34 $\pm$ 0.07( 1.00) &   -   &   -   &   -   & 0.40 $\pm$ 0.08( 1.00) & 0.29 $\pm$ 0.03( 3.00) & 0.27 $\pm$ 0.04( 5.00) & 0.22 $\pm$ 0.02( 10.00) \\
 MiceBone & GT+CB &  -   &   -   &   -   &   -   &   -   &   -   &   -   &   -   &   -   & 0.35 $\pm$ 0.05( 1.00) & 0.24 $\pm$ 0.01( 3.00) & 0.25 $\pm$ 0.01( 5.00) & 0.23 $\pm$ 0.02( 10.00) \\
 MiceBone & \spaplus (inc. un.) &0.37 $\pm$ 0.05( 1.20) & 0.31 $\pm$ 0.03( 3.20) & 0.30 $\pm$ 0.02( 5.20) & 0.27 $\pm$ 0.02( 10.20) &   -   &   -   &   -   &   -   &   -   &   -   &   -   &   -   &   -   \\
 MiceBone & \spaplus (ours) &0.33 $\pm$ 0.03( 0.40) & 0.33 $\pm$ 0.03( 0.80) & 0.33 $\pm$ 0.05( 1.20) & 0.34 $\pm$ 0.05( 2.20) & 0.32 $\pm$ 0.06( 1.00) &   -   & 0.31 $\pm$ 0.03( 2.00) & 0.28 $\pm$ 0.03( 3.00) & 0.30 $\pm$ 0.04( 5.50) & 0.31 $\pm$ 0.03( 2.00) & 0.30 $\pm$ 0.00( 4.00) & 0.27 $\pm$ 0.01( 6.00) & 0.26 $\pm$ 0.04( 11.00) \\
 MiceBone & \spaplus (inc. un.) $S$=2.5 &0.37 $\pm$ 0.05( 0.60) & 0.31 $\pm$ 0.03( 1.40) & 0.30 $\pm$ 0.02( 2.20) & 0.27 $\pm$ 0.02( 4.20) &   -   &   -   &   -   &   -   &   -   &   -   &   -   &   -   &   -   \\
 MiceBone & \spaplus (inc. un.) $S$=10 &0.37 $\pm$ 0.05( 0.30) & 0.31 $\pm$ 0.03( 0.50) & 0.30 $\pm$ 0.02( 0.70) & 0.27 $\pm$ 0.02( 1.20) &   -   &   -   &   -   &   -   &   -   &   -   &   -   &   -   &   -   \\
 MiceBone & \spaplus (ours) $S$=2.5 &0.33 $\pm$ 0.03( 0.28) & 0.33 $\pm$ 0.03( 0.44) & 0.33 $\pm$ 0.05( 0.60) & 0.34 $\pm$ 0.05( 1.00) & 0.32 $\pm$ 0.06( 0.70) &   -   & 0.31 $\pm$ 0.03( 1.10) & 0.28 $\pm$ 0.03( 1.50) & 0.30 $\pm$ 0.04( 2.50) & 0.31 $\pm$ 0.03( 1.40) & 0.30 $\pm$ 0.00( 2.20) & 0.27 $\pm$ 0.01( 3.00) & 0.26 $\pm$ 0.04( 5.00) \\
 MiceBone & \spaplus (ours) $S$=10 &0.33 $\pm$ 0.03( 0.22) & 0.33 $\pm$ 0.03( 0.26) & 0.33 $\pm$ 0.05( 0.30) & 0.34 $\pm$ 0.05( 0.40) & 0.32 $\pm$ 0.06( 0.55) &   -   & 0.31 $\pm$ 0.03( 0.65) & 0.28 $\pm$ 0.03( 0.75) & 0.30 $\pm$ 0.04( 1.00) & 0.31 $\pm$ 0.03( 1.10) & 0.30 $\pm$ 0.00( 1.30) & 0.27 $\pm$ 0.01( 1.50) & 0.26 $\pm$ 0.04( 2.00) \\
 Pig & Baseline &0.79 $\pm$ 0.06( 0.20) &   -   & 0.61 $\pm$ 0.01( 1.00) &   -   & 0.79 $\pm$ 0.05( 0.50) & 0.62 $\pm$ 0.07( 1.00) &   -   &   -   &   -   & 0.75 $\pm$ 0.05( 1.00) & 0.56 $\pm$ 0.02( 3.00) & 0.62 $\pm$ 0.04( 5.00) & 0.54 $\pm$ 0.04( 10.00) \\
 Pig & DivideMix &0.84 $\pm$ 0.05( 0.20) &   -   & 0.86 $\pm$ 0.17( 1.00) &   -   & 0.94 $\pm$ 0.07( 0.50) & 1.11 $\pm$ 0.18( 1.00) &   -   &   -   &   -   & 0.95 $\pm$ 0.04( 1.00) & 1.11 $\pm$ 0.32( 3.00) & 1.36 $\pm$ 0.19( 5.00) & 1.04 $\pm$ 0.25( 10.00) \\
 Pig & Pseudo soft &0.74 $\pm$ 0.10( 0.20) &   -   & 0.61 $\pm$ 0.03( 1.00) &   -   & 0.68 $\pm$ 0.08( 0.50) & 0.67 $\pm$ 0.01( 1.00) &   -   &   -   &   -   & 0.70 $\pm$ 0.06( 1.00) & 0.63 $\pm$ 0.03( 3.00) & 0.59 $\pm$ 0.04( 5.00) & 0.61 $\pm$ 0.07( 10.00) \\
 Pig & GT+CB &  -   &   -   &   -   &   -   &   -   &   -   &   -   &   -   &   -   & 0.67 $\pm$ 0.08( 1.00) & 0.55 $\pm$ 0.01( 3.00) & 0.51 $\pm$ 0.01( 5.00) & 0.51 $\pm$ 0.04( 10.00) \\
 Pig & \spaplus (inc. un.) &0.72 $\pm$ 0.02( 1.20) & 0.55 $\pm$ 0.02( 3.20) & 0.55 $\pm$ 0.01( 5.20) & 0.54 $\pm$ 0.02( 10.20) &   -   &   -   &   -   &   -   &   -   &   -   &   -   &   -   &   -   \\
 Pig & \spaplus (ours) &0.68 $\pm$ 0.03( 0.40) & 0.62 $\pm$ 0.06( 0.80) & 0.66 $\pm$ 0.04( 1.20) & 0.69 $\pm$ 0.11( 2.20) & 0.60 $\pm$ 0.06( 1.00) &   -   & 0.62 $\pm$ 0.06( 2.00) & 0.58 $\pm$ 0.01( 3.00) & 0.57 $\pm$ 0.03( 5.50) & 0.61 $\pm$ 0.02( 2.00) & 0.54 $\pm$ 0.02( 4.00) & 0.62 $\pm$ 0.14( 6.00) & 0.51 $\pm$ 0.00( 11.00) \\
 Pig & \spaplus (inc. un.) $S$=2.5 &0.72 $\pm$ 0.02( 0.60) & 0.55 $\pm$ 0.02( 1.40) & 0.55 $\pm$ 0.01( 2.20) & 0.54 $\pm$ 0.02( 4.20) &   -   &   -   &   -   &   -   &   -   &   -   &   -   &   -   &   -   \\
 Pig & \spaplus (inc. un.) $S$=10 &0.72 $\pm$ 0.02( 0.30) & 0.55 $\pm$ 0.02( 0.50) & 0.55 $\pm$ 0.01( 0.70) & 0.54 $\pm$ 0.02( 1.20) &   -   &   -   &   -   &   -   &   -   &   -   &   -   &   -   &   -   \\
 Pig & \spaplus (ours) $S$=2.5 &0.68 $\pm$ 0.03( 0.28) & 0.62 $\pm$ 0.06( 0.44) & 0.66 $\pm$ 0.04( 0.60) & 0.69 $\pm$ 0.11( 1.00) & 0.60 $\pm$ 0.06( 0.70) &   -   & 0.62 $\pm$ 0.06( 1.10) & 0.58 $\pm$ 0.01( 1.50) & 0.57 $\pm$ 0.03( 2.50) & 0.61 $\pm$ 0.02( 1.40) & 0.54 $\pm$ 0.02( 2.20) & 0.62 $\pm$ 0.14( 3.00) & 0.51 $\pm$ 0.00( 5.00) \\
 Pig & \spaplus (ours) $S$=10 &0.68 $\pm$ 0.03( 0.22) & 0.62 $\pm$ 0.06( 0.26) & 0.66 $\pm$ 0.04( 0.30) & 0.69 $\pm$ 0.11( 0.40) & 0.60 $\pm$ 0.06( 0.55) &   -   & 0.62 $\pm$ 0.06( 0.65) & 0.58 $\pm$ 0.01( 0.75) & 0.57 $\pm$ 0.03( 1.00) & 0.61 $\pm$ 0.02( 1.10) & 0.54 $\pm$ 0.02( 1.30) & 0.62 $\pm$ 0.14( 1.50) & 0.51 $\pm$ 0.00( 2.00) \\
 Plankton & Baseline &0.43 $\pm$ 0.01( 0.20) &   -   & 0.34 $\pm$ 0.03( 1.00) &   -   & 0.36 $\pm$ 0.03( 0.50) & 0.28 $\pm$ 0.02( 1.00) &   -   &   -   &   -   & 0.34 $\pm$ 0.02( 1.00) & 0.23 $\pm$ 0.02( 3.00) & 0.22 $\pm$ 0.01( 5.00) & 0.20 $\pm$ 0.02( 10.00) \\
 Plankton & DivideMix &0.48 $\pm$ 0.03( 0.20) &   -   & 0.48 $\pm$ 0.02( 1.00) &   -   & 0.35 $\pm$ 0.02( 0.50) & 0.44 $\pm$ 0.03( 1.00) &   -   &   -   &   -   & 0.34 $\pm$ 0.01( 1.00) & 0.34 $\pm$ 0.01( 3.00) & 0.34 $\pm$ 0.01( 5.00) & 0.34 $\pm$ 0.01( 10.00) \\
 Plankton & Pseudo soft &0.41 $\pm$ 0.05( 0.20) &   -   & 0.32 $\pm$ 0.04( 1.00) &   -   & 0.44 $\pm$ 0.13( 0.50) & 0.46 $\pm$ 0.20( 1.00) &   -   &   -   &   -   & 0.32 $\pm$ 0.06( 1.00) & 0.33 $\pm$ 0.08( 3.00) & 0.24 $\pm$ 0.03( 5.00) & 0.25 $\pm$ 0.01( 10.00) \\
 Plankton & GT+CB &  -   &   -   &   -   &   -   &   -   &   -   &   -   &   -   &   -   & 0.25 $\pm$ 0.01( 1.00) & 0.21 $\pm$ 0.02( 3.00) & 0.20 $\pm$ 0.01( 5.00) & 0.20 $\pm$ 0.01( 10.00) \\
 Plankton & \spaplus (inc. un.) &0.37 $\pm$ 0.01( 1.20) & 0.30 $\pm$ 0.01( 3.20) & 0.30 $\pm$ 0.02( 5.20) & 0.28 $\pm$ 0.02( 10.20) &   -   &   -   &   -   &   -   &   -   &   -   &   -   &   -   &   -   \\
 Plankton & \spaplus (ours) &0.41 $\pm$ 0.02( 0.40) & 0.41 $\pm$ 0.02( 0.80) & 0.41 $\pm$ 0.02( 1.20) & 0.42 $\pm$ 0.03( 2.20) & 0.28 $\pm$ 0.02( 1.00) &   -   & 0.27 $\pm$ 0.02( 2.00) & 0.27 $\pm$ 0.02( 3.00) & 0.27 $\pm$ 0.02( 5.50) & 0.26 $\pm$ 0.01( 2.00) & 0.23 $\pm$ 0.01( 4.00) & 0.23 $\pm$ 0.01( 6.00) & 0.22 $\pm$ 0.01( 11.00) \\
 Plankton & \spaplus (inc. un.) $S$=2.5 &0.37 $\pm$ 0.01( 0.60) & 0.30 $\pm$ 0.01( 1.40) & 0.30 $\pm$ 0.02( 2.20) & 0.28 $\pm$ 0.02( 4.20) &   -   &   -   &   -   &   -   &   -   &   -   &   -   &   -   &   -   \\
 Plankton & \spaplus (inc. un.) $S$=10 &0.37 $\pm$ 0.01( 0.30) & 0.30 $\pm$ 0.01( 0.50) & 0.30 $\pm$ 0.02( 0.70) & 0.28 $\pm$ 0.02( 1.20) &   -   &   -   &   -   &   -   &   -   &   -   &   -   &   -   &   -   \\
 Plankton & \spaplus (ours) $S$=2.5 &0.41 $\pm$ 0.02( 0.28) & 0.41 $\pm$ 0.02( 0.44) & 0.41 $\pm$ 0.02( 0.60) & 0.42 $\pm$ 0.03( 1.00) & 0.28 $\pm$ 0.02( 0.70) &   -   & 0.27 $\pm$ 0.02( 1.10) & 0.27 $\pm$ 0.02( 1.50) & 0.27 $\pm$ 0.02( 2.50) & 0.26 $\pm$ 0.01( 1.40) & 0.23 $\pm$ 0.01( 2.20) & 0.23 $\pm$ 0.01( 3.00) & 0.22 $\pm$ 0.01( 5.00) \\
 Plankton & \spaplus (ours) $S$=10 &0.41 $\pm$ 0.02( 0.22) & 0.41 $\pm$ 0.02( 0.26) & 0.41 $\pm$ 0.02( 0.30) & 0.42 $\pm$ 0.03( 0.40) & 0.28 $\pm$ 0.02( 0.55) &   -   & 0.27 $\pm$ 0.02( 0.65) & 0.27 $\pm$ 0.02( 0.75) & 0.27 $\pm$ 0.02( 1.00) & 0.26 $\pm$ 0.01( 1.10) & 0.23 $\pm$ 0.01( 1.30) & 0.23 $\pm$ 0.01( 1.50) & 0.22 $\pm$ 0.01( 2.00) \\
 QualityMRI & Baseline &3.12 $\pm$ 1.42( 0.20) &   -   & 0.56 $\pm$ 0.25( 1.00) &   -   & 1.12 $\pm$ 0.36( 0.50) & 0.41 $\pm$ 0.31( 1.00) &   -   &   -   &   -   & 1.73 $\pm$ 0.48( 1.00) & 0.39 $\pm$ 0.19( 3.00) & 0.52 $\pm$ 0.28( 5.00) & 0.31 $\pm$ 0.14( 10.00) \\
 QualityMRI & DivideMix &0.24 $\pm$ 0.13( 0.20) &   -   & 0.26 $\pm$ 0.07( 1.00) &   -   & 0.45 $\pm$ 0.38( 0.50) & 0.91 $\pm$ 0.32( 1.00) &   -   &   -   &   -   & 0.46 $\pm$ 0.26( 1.00) & 0.60 $\pm$ 0.58( 3.00) & 0.17 $\pm$ 0.05( 5.00) & 1.34 $\pm$ 0.84( 10.00) \\
 QualityMRI & Pseudo soft &2.78 $\pm$ 0.67( 0.20) &   -   & 0.20 $\pm$ 0.04( 1.00) &   -   & 1.16 $\pm$ 0.34( 0.50) & 0.65 $\pm$ 0.38( 1.00) &   -   &   -   &   -   & 0.62 $\pm$ 0.16( 1.00) & 0.21 $\pm$ 0.10( 3.00) & 0.28 $\pm$ 0.03( 5.00) & 0.12 $\pm$ 0.04( 10.00) \\
 QualityMRI & GT+CB &  -   &   -   &   -   &   -   &   -   &   -   &   -   &   -   &   -   & 0.47 $\pm$ 0.10( 1.00) & 0.22 $\pm$ 0.17( 3.00) & 0.15 $\pm$ 0.01( 5.00) & 0.24 $\pm$ 0.13( 10.00) \\
 QualityMRI & \spaplus (inc. un.) &0.22 $\pm$ 0.14( 1.20) & 0.14 $\pm$ 0.08( 3.20) & 0.08 $\pm$ 0.00( 5.20) & 0.10 $\pm$ 0.03( 10.20) &   -   &   -   &   -   &   -   &   -   &   -   &   -   &   -   &   -   \\
 QualityMRI & \spaplus (ours) &0.24 $\pm$ 0.10( 0.40) & 0.16 $\pm$ 0.04( 0.80) & 0.14 $\pm$ 0.02( 1.20) & 0.17 $\pm$ 0.05( 2.20) & 0.38 $\pm$ 0.12( 1.00) &   -   & 0.24 $\pm$ 0.10( 2.00) & 0.16 $\pm$ 0.07( 3.00) & 0.38 $\pm$ 0.10( 5.50) & 0.24 $\pm$ 0.09( 2.00) & 0.12 $\pm$ 0.03( 4.00) & 0.11 $\pm$ 0.07( 6.00) & 0.11 $\pm$ 0.05( 11.00) \\
 QualityMRI & \spaplus (inc. un.) $S$=2.5 &0.22 $\pm$ 0.14( 0.60) & 0.14 $\pm$ 0.08( 1.40) & 0.08 $\pm$ 0.00( 2.20) & 0.10 $\pm$ 0.03( 4.20) &   -   &   -   &   -   &   -   &   -   &   -   &   -   &   -   &   -   \\
 QualityMRI & \spaplus (inc. un.) $S$=10 &0.22 $\pm$ 0.14( 0.30) & 0.14 $\pm$ 0.08( 0.50) & 0.08 $\pm$ 0.00( 0.70) & 0.10 $\pm$ 0.03( 1.20) &   -   &   -   &   -   &   -   &   -   &   -   &   -   &   -   &   -   \\
 QualityMRI & \spaplus (ours) $S$=2.5 &0.24 $\pm$ 0.10( 0.28) & 0.16 $\pm$ 0.04( 0.44) & 0.14 $\pm$ 0.02( 0.60) & 0.17 $\pm$ 0.05( 1.00) & 0.38 $\pm$ 0.12( 0.70) &   -   & 0.24 $\pm$ 0.10( 1.10) & 0.16 $\pm$ 0.07( 1.50) & 0.38 $\pm$ 0.10( 2.50) & 0.24 $\pm$ 0.09( 1.40) & 0.12 $\pm$ 0.03( 2.20) & 0.11 $\pm$ 0.07( 3.00) & 0.11 $\pm$ 0.05( 5.00) \\
 QualityMRI & \spaplus (ours) $S$=10 &0.24 $\pm$ 0.10( 0.22) & 0.16 $\pm$ 0.04( 0.26) & 0.14 $\pm$ 0.02( 0.30) & 0.17 $\pm$ 0.05( 0.40) & 0.38 $\pm$ 0.12( 0.55) &   -   & 0.24 $\pm$ 0.10( 0.65) & 0.16 $\pm$ 0.07( 0.75) & 0.38 $\pm$ 0.10( 1.00) & 0.24 $\pm$ 0.09( 1.10) & 0.12 $\pm$ 0.03( 1.30) & 0.11 $\pm$ 0.07( 1.50) & 0.11 $\pm$ 0.05( 2.00) \\
 Synthetic & Baseline &0.16 $\pm$ 0.02( 0.20) &   -   & 0.07 $\pm$ 0.01( 1.00) &   -   & 0.12 $\pm$ 0.02( 0.50) & 0.08 $\pm$ 0.01( 1.00) &   -   &   -   &   -   & 0.08 $\pm$ 0.01( 1.00) & 0.05 $\pm$ 0.00( 3.00) & 0.06 $\pm$ 0.01( 5.00) & 0.04 $\pm$ 0.00( 10.00) \\
 Synthetic & DivideMix &0.25 $\pm$ 0.02( 0.20) &   -   & 0.71 $\pm$ 0.10( 1.00) &   -   & 0.28 $\pm$ 0.02( 0.50) & 1.03 $\pm$ 0.07( 1.00) &   -   &   -   &   -   & 0.33 $\pm$ 0.00( 1.00) & 0.48 $\pm$ 0.06( 3.00) & 0.42 $\pm$ 0.07( 5.00) & 0.43 $\pm$ 0.04( 10.00) \\
 Synthetic & Pseudo soft &0.32 $\pm$ 0.06( 0.20) &   -   & 0.11 $\pm$ 0.03( 1.00) &   -   & 0.13 $\pm$ 0.02( 0.50) & 0.11 $\pm$ 0.02( 1.00) &   -   &   -   &   -   & 0.10 $\pm$ 0.01( 1.00) & 0.07 $\pm$ 0.00( 3.00) & 0.08 $\pm$ 0.01( 5.00) & 0.06 $\pm$ 0.01( 10.00) \\
 Synthetic & GT+CB &  -   &   -   &   -   &   -   &   -   &   -   &   -   &   -   &   -   & 0.10 $\pm$ 0.00( 1.00) & 0.08 $\pm$ 0.00( 3.00) & 0.07 $\pm$ 0.01( 5.00) & 0.07 $\pm$ 0.00( 10.00) \\
 Synthetic & \spaplus (inc. un.) &0.11 $\pm$ 0.02( 1.20) & 0.10 $\pm$ 0.01( 3.20) & 0.09 $\pm$ 0.00( 5.20) & 0.08 $\pm$ 0.01( 10.20) &   -   &   -   &   -   &   -   &   -   &   -   &   -   &   -   &   -   \\
 Synthetic & \spaplus (ours) &0.18 $\pm$ 0.01( 0.40) & 0.18 $\pm$ 0.01( 0.80) & 0.17 $\pm$ 0.01( 1.20) & 0.17 $\pm$ 0.01( 2.20) & 0.12 $\pm$ 0.01( 1.00) &   -   & 0.12 $\pm$ 0.01( 2.00) & 0.12 $\pm$ 0.01( 3.00) & 0.12 $\pm$ 0.00( 5.50) & 0.08 $\pm$ 0.01( 2.00) & 0.07 $\pm$ 0.01( 4.00) & 0.08 $\pm$ 0.00( 6.00) & 0.08 $\pm$ 0.00( 11.00) \\
 Synthetic & \spaplus (inc. un.) $S$=2.5 &0.11 $\pm$ 0.02( 0.60) & 0.10 $\pm$ 0.01( 1.40) & 0.09 $\pm$ 0.00( 2.20) & 0.08 $\pm$ 0.01( 4.20) &   -   &   -   &   -   &   -   &   -   &   -   &   -   &   -   &   -   \\
 Synthetic & \spaplus (inc. un.) $S$=10 &0.11 $\pm$ 0.02( 0.30) & 0.10 $\pm$ 0.01( 0.50) & 0.09 $\pm$ 0.00( 0.70) & 0.08 $\pm$ 0.01( 1.20) &   -   &   -   &   -   &   -   &   -   &   -   &   -   &   -   &   -   \\
 Synthetic & \spaplus (ours) $S$=2.5 &0.18 $\pm$ 0.01( 0.28) & 0.18 $\pm$ 0.01( 0.44) & 0.17 $\pm$ 0.01( 0.60) & 0.17 $\pm$ 0.01( 1.00) & 0.12 $\pm$ 0.01( 0.70) &   -   & 0.12 $\pm$ 0.01( 1.10) & 0.12 $\pm$ 0.01( 1.50) & 0.12 $\pm$ 0.00( 2.50) & 0.08 $\pm$ 0.01( 1.40) & 0.07 $\pm$ 0.01( 2.20) & 0.08 $\pm$ 0.00( 3.00) & 0.08 $\pm$ 0.00( 5.00) \\
 Synthetic & \spaplus (ours) $S$=10 &0.18 $\pm$ 0.01( 0.22) & 0.18 $\pm$ 0.01( 0.26) & 0.17 $\pm$ 0.01( 0.30) & 0.17 $\pm$ 0.01( 0.40) & 0.12 $\pm$ 0.01( 0.55) &   -   & 0.12 $\pm$ 0.01( 0.65) & 0.12 $\pm$ 0.01( 0.75) & 0.12 $\pm$ 0.00( 1.00) & 0.08 $\pm$ 0.01( 1.10) & 0.07 $\pm$ 0.01( 1.30) & 0.08 $\pm$ 0.00( 1.50) & 0.08 $\pm$ 0.00( 2.00) \\
 Treeversity\#1 & Baseline &0.55 $\pm$ 0.01( 0.20) &   -   & 0.51 $\pm$ 0.01( 1.00) &   -   & 0.46 $\pm$ 0.03( 0.50) & 0.43 $\pm$ 0.03( 1.00) &   -   &   -   &   -   & 0.49 $\pm$ 0.04( 1.00) & 0.37 $\pm$ 0.01( 3.00) & 0.37 $\pm$ 0.00( 5.00) & 0.36 $\pm$ 0.02( 10.00) \\
 Treeversity\#1 & DivideMix &0.53 $\pm$ 0.00( 0.20) &   -   & 0.63 $\pm$ 0.02( 1.00) &   -   & 0.47 $\pm$ 0.01( 0.50) & 0.60 $\pm$ 0.03( 1.00) &   -   &   -   &   -   & 0.47 $\pm$ 0.01( 1.00) & 0.46 $\pm$ 0.01( 3.00) & 0.46 $\pm$ 0.01( 5.00) & 0.46 $\pm$ 0.01( 10.00) \\
 Treeversity\#1 & Pseudo soft &0.70 $\pm$ 0.06( 0.20) &   -   & 0.54 $\pm$ 0.01( 1.00) &   -   & 0.53 $\pm$ 0.03( 0.50) & 0.50 $\pm$ 0.02( 1.00) &   -   &   -   &   -   & 0.46 $\pm$ 0.01( 1.00) & 0.41 $\pm$ 0.02( 3.00) & 0.39 $\pm$ 0.02( 5.00) & 0.37 $\pm$ 0.02( 10.00) \\
 Treeversity\#1 & GT+CB &  -   &   -   &   -   &   -   &   -   &   -   &   -   &   -   &   -   & 0.41 $\pm$ 0.01( 1.00) & 0.37 $\pm$ 0.02( 3.00) & 0.35 $\pm$ 0.02( 5.00) & 0.36 $\pm$ 0.02( 10.00) \\
 Treeversity\#1 & \spaplus (inc. un.) &0.44 $\pm$ 0.02( 1.20) & 0.38 $\pm$ 0.02( 3.20) & 0.37 $\pm$ 0.02( 5.20) & 0.38 $\pm$ 0.02( 10.20) &   -   &   -   &   -   &   -   &   -   &   -   &   -   &   -   &   -   \\
 Treeversity\#1 & \spaplus (ours) &0.49 $\pm$ 0.02( 0.40) & 0.50 $\pm$ 0.01( 0.80) & 0.50 $\pm$ 0.02( 1.20) & 0.50 $\pm$ 0.02( 2.20) & 0.42 $\pm$ 0.01( 1.00) &   -   & 0.41 $\pm$ 0.02( 2.00) & 0.41 $\pm$ 0.01( 3.00) & 0.40 $\pm$ 0.00( 5.50) & 0.39 $\pm$ 0.02( 2.00) & 0.37 $\pm$ 0.01( 4.00) & 0.37 $\pm$ 0.02( 6.00) & 0.35 $\pm$ 0.01( 11.00) \\
 Treeversity\#1 & \spaplus (inc. un.) $S$=2.5 &0.44 $\pm$ 0.02( 0.60) & 0.38 $\pm$ 0.02( 1.40) & 0.37 $\pm$ 0.02( 2.20) & 0.38 $\pm$ 0.02( 4.20) &   -   &   -   &   -   &   -   &   -   &   -   &   -   &   -   &   -   \\
 Treeversity\#1 & \spaplus (inc. un.) $S$=10 &0.44 $\pm$ 0.02( 0.30) & 0.38 $\pm$ 0.02( 0.50) & 0.37 $\pm$ 0.02( 0.70) & 0.38 $\pm$ 0.02( 1.20) &   -   &   -   &   -   &   -   &   -   &   -   &   -   &   -   &   -   \\
 Treeversity\#1 & \spaplus (ours) $S$=2.5 &0.49 $\pm$ 0.02( 0.28) & 0.50 $\pm$ 0.01( 0.44) & 0.50 $\pm$ 0.02( 0.60) & 0.50 $\pm$ 0.02( 1.00) & 0.42 $\pm$ 0.01( 0.70) &   -   & 0.41 $\pm$ 0.02( 1.10) & 0.41 $\pm$ 0.01( 1.50) & 0.40 $\pm$ 0.00( 2.50) & 0.39 $\pm$ 0.02( 1.40) & 0.37 $\pm$ 0.01( 2.20) & 0.37 $\pm$ 0.02( 3.00) & 0.35 $\pm$ 0.01( 5.00) \\
 Treeversity\#1 & \spaplus (ours) $S$=10 &0.49 $\pm$ 0.02( 0.22) & 0.50 $\pm$ 0.01( 0.26) & 0.50 $\pm$ 0.02( 0.30) & 0.50 $\pm$ 0.02( 0.40) & 0.42 $\pm$ 0.01( 0.55) &   -   & 0.41 $\pm$ 0.02( 0.65) & 0.41 $\pm$ 0.01( 0.75) & 0.40 $\pm$ 0.00( 1.00) & 0.39 $\pm$ 0.02( 1.10) & 0.37 $\pm$ 0.01( 1.30) & 0.37 $\pm$ 0.02( 1.50) & 0.35 $\pm$ 0.01( 2.00) \\
 Treeversity\#6 & Baseline &1.42 $\pm$ 0.14( 0.20) &   -   & 0.46 $\pm$ 0.01( 1.00) &   -   & 1.29 $\pm$ 0.09( 0.50) & 0.64 $\pm$ 0.01( 1.00) &   -   &   -   &   -   & 1.02 $\pm$ 0.03( 1.00) & 0.50 $\pm$ 0.02( 3.00) & 0.38 $\pm$ 0.00( 5.00) & 0.33 $\pm$ 0.01( 10.00) \\
 Treeversity\#6 & DivideMix &0.55 $\pm$ 0.05( 0.20) &   -   & 0.88 $\pm$ 0.04( 1.00) &   -   & 0.54 $\pm$ 0.05( 0.50) & 1.03 $\pm$ 0.09( 1.00) &   -   &   -   &   -   & 0.62 $\pm$ 0.05( 1.00) & 0.71 $\pm$ 0.02( 3.00) & 0.70 $\pm$ 0.04( 5.00) & 0.77 $\pm$ 0.03( 10.00) \\
 Treeversity\#6 & Pseudo soft &0.97 $\pm$ 0.10( 0.20) &   -   & 0.49 $\pm$ 0.04( 1.00) &   -   & 0.85 $\pm$ 0.04( 0.50) & 0.60 $\pm$ 0.02( 1.00) &   -   &   -   &   -   & 0.83 $\pm$ 0.13( 1.00) & 0.47 $\pm$ 0.02( 3.00) & 0.40 $\pm$ 0.02( 5.00) & 0.35 $\pm$ 0.01( 10.00) \\
 Treeversity\#6 & GT+CB &  -   &   -   &   -   &   -   &   -   &   -   &   -   &   -   &   -   & 0.61 $\pm$ 0.03( 1.00) & 0.37 $\pm$ 0.00( 3.00) & 0.33 $\pm$ 0.01( 5.00) & 0.31 $\pm$ 0.00( 10.00) \\
 Treeversity\#6 & \spaplus (inc. un.) &0.69 $\pm$ 0.04( 1.20) & 0.42 $\pm$ 0.01( 3.20) & 0.36 $\pm$ 0.01( 5.20) & 0.32 $\pm$ 0.00( 10.20) &   -   &   -   &   -   &   -   &   -   &   -   &   -   &   -   &   -   \\
 Treeversity\#6 & \spaplus (ours) &0.47 $\pm$ 0.00( 0.40) & 0.45 $\pm$ 0.00( 0.80) & 0.45 $\pm$ 0.01( 1.20) & 0.45 $\pm$ 0.02( 2.20) & 0.44 $\pm$ 0.01( 1.00) &   -   & 0.41 $\pm$ 0.02( 2.00) & 0.38 $\pm$ 0.02( 3.00) & 0.37 $\pm$ 0.01( 5.50) & 0.47 $\pm$ 0.01( 2.00) & 0.37 $\pm$ 0.02( 4.00) & 0.34 $\pm$ 0.02( 6.00) & 0.33 $\pm$ 0.01( 11.00) \\
 Treeversity\#6 & \spaplus (inc. un.) $S$=2.5 &0.69 $\pm$ 0.04( 0.60) & 0.42 $\pm$ 0.01( 1.40) & 0.36 $\pm$ 0.01( 2.20) & 0.32 $\pm$ 0.00( 4.20) &   -   &   -   &   -   &   -   &   -   &   -   &   -   &   -   &   -   \\
 Treeversity\#6 & \spaplus (inc. un.) $S$=10 &0.69 $\pm$ 0.04( 0.30) & 0.42 $\pm$ 0.01( 0.50) & 0.36 $\pm$ 0.01( 0.70) & 0.32 $\pm$ 0.00( 1.20) &   -   &   -   &   -   &   -   &   -   &   -   &   -   &   -   &   -   \\
 Treeversity\#6 & \spaplus (ours) $S$=2.5 &0.47 $\pm$ 0.00( 0.28) & 0.45 $\pm$ 0.00( 0.44) & 0.45 $\pm$ 0.01( 0.60) & 0.45 $\pm$ 0.02( 1.00) & 0.44 $\pm$ 0.01( 0.70) &   -   & 0.41 $\pm$ 0.02( 1.10) & 0.38 $\pm$ 0.02( 1.50) & 0.37 $\pm$ 0.01( 2.50) & 0.47 $\pm$ 0.01( 1.40) & 0.37 $\pm$ 0.02( 2.20) & 0.34 $\pm$ 0.02( 3.00) & 0.33 $\pm$ 0.01( 5.00) \\
 Treeversity\#6 & \spaplus (ours) $S$=10 &0.47 $\pm$ 0.00( 0.22) & 0.45 $\pm$ 0.00( 0.26) & 0.45 $\pm$ 0.01( 0.30) & 0.45 $\pm$ 0.02( 0.40) & 0.44 $\pm$ 0.01( 0.55) &   -   & 0.41 $\pm$ 0.02( 0.65) & 0.38 $\pm$ 0.02( 0.75) & 0.37 $\pm$ 0.01( 1.00) & 0.47 $\pm$ 0.01( 1.10) & 0.37 $\pm$ 0.02( 1.30) & 0.34 $\pm$ 0.02( 1.50) & 0.33 $\pm$ 0.01( 2.00) \\
 Turkey & Baseline &0.64 $\pm$ 0.16( 0.20) &   -   & 0.32 $\pm$ 0.09( 1.00) &   -   & 0.51 $\pm$ 0.08( 0.50) & 0.43 $\pm$ 0.08( 1.00) &   -   &   -   &   -   & 0.40 $\pm$ 0.08( 1.00) & 0.27 $\pm$ 0.01( 3.00) & 0.38 $\pm$ 0.24( 5.00) & 0.24 $\pm$ 0.06( 10.00) \\
 Turkey & DivideMix &0.52 $\pm$ 0.06( 0.20) &   -   & 0.61 $\pm$ 0.11( 1.00) &   -   & 0.46 $\pm$ 0.07( 0.50) & 0.57 $\pm$ 0.13( 1.00) &   -   &   -   &   -   & 0.43 $\pm$ 0.08( 1.00) & 0.46 $\pm$ 0.04( 3.00) & 0.39 $\pm$ 0.06( 5.00) & 0.41 $\pm$ 0.07( 10.00) \\
 Turkey & Pseudo soft &0.45 $\pm$ 0.07( 0.20) &   -   & 0.27 $\pm$ 0.01( 1.00) &   -   & 0.43 $\pm$ 0.09( 0.50) & 0.30 $\pm$ 0.04( 1.00) &   -   &   -   &   -   & 0.37 $\pm$ 0.08( 1.00) & 0.28 $\pm$ 0.05( 3.00) & 0.22 $\pm$ 0.02( 5.00) & 0.21 $\pm$ 0.02( 10.00) \\
 Turkey & GT+CB &  -   &   -   &   -   &   -   &   -   &   -   &   -   &   -   &   -   & 2.09 $\pm$ 2.55( 1.00) & 0.26 $\pm$ 0.04( 3.00) & 0.22 $\pm$ 0.01( 5.00) & 0.19 $\pm$ 0.01( 10.00) \\
 Turkey & \spaplus (inc. un.) &0.42 $\pm$ 0.08( 1.20) & 0.30 $\pm$ 0.11( 3.20) & 0.23 $\pm$ 0.01( 5.20) & 0.22 $\pm$ 0.02( 10.20) &   -   &   -   &   -   &   -   &   -   &   -   &   -   &   -   &   -   \\
 Turkey & \spaplus (ours) &0.44 $\pm$ 0.07( 0.40) & 0.42 $\pm$ 0.08( 0.80) & 0.41 $\pm$ 0.07( 1.20) & 0.43 $\pm$ 0.03( 2.20) & 0.36 $\pm$ 0.09( 1.00) &   -   & 0.30 $\pm$ 0.06( 2.00) & 0.29 $\pm$ 0.04( 3.00) & 0.35 $\pm$ 0.07( 5.50) & 0.27 $\pm$ 0.00( 2.00) & 0.25 $\pm$ 0.01( 4.00) & 0.23 $\pm$ 0.02( 6.00) & 0.22 $\pm$ 0.02( 11.00) \\
 Turkey & \spaplus (inc. un.) $S$=2.5 &0.42 $\pm$ 0.08( 0.60) & 0.30 $\pm$ 0.11( 1.40) & 0.23 $\pm$ 0.01( 2.20) & 0.22 $\pm$ 0.02( 4.20) &   -   &   -   &   -   &   -   &   -   &   -   &   -   &   -   &   -   \\
 Turkey & \spaplus (inc. un.) $S$=10 &0.42 $\pm$ 0.08( 0.30) & 0.30 $\pm$ 0.11( 0.50) & 0.23 $\pm$ 0.01( 0.70) & 0.22 $\pm$ 0.02( 1.20) &   -   &   -   &   -   &   -   &   -   &   -   &   -   &   -   &   -   \\
 Turkey & \spaplus (ours) $S$=2.5 &0.44 $\pm$ 0.07( 0.28) & 0.42 $\pm$ 0.08( 0.44) & 0.41 $\pm$ 0.07( 0.60) & 0.43 $\pm$ 0.03( 1.00) & 0.36 $\pm$ 0.09( 0.70) &   -   & 0.30 $\pm$ 0.06( 1.10) & 0.29 $\pm$ 0.04( 1.50) & 0.35 $\pm$ 0.07( 2.50) & 0.27 $\pm$ 0.00( 1.40) & 0.25 $\pm$ 0.01( 2.20) & 0.23 $\pm$ 0.02( 3.00) & 0.22 $\pm$ 0.02( 5.00) \\
 Turkey & \spaplus (ours) $S$=10 &0.44 $\pm$ 0.07( 0.22) & 0.42 $\pm$ 0.08( 0.26) & 0.41 $\pm$ 0.07( 0.30) & 0.43 $\pm$ 0.03( 0.40) & 0.36 $\pm$ 0.09( 0.55) &   -   & 0.30 $\pm$ 0.06( 0.65) & 0.29 $\pm$ 0.04( 0.75) & 0.35 $\pm$ 0.07( 1.00) & 0.27 $\pm$ 0.00( 1.10) & 0.25 $\pm$ 0.01( 1.30) & 0.23 $\pm$ 0.02( 1.50) & 0.22 $\pm$ 0.02( 2.00) \\

		\end{tabular}
	}
		
\end{table*}
}

\newcommand{\tblBenchmarkSemiSupervisedMedian}{
\begin{table*}[tb]
	\caption{
		Benchmark results  with 20, 50 and 100\% in. sup. with median aggregation, metric: KL $\pm$ STD (budget), columns are the number of annotations used, $S$ describes the expected speedup, best viewed in digital form
	}
	\label{tbl:BenchmarkSemiSupervisedMedian}
	\resizebox{\linewidth}{!}{
		\begin{tabular}{l | c c c c | c c c c c  | c c c c c c}
			\toprule

  in. sup. & 20\% & & & & 50 \% & & & & & 100\% \\
			
		 method & 1 & 3 & 5 & 10 & 1 & 2 & 3 & 5 & 10 & 1 & 3 & 5 & 10 \\
  \midrule 
 Baseline &0.63 $\pm$ 0.06( 0.20) &   -   & 0.49 $\pm$ 0.02( 1.00) &   -   & 0.52 $\pm$ 0.06( 0.50) & 0.42 $\pm$ 0.03( 1.00) &   -   &   -   &   -   & 0.52 $\pm$ 0.04( 1.00) & 0.34 $\pm$ 0.02( 3.00) & 0.37 $\pm$ 0.03( 5.00) & 0.29 $\pm$ 0.02( 10.00) \\
 DivideMix &0.53 $\pm$ 0.05( 0.20) &   -   & 0.64 $\pm$ 0.09( 1.00) &   -   & 0.45 $\pm$ 0.04( 0.50) & 0.75 $\pm$ 0.08( 1.00) &   -   &   -   &   -   & 0.44 $\pm$ 0.05( 1.00) & 0.47 $\pm$ 0.03( 3.00) & 0.41 $\pm$ 0.04( 5.00) & 0.45 $\pm$ 0.04( 10.00) \\
 Pseudo soft &0.72 $\pm$ 0.06( 0.20) &   -   & 0.40 $\pm$ 0.03( 1.00) &   -   & 0.61 $\pm$ 0.06( 0.50) & 0.55 $\pm$ 0.05( 1.00) &   -   &   -   &   -   & 0.43 $\pm$ 0.07( 1.00) & 0.37 $\pm$ 0.03( 3.00) & 0.34 $\pm$ 0.02( 5.00) & 0.30 $\pm$ 0.02( 10.00) \\
 GT+CB &  -   &   -   &   -   &   -   &   -   &   -   &   -   &   -   &   -   & 0.44 $\pm$ 0.03( 1.00) & 0.26 $\pm$ 0.01( 3.00) & 0.25 $\pm$ 0.01( 5.00) & 0.24 $\pm$ 0.01( 10.00) \\
 \spaplus (inc. un.) &0.40 $\pm$ 0.03( 1.20) & 0.31 $\pm$ 0.02( 3.20) & 0.30 $\pm$ 0.01( 5.20) & 0.28 $\pm$ 0.02( 10.20) &   -   &   -   &   -   &   -   &   -   &   -   &   -   &   -   &   -   \\
 \spaplus (ours) &0.46 $\pm$ 0.02( 0.40) & 0.44 $\pm$ 0.03( 0.80) & 0.43 $\pm$ 0.03( 1.20) & 0.44 $\pm$ 0.03( 2.20) & 0.37 $\pm$ 0.02( 1.00) &   -   & 0.33 $\pm$ 0.03( 2.00) & 0.31 $\pm$ 0.02( 3.00) & 0.36 $\pm$ 0.02( 5.50) & 0.29 $\pm$ 0.02( 2.00) & 0.28 $\pm$ 0.01( 4.00) & 0.26 $\pm$ 0.02( 6.00) & 0.25 $\pm$ 0.01( 11.00) \\
 \spaplus (inc. un.) $S$=2.5 &0.40 $\pm$ 0.03( 0.60) & 0.31 $\pm$ 0.02( 1.40) & 0.30 $\pm$ 0.01( 2.20) & 0.28 $\pm$ 0.02( 4.20) &   -   &   -   &   -   &   -   &   -   &   -   &   -   &   -   &   -   \\
 \spaplus (inc. un.) $S$=10 &0.40 $\pm$ 0.03( 0.30) & 0.31 $\pm$ 0.02( 0.50) & 0.30 $\pm$ 0.01( 0.70) & 0.28 $\pm$ 0.02( 1.20) &   -   &   -   &   -   &   -   &   -   &   -   &   -   &   -   &   -   \\
 \spaplus (ours) $S$=2.5 &0.46 $\pm$ 0.02( 0.28) & 0.44 $\pm$ 0.03( 0.44) & 0.43 $\pm$ 0.03( 0.60) & 0.44 $\pm$ 0.03( 1.00) & 0.37 $\pm$ 0.02( 0.70) &   -   & 0.33 $\pm$ 0.03( 1.10) & 0.31 $\pm$ 0.02( 1.50) & 0.36 $\pm$ 0.02( 2.50) & 0.29 $\pm$ 0.02( 1.40) & 0.28 $\pm$ 0.01( 2.20) & 0.26 $\pm$ 0.02( 3.00) & 0.25 $\pm$ 0.01( 5.00) \\
 \spaplus (ours) $S$=10 &0.46 $\pm$ 0.02( 0.22) & 0.44 $\pm$ 0.03( 0.26) & 0.43 $\pm$ 0.03( 0.30) & 0.44 $\pm$ 0.03( 0.40) & 0.37 $\pm$ 0.02( 0.55) &   -   & 0.33 $\pm$ 0.03( 0.65) & 0.31 $\pm$ 0.02( 0.75) & 0.36 $\pm$ 0.02( 1.00) & 0.29 $\pm$ 0.02( 1.10) & 0.28 $\pm$ 0.01( 1.30) & 0.26 $\pm$ 0.02( 1.50) & 0.25 $\pm$ 0.01( 2.00) \\

		\end{tabular}
	}
		
\end{table*}
}

\newcommand{\tblBenchmarkSemiSupervisedMean}{
\begin{table*}[tb]
	\caption{
		Benchmark results  with 20, 50 and 100\% in. sup. with mean aggregation, metric: KL $\pm$ STD (budget), columns are the number of annotations used, $S$ describes the expected speedup
	}
	\label{tbl:BenchmarkSemiSupervisedMean}
	\resizebox{\linewidth}{!}{
		\begin{tabular}{l | c c c c | c c c c c  | c c c c c c}
			\toprule

  in. sup. & 20\% & & & & 50 \% & & & & & 100\% \\
			
method & 1 & 3 & 5 & 10 & 1 & 2 & 3 & 5 & 10 & 1 & 3 & 5 & 10 \\
 Baseline &0.98 $\pm$ 0.20( 0.20) &   -   & 0.49 $\pm$ 0.05( 1.00) &   -   & 0.70 $\pm$ 0.08( 0.50) & 0.47 $\pm$ 0.07( 1.00) &   -   &   -   &   -   & 0.69 $\pm$ 0.08( 1.00) & 0.37 $\pm$ 0.04( 3.00) & 0.39 $\pm$ 0.07( 5.00) & 0.33 $\pm$ 0.04( 10.00) \\
 DivideMix &0.55 $\pm$ 0.05( 0.20) &   -   & 0.66 $\pm$ 0.08( 1.00) &   -   & 0.51 $\pm$ 0.07( 0.50) & 0.76 $\pm$ 0.10( 1.00) &   -   &   -   &   -   & 0.52 $\pm$ 0.06( 1.00) & 0.56 $\pm$ 0.11( 3.00) & 0.54 $\pm$ 0.05( 5.00) & 0.64 $\pm$ 0.14( 10.00) \\
 Pseudo soft &0.90 $\pm$ 0.12( 0.20) &   -   & 0.47 $\pm$ 0.03( 1.00) &   -   & 0.64 $\pm$ 0.08( 0.50) & 0.52 $\pm$ 0.09( 1.00) &   -   &   -   &   -   & 0.52 $\pm$ 0.07( 1.00) & 0.39 $\pm$ 0.04( 3.00) & 0.36 $\pm$ 0.03( 5.00) & 0.33 $\pm$ 0.02( 10.00) \\
 GT+CB &  -   &   -   &   -   &   -   &   -   &   -   &   -   &   -   &   -   & 0.60 $\pm$ 0.29( 1.00) & 0.32 $\pm$ 0.03( 3.00) & 0.30 $\pm$ 0.01( 5.00) & 0.30 $\pm$ 0.02( 10.00) \\
 \spaplus (inc. un.) &0.46 $\pm$ 0.05( 1.20) & 0.35 $\pm$ 0.03( 3.20) & 0.33 $\pm$ 0.02( 5.20) & 0.31 $\pm$ 0.02( 10.20) &   -   &   -   &   -   &   -   &   -   &   -   &   -   &   -   &   -   \\
 \spaplus (ours) &0.47 $\pm$ 0.03( 0.40) & 0.46 $\pm$ 0.03( 0.80) & 0.45 $\pm$ 0.03( 1.20) & 0.46 $\pm$ 0.04( 2.20) & 0.40 $\pm$ 0.04( 1.00) &   -   & 0.39 $\pm$ 0.06( 2.00) & 0.35 $\pm$ 0.02( 3.00) & 0.38 $\pm$ 0.03( 5.50) & 0.37 $\pm$ 0.02( 2.00) & 0.33 $\pm$ 0.02( 4.00) & 0.32 $\pm$ 0.03( 6.00) & 0.30 $\pm$ 0.02( 11.00) \\
 \spaplus (inc. un.) $S$=2.5 &0.46 $\pm$ 0.05( 0.60) & 0.35 $\pm$ 0.03( 1.40) & 0.33 $\pm$ 0.02( 2.20) & 0.31 $\pm$ 0.02( 4.20) &   -   &   -   &   -   &   -   &   -   &   -   &   -   &   -   &   -   \\
 \spaplus (inc. un.) $S$=10 &0.46 $\pm$ 0.05( 0.30) & 0.35 $\pm$ 0.03( 0.50) & 0.33 $\pm$ 0.02( 0.70) & 0.31 $\pm$ 0.02( 1.20) &   -   &   -   &   -   &   -   &   -   &   -   &   -   &   -   &   -   \\
 \spaplus (ours) $S$=2.5 &0.47 $\pm$ 0.03( 0.28) & 0.46 $\pm$ 0.03( 0.44) & 0.45 $\pm$ 0.03( 0.60) & 0.46 $\pm$ 0.04( 1.00) & 0.40 $\pm$ 0.04( 0.70) &   -   & 0.39 $\pm$ 0.06( 1.10) & 0.35 $\pm$ 0.02( 1.50) & 0.38 $\pm$ 0.03( 2.50) & 0.37 $\pm$ 0.02( 1.40) & 0.33 $\pm$ 0.02( 2.20) & 0.32 $\pm$ 0.03( 3.00) & 0.30 $\pm$ 0.02( 5.00) \\
 \spaplus (ours) $S$=10 &0.47 $\pm$ 0.03( 0.22) & 0.46 $\pm$ 0.03( 0.26) & 0.45 $\pm$ 0.03( 0.30) & 0.46 $\pm$ 0.04( 0.40) & 0.40 $\pm$ 0.04( 0.55) &   -   & 0.39 $\pm$ 0.06( 0.65) & 0.35 $\pm$ 0.02( 0.75) & 0.38 $\pm$ 0.03( 1.00) & 0.37 $\pm$ 0.02( 1.10) & 0.33 $\pm$ 0.02( 1.30) & 0.32 $\pm$ 0.03( 1.50) & 0.30 $\pm$ 0.02( 2.00) \\

		\end{tabular}
	}
		
\end{table*}
}

\newcommand{\tblBenchmarkAblationFirstPart}{
\begin{table*}[tb]
	\caption{
		Benchmark results ablation study,  metric: $KL$ $\pm$ STD, columns are the number of annotations used
	}
	\label{tbl:BenchmarkAblationFirstPart}
	\resizebox{\linewidth}{!}{
		\begin{tabular}{l c c c c c c c c c c c c c c c c}
			\toprule
			
		dataset & method &  1 & 3 & 5 & 10 & 20 & 50 & 100 \\

  \midrule 
  
 Benthic & \spaplus (ours) &0.78 $\pm$ 0.03 & 0.76 $\pm$ 0.06 & 0.71 $\pm$ 0.05 & 0.67 $\pm$ 0.02 & 0.68 $\pm$ 0.02 & 0.67 $\pm$ 0.06 & 0.65 $\pm$ 0.03 \\
 Benthic & \spaplus (+ GT $\delta$) &0.80 $\pm$ 0.03 & 0.78 $\pm$ 0.02 & 0.72 $\pm$ 0.07 & 0.67 $\pm$ 0.04 & 0.69 $\pm$ 0.03 & 0.67 $\pm$ 0.03 & 0.65 $\pm$ 0.02 \\
 Benthic & Only CB &0.79 $\pm$ 0.05 & 0.75 $\pm$ 0.03 & 0.68 $\pm$ 0.01 & 0.70 $\pm$ 0.02 & 0.65 $\pm$ 0.03 & 0.67 $\pm$ 0.03 & 0.69 $\pm$ 0.02 \\
 Benthic & Only CB ($\mu$=0.25) &0.82 $\pm$ 0.05 & 0.76 $\pm$ 0.01 & 0.74 $\pm$ 0.03 & 0.72 $\pm$ 0.03 &   -   &   -   &   -   \\
 Benthic & Only BC &  -   & 0.97 $\pm$ 0.20 & 0.76 $\pm$ 0.05 & 0.71 $\pm$ 0.02 &   -   &   -   &   -   \\
 CIFAR10H & \spaplus (ours) &0.25 $\pm$ 0.01 & 0.26 $\pm$ 0.01 & 0.24 $\pm$ 0.01 & 0.24 $\pm$ 0.01 & 0.24 $\pm$ 0.01 & 0.25 $\pm$ 0.02 & 0.24 $\pm$ 0.01 \\
 CIFAR10H & \spaplus (+ GT $\delta$) &0.26 $\pm$ 0.01 & 0.25 $\pm$ 0.00 & 0.25 $\pm$ 0.01 & 0.25 $\pm$ 0.02 & 0.24 $\pm$ 0.01 & 0.23 $\pm$ 0.01 & 0.24 $\pm$ 0.01 \\
 CIFAR10H & Only CB &0.25 $\pm$ 0.02 & 0.25 $\pm$ 0.01 & 0.24 $\pm$ 0.01 & 0.25 $\pm$ 0.01 & 0.24 $\pm$ 0.01 & 0.25 $\pm$ 0.01 & 0.25 $\pm$ 0.01 \\
 CIFAR10H & Only CB ($\mu$=0.25) &0.30 $\pm$ 0.02 & 0.28 $\pm$ 0.01 & 0.28 $\pm$ 0.01 & 0.28 $\pm$ 0.01 &   -   &   -   &   -   \\
 CIFAR10H & Only BC &  -   & 0.28 $\pm$ 0.01 & 0.27 $\pm$ 0.02 & 0.26 $\pm$ 0.02 &   -   &   -   &   -   \\
 MiceBone & \spaplus (ours) &0.31 $\pm$ 0.03 & 0.30 $\pm$ 0.00 & 0.27 $\pm$ 0.01 & 0.26 $\pm$ 0.04 & 0.25 $\pm$ 0.04 & 0.25 $\pm$ 0.02 & 0.23 $\pm$ 0.03 \\
 MiceBone & \spaplus (+ GT $\delta$) &0.31 $\pm$ 0.02 & 0.29 $\pm$ 0.01 & 0.27 $\pm$ 0.05 & 0.25 $\pm$ 0.01 & 0.24 $\pm$ 0.02 & 0.23 $\pm$ 0.02 & 0.23 $\pm$ 0.02 \\
 MiceBone & Only CB &0.36 $\pm$ 0.04 & 0.28 $\pm$ 0.04 & 0.27 $\pm$ 0.02 & 0.25 $\pm$ 0.05 & 0.25 $\pm$ 0.02 & 0.24 $\pm$ 0.02 & 0.25 $\pm$ 0.04 \\
 MiceBone & Only CB ($\mu$=0.25) &0.34 $\pm$ 0.04 & 0.29 $\pm$ 0.01 & 0.30 $\pm$ 0.03 & 0.30 $\pm$ 0.01 &   -   &   -   &   -   \\
 MiceBone & Only BC &  -   & 0.32 $\pm$ 0.04 & 0.32 $\pm$ 0.02 & 0.26 $\pm$ 0.02 &   -   &   -   &   -   \\
 Pig & \spaplus (ours) &0.61 $\pm$ 0.02 & 0.54 $\pm$ 0.02 & 0.62 $\pm$ 0.14 & 0.51 $\pm$ 0.00 & 0.58 $\pm$ 0.08 & 0.50 $\pm$ 0.02 & 0.50 $\pm$ 0.02 \\
 Pig & \spaplus (+ GT $\delta$) &0.57 $\pm$ 0.01 & 0.53 $\pm$ 0.03 & 0.61 $\pm$ 0.07 & 0.55 $\pm$ 0.04 & 0.50 $\pm$ 0.01 & 0.52 $\pm$ 0.01 & 0.51 $\pm$ 0.01 \\
 Pig & Only CB &0.58 $\pm$ 0.05 & 0.53 $\pm$ 0.01 & 0.56 $\pm$ 0.02 & 0.60 $\pm$ 0.06 & 0.51 $\pm$ 0.02 & 0.50 $\pm$ 0.01 & 0.52 $\pm$ 0.04 \\
 Pig & Only CB ($\mu$=0.25) &0.55 $\pm$ 0.01 & 0.52 $\pm$ 0.02 & 0.53 $\pm$ 0.02 & 0.55 $\pm$ 0.04 &   -   &   -   &   -   \\
 Pig & Only BC &  -   & 0.59 $\pm$ 0.03 & 0.57 $\pm$ 0.02 & 0.52 $\pm$ 0.01 &   -   &   -   &   -   \\
 Plankton & \spaplus (ours) &0.26 $\pm$ 0.01 & 0.23 $\pm$ 0.01 & 0.23 $\pm$ 0.01 & 0.22 $\pm$ 0.01 & 0.22 $\pm$ 0.01 & 0.22 $\pm$ 0.01 & 0.22 $\pm$ 0.01 \\
 Plankton & \spaplus (+ GT $\delta$) &0.26 $\pm$ 0.02 & 0.27 $\pm$ 0.01 & 0.27 $\pm$ 0.01 & 0.23 $\pm$ 0.01 & 0.21 $\pm$ 0.01 & 0.20 $\pm$ 0.02 & 0.20 $\pm$ 0.01 \\
 Plankton & Only CB &0.26 $\pm$ 0.01 & 0.23 $\pm$ 0.01 & 0.23 $\pm$ 0.01 & 0.22 $\pm$ 0.01 & 0.22 $\pm$ 0.01 & 0.22 $\pm$ 0.01 & 0.22 $\pm$ 0.01 \\
 Plankton & Only CB ($\mu$=0.25) &0.28 $\pm$ 0.02 & 0.25 $\pm$ 0.01 & 0.26 $\pm$ 0.02 & 0.26 $\pm$ 0.01 &   -   &   -   &   -   \\
 Plankton & Only BC &  -   & 0.28 $\pm$ 0.02 & 0.25 $\pm$ 0.01 & 0.24 $\pm$ 0.01 &   -   &   -   &   -   \\
 QualityMRI & \spaplus (ours) &0.24 $\pm$ 0.09 & 0.12 $\pm$ 0.03 & 0.11 $\pm$ 0.07 & 0.11 $\pm$ 0.05 & 0.10 $\pm$ 0.04 & 0.12 $\pm$ 0.04 & 0.16 $\pm$ 0.04 \\
 QualityMRI & \spaplus (+ GT $\delta$) &0.18 $\pm$ 0.10 & 0.18 $\pm$ 0.13 & 0.24 $\pm$ 0.08 & 0.11 $\pm$ 0.02 & 0.20 $\pm$ 0.05 & 0.24 $\pm$ 0.03 & 0.14 $\pm$ 0.05 \\
 QualityMRI & Only CB &0.20 $\pm$ 0.10 & 0.22 $\pm$ 0.02 & 0.15 $\pm$ 0.04 & 0.18 $\pm$ 0.10 & 0.15 $\pm$ 0.02 & 0.15 $\pm$ 0.09 & 0.16 $\pm$ 0.07 \\
 QualityMRI & Only CB ($\mu$=0.25) &0.11 $\pm$ 0.03 & 0.14 $\pm$ 0.05 & 0.07 $\pm$ 0.01 & 0.11 $\pm$ 0.05 &   -   &   -   &   -   \\
 QualityMRI & Only BC &  -   & 0.27 $\pm$ 0.18 & 0.27 $\pm$ 0.18 & 0.18 $\pm$ 0.14 &   -   &   -   &   -   \\
 Synthetic & \spaplus (ours) &0.08 $\pm$ 0.01 & 0.07 $\pm$ 0.01 & 0.08 $\pm$ 0.00 & 0.08 $\pm$ 0.00 & 0.08 $\pm$ 0.00 & 0.07 $\pm$ 0.01 & 0.07 $\pm$ 0.00 \\
 Synthetic & \spaplus (+ GT $\delta$) &0.08 $\pm$ 0.01 & 0.08 $\pm$ 0.01 & 0.08 $\pm$ 0.01 & 0.08 $\pm$ 0.00 & 0.08 $\pm$ 0.00 & 0.07 $\pm$ 0.01 & 0.08 $\pm$ 0.01 \\
 Synthetic & Only CB &0.08 $\pm$ 0.00 & 0.09 $\pm$ 0.01 & 0.08 $\pm$ 0.01 & 0.08 $\pm$ 0.00 & 0.08 $\pm$ 0.01 & 0.09 $\pm$ 0.02 & 0.09 $\pm$ 0.01 \\
 Synthetic & Only CB ($\mu$=0.25) &0.17 $\pm$ 0.01 & 0.16 $\pm$ 0.01 & 0.17 $\pm$ 0.01 & 0.18 $\pm$ 0.01 &   -   &   -   &   -   \\
 Synthetic & Only BC &  -   & 0.06 $\pm$ 0.00 & 0.05 $\pm$ 0.00 & 0.04 $\pm$ 0.00 &   -   &   -   &   -   \\
 Treeversity\#1 & \spaplus (ours) &0.39 $\pm$ 0.02 & 0.37 $\pm$ 0.01 & 0.37 $\pm$ 0.02 & 0.35 $\pm$ 0.01 & 0.35 $\pm$ 0.01 & 0.35 $\pm$ 0.02 & 0.35 $\pm$ 0.01 \\
 Treeversity\#1 & \spaplus (+ GT $\delta$) &0.39 $\pm$ 0.01 & 0.37 $\pm$ 0.01 & 0.37 $\pm$ 0.02 & 0.35 $\pm$ 0.01 & 0.36 $\pm$ 0.00 & 0.35 $\pm$ 0.01 & 0.36 $\pm$ 0.02 \\
 Treeversity\#1 & Only CB &0.40 $\pm$ 0.00 & 0.37 $\pm$ 0.01 & 0.36 $\pm$ 0.01 & 0.37 $\pm$ 0.02 & 0.36 $\pm$ 0.01 & 0.36 $\pm$ 0.01 & 0.36 $\pm$ 0.01 \\
 Treeversity\#1 & Only CB ($\mu$=0.25) &0.43 $\pm$ 0.01 & 0.41 $\pm$ 0.01 & 0.41 $\pm$ 0.01 & 0.40 $\pm$ 0.01 &   -   &   -   &   -   \\
 Treeversity\#1 & Only BC &  -   & 0.37 $\pm$ 0.02 & 0.37 $\pm$ 0.01 & 0.36 $\pm$ 0.01 &   -   &   -   &   -   \\
 Treeversity\#6 & \spaplus (ours) &0.47 $\pm$ 0.01 & 0.37 $\pm$ 0.02 & 0.34 $\pm$ 0.02 & 0.33 $\pm$ 0.01 & 0.30 $\pm$ 0.01 & 0.30 $\pm$ 0.01 & 0.28 $\pm$ 0.00 \\
 Treeversity\#6 & \spaplus (+ GT $\delta$) &0.47 $\pm$ 0.02 & 0.40 $\pm$ 0.01 & 0.35 $\pm$ 0.00 & 0.33 $\pm$ 0.01 & 0.31 $\pm$ 0.02 & 0.30 $\pm$ 0.00 & 0.30 $\pm$ 0.01 \\
 Treeversity\#6 & Only CB &0.47 $\pm$ 0.02 & 0.37 $\pm$ 0.00 & 0.36 $\pm$ 0.01 & 0.32 $\pm$ 0.00 & 0.31 $\pm$ 0.01 & 0.30 $\pm$ 0.00 & 0.30 $\pm$ 0.01 \\
 Treeversity\#6 & Only CB ($\mu$=0.25) &0.48 $\pm$ 0.01 & 0.45 $\pm$ 0.01 & 0.42 $\pm$ 0.01 & 0.41 $\pm$ 0.01 &   -   &   -   &   -   \\
 Treeversity\#6 & Only BC &  -   & 0.46 $\pm$ 0.03 & 0.42 $\pm$ 0.01 & 0.33 $\pm$ 0.02 &   -   &   -   &   -   \\
 Turkey & \spaplus (ours) &0.27 $\pm$ 0.00 & 0.25 $\pm$ 0.01 & 0.23 $\pm$ 0.02 & 0.22 $\pm$ 0.02 & 0.28 $\pm$ 0.10 & 0.20 $\pm$ 0.02 & 0.19 $\pm$ 0.01 \\
 Turkey & \spaplus (+ GT $\delta$) &0.29 $\pm$ 0.04 & 0.24 $\pm$ 0.01 & 0.20 $\pm$ 0.01 & 0.22 $\pm$ 0.03 & 0.20 $\pm$ 0.02 & 0.23 $\pm$ 0.05 & 0.19 $\pm$ 0.01 \\
 Turkey & Only CB &0.33 $\pm$ 0.07 & 0.29 $\pm$ 0.07 & 0.26 $\pm$ 0.06 & 0.22 $\pm$ 0.02 & 0.20 $\pm$ 0.01 & 0.21 $\pm$ 0.03 & 0.20 $\pm$ 0.02 \\
 Turkey & Only CB ($\mu$=0.25) &0.32 $\pm$ 0.02 & 0.26 $\pm$ 0.03 & 0.35 $\pm$ 0.12 & 0.24 $\pm$ 0.01 &   -   &   -   &   -   \\
 Turkey & Only BC &  -   & 0.24 $\pm$ 0.01 & 0.23 $\pm$ 0.05 & 0.23 $\pm$ 0.01 &   -   &   -   &   -   \\

		\end{tabular}
	}
		
\end{table*}
}

\newcommand{\tblBenchmarkAblationSecondPart}{
\begin{table*}[tb]
	\caption{
		Benchmark results ablation study,   metric: $\hat{KL}$ $\pm$ STD, columns are the number of annotations used
	}
	\label{tbl:BenchmarkAblationSecondPart}
	\resizebox{\linewidth}{!}{
		\begin{tabular}{l c c c c c c c c c c c c c c c c}
			\toprule
			
		dataset & method & 1 & 3 & 5 & 10 & 20 & 50 & 100 \\

  \midrule

Benthic & \spaplus (ours) &1.05 $\pm$ 0.03 & 0.51 $\pm$ 0.01 & 0.31 $\pm$ 0.01 & 0.20 $\pm$ 0.00 & 0.16 $\pm$ 0.00 & 0.13 $\pm$ 0.00 & 0.12 $\pm$ 0.00 \\
 Benthic & \spaplus (+ GT $\delta$) &1.05 $\pm$ 0.03 & 0.58 $\pm$ 0.00 & 0.39 $\pm$ 0.01 & 0.20 $\pm$ 0.00 & 0.11 $\pm$ 0.01 & 0.07 $\pm$ 0.00 & 0.05 $\pm$ 0.00 \\
 Benthic & Only CB &1.05 $\pm$ 0.03 & 0.52 $\pm$ 0.01 & 0.32 $\pm$ 0.01 & 0.22 $\pm$ 0.00 & 0.18 $\pm$ 0.00 & 0.16 $\pm$ 0.00 & 0.15 $\pm$ 0.00 \\
 Benthic & Only CB ($\mu$=0.25) &0.88 $\pm$ 0.03 & 0.55 $\pm$ 0.01 & 0.41 $\pm$ 0.01 & 0.35 $\pm$ 0.01 &   -   &   -   &   -   \\
 Benthic & Only BC &  -   & 1.43 $\pm$ 0.04 & 0.88 $\pm$ 0.03 & 0.35 $\pm$ 0.02 &   -   &   -   &   -   \\
 CIFAR10H & \spaplus (ours) &0.52 $\pm$ 0.03 & 0.17 $\pm$ 0.01 & 0.13 $\pm$ 0.00 & 0.10 $\pm$ 0.00 & 0.07 $\pm$ 0.00 & 0.04 $\pm$ 0.00 & 0.03 $\pm$ 0.00 \\
 CIFAR10H & \spaplus (+ GT $\delta$) &0.52 $\pm$ 0.03 & 0.16 $\pm$ 0.00 & 0.13 $\pm$ 0.00 & 0.09 $\pm$ 0.00 & 0.06 $\pm$ 0.00 & 0.03 $\pm$ 0.00 & 0.02 $\pm$ 0.00 \\
 CIFAR10H & Only CB &0.52 $\pm$ 0.03 & 0.17 $\pm$ 0.01 & 0.13 $\pm$ 0.00 & 0.09 $\pm$ 0.00 & 0.06 $\pm$ 0.00 & 0.03 $\pm$ 0.00 & 0.02 $\pm$ 0.00 \\
 CIFAR10H & Only CB ($\mu$=0.25) &0.49 $\pm$ 0.03 & 0.17 $\pm$ 0.01 & 0.13 $\pm$ 0.00 & 0.11 $\pm$ 0.00 &   -   &   -   &   -   \\
 CIFAR10H & Only BC &  -   & 0.29 $\pm$ 0.01 & 0.22 $\pm$ 0.00 & 0.15 $\pm$ 0.00 &   -   &   -   &   -   \\
 MiceBone & \spaplus (ours) &0.61 $\pm$ 0.02 & 0.28 $\pm$ 0.00 & 0.21 $\pm$ 0.01 & 0.14 $\pm$ 0.00 & 0.11 $\pm$ 0.01 & 0.09 $\pm$ 0.00 & 0.09 $\pm$ 0.01 \\
 MiceBone & \spaplus (+ GT $\delta$) &0.61 $\pm$ 0.02 & 0.34 $\pm$ 0.01 & 0.24 $\pm$ 0.02 & 0.13 $\pm$ 0.01 & 0.08 $\pm$ 0.00 & 0.05 $\pm$ 0.00 & 0.04 $\pm$ 0.00 \\
 MiceBone & Only CB &0.61 $\pm$ 0.02 & 0.28 $\pm$ 0.00 & 0.21 $\pm$ 0.01 & 0.15 $\pm$ 0.01 & 0.13 $\pm$ 0.01 & 0.11 $\pm$ 0.01 & 0.11 $\pm$ 0.01 \\
 MiceBone & Only CB ($\mu$=0.25) &0.40 $\pm$ 0.02 & 0.26 $\pm$ 0.01 & 0.24 $\pm$ 0.01 & 0.22 $\pm$ 0.01 &   -   &   -   &   -   \\
 MiceBone & Only BC &  -   & 1.39 $\pm$ 0.07 & 0.91 $\pm$ 0.04 & 0.40 $\pm$ 0.02 &   -   &   -   &   -   \\
 Pig & \spaplus (ours) &0.97 $\pm$ 0.03 & 0.44 $\pm$ 0.01 & 0.30 $\pm$ 0.01 & 0.19 $\pm$ 0.01 & 0.13 $\pm$ 0.00 & 0.10 $\pm$ 0.00 & 0.09 $\pm$ 0.00 \\
 Pig & \spaplus (+ GT $\delta$) &0.97 $\pm$ 0.03 & 0.48 $\pm$ 0.02 & 0.33 $\pm$ 0.00 & 0.19 $\pm$ 0.01 & 0.12 $\pm$ 0.00 & 0.07 $\pm$ 0.00 & 0.06 $\pm$ 0.00 \\
 Pig & Only CB &0.97 $\pm$ 0.03 & 0.44 $\pm$ 0.01 & 0.31 $\pm$ 0.01 & 0.20 $\pm$ 0.01 & 0.16 $\pm$ 0.01 & 0.13 $\pm$ 0.01 & 0.12 $\pm$ 0.00 \\
 Pig & Only CB ($\mu$=0.25) &0.57 $\pm$ 0.01 & 0.38 $\pm$ 0.00 & 0.33 $\pm$ 0.01 & 0.29 $\pm$ 0.01 &   -   &   -   &   -   \\
 Pig & Only BC &  -   & 2.02 $\pm$ 0.05 & 1.38 $\pm$ 0.02 & 0.56 $\pm$ 0.02 &   -   &   -   &   -   \\
 Plankton & \spaplus (ours) &0.59 $\pm$ 0.02 & 0.35 $\pm$ 0.03 & 0.29 $\pm$ 0.01 & 0.22 $\pm$ 0.01 & 0.16 $\pm$ 0.01 & 0.13 $\pm$ 0.00 & 0.12 $\pm$ 0.00 \\
 Plankton & \spaplus (+ GT $\delta$) &0.59 $\pm$ 0.02 & 0.50 $\pm$ 0.01 & 0.38 $\pm$ 0.02 & 0.21 $\pm$ 0.01 & 0.12 $\pm$ 0.01 & 0.06 $\pm$ 0.00 & 0.04 $\pm$ 0.00 \\
 Plankton & Only CB &0.59 $\pm$ 0.02 & 0.35 $\pm$ 0.03 & 0.29 $\pm$ 0.01 & 0.22 $\pm$ 0.01 & 0.16 $\pm$ 0.01 & 0.13 $\pm$ 0.00 & 0.12 $\pm$ 0.00 \\
 Plankton & Only CB ($\mu$=0.25) &0.54 $\pm$ 0.02 & 0.35 $\pm$ 0.03 & 0.32 $\pm$ 0.01 & 0.28 $\pm$ 0.01 &   -   &   -   &   -   \\
 Plankton & Only BC &  -   & 0.79 $\pm$ 0.06 & 0.68 $\pm$ 0.05 & 0.36 $\pm$ 0.02 &   -   &   -   &   -   \\
 QualityMRI & \spaplus (ours) &0.54 $\pm$ 0.06 & 0.16 $\pm$ 0.02 & 0.09 $\pm$ 0.02 & 0.05 $\pm$ 0.01 & 0.03 $\pm$ 0.01 & 0.02 $\pm$ 0.00 & 0.01 $\pm$ 0.00 \\
 QualityMRI & \spaplus (+ GT $\delta$) &0.54 $\pm$ 0.06 & 0.14 $\pm$ 0.02 & 0.10 $\pm$ 0.01 & 0.05 $\pm$ 0.00 & 0.02 $\pm$ 0.01 & 0.01 $\pm$ 0.00 & 0.01 $\pm$ 0.00 \\
 QualityMRI & Only CB &0.54 $\pm$ 0.06 & 0.15 $\pm$ 0.02 & 0.07 $\pm$ 0.02 & 0.04 $\pm$ 0.01 & 0.02 $\pm$ 0.01 & 0.01 $\pm$ 0.00 & 0.01 $\pm$ 0.00 \\
 QualityMRI & Only CB ($\mu$=0.25) &0.21 $\pm$ 0.03 & 0.09 $\pm$ 0.01 & 0.06 $\pm$ 0.01 & 0.06 $\pm$ 0.01 &   -   &   -   &   -   \\
 QualityMRI & Only BC &  -   & 1.06 $\pm$ 0.18 & 0.42 $\pm$ 0.10 & 0.10 $\pm$ 0.02 &   -   &   -   &   -   \\
 Synthetic & \spaplus (ours) &0.90 $\pm$ 0.01 & 0.44 $\pm$ 0.00 & 0.32 $\pm$ 0.00 & 0.19 $\pm$ 0.00 & 0.12 $\pm$ 0.00 & 0.08 $\pm$ 0.00 & 0.07 $\pm$ 0.00 \\
 Synthetic & \spaplus (+ GT $\delta$) &0.90 $\pm$ 0.01 & 0.49 $\pm$ 0.01 & 0.38 $\pm$ 0.00 & 0.21 $\pm$ 0.01 & 0.13 $\pm$ 0.00 & 0.07 $\pm$ 0.00 & 0.06 $\pm$ 0.00 \\
 Synthetic & Only CB &0.90 $\pm$ 0.01 & 0.42 $\pm$ 0.00 & 0.30 $\pm$ 0.01 & 0.19 $\pm$ 0.00 & 0.13 $\pm$ 0.00 & 0.09 $\pm$ 0.00 & 0.08 $\pm$ 0.00 \\
 Synthetic & Only CB ($\mu$=0.25) &0.64 $\pm$ 0.01 & 0.39 $\pm$ 0.00 & 0.34 $\pm$ 0.01 & 0.27 $\pm$ 0.00 &   -   &   -   &   -   \\
 Synthetic & Only BC &  -   & 1.52 $\pm$ 0.01 & 1.16 $\pm$ 0.01 & 0.54 $\pm$ 0.01 &   -   &   -   &   -   \\
 Treeversity\#1 & \spaplus (ours) &0.76 $\pm$ 0.04 & 0.27 $\pm$ 0.02 & 0.20 $\pm$ 0.01 & 0.12 $\pm$ 0.01 & 0.08 $\pm$ 0.00 & 0.06 $\pm$ 0.00 & 0.05 $\pm$ 0.00 \\
 Treeversity\#1 & \spaplus (+ GT $\delta$) &0.76 $\pm$ 0.04 & 0.27 $\pm$ 0.02 & 0.20 $\pm$ 0.01 & 0.12 $\pm$ 0.01 & 0.08 $\pm$ 0.00 & 0.05 $\pm$ 0.00 & 0.04 $\pm$ 0.00 \\
 Treeversity\#1 & Only CB &0.76 $\pm$ 0.04 & 0.27 $\pm$ 0.02 & 0.20 $\pm$ 0.01 & 0.12 $\pm$ 0.01 & 0.09 $\pm$ 0.00 & 0.07 $\pm$ 0.00 & 0.06 $\pm$ 0.00 \\
 Treeversity\#1 & Only CB ($\mu$=0.25) &0.64 $\pm$ 0.03 & 0.30 $\pm$ 0.01 & 0.25 $\pm$ 0.00 & 0.20 $\pm$ 0.00 &   -   &   -   &   -   \\
 Treeversity\#1 & Only BC &  -   & 0.68 $\pm$ 0.04 & 0.50 $\pm$ 0.02 & 0.23 $\pm$ 0.01 &   -   &   -   &   -   \\
 Treeversity\#6 & \spaplus (ours) &1.11 $\pm$ 0.03 & 0.46 $\pm$ 0.01 & 0.30 $\pm$ 0.01 & 0.18 $\pm$ 0.01 & 0.11 $\pm$ 0.00 & 0.07 $\pm$ 0.00 & 0.06 $\pm$ 0.00 \\
 Treeversity\#6 & \spaplus (+ GT $\delta$) &1.11 $\pm$ 0.03 & 0.50 $\pm$ 0.01 & 0.36 $\pm$ 0.01 & 0.20 $\pm$ 0.00 & 0.12 $\pm$ 0.00 & 0.07 $\pm$ 0.00 & 0.06 $\pm$ 0.00 \\
 Treeversity\#6 & Only CB &1.11 $\pm$ 0.03 & 0.45 $\pm$ 0.01 & 0.29 $\pm$ 0.00 & 0.17 $\pm$ 0.00 & 0.12 $\pm$ 0.00 & 0.09 $\pm$ 0.00 & 0.08 $\pm$ 0.00 \\
 Treeversity\#6 & Only CB ($\mu$=0.25) &0.74 $\pm$ 0.02 & 0.42 $\pm$ 0.01 & 0.33 $\pm$ 0.01 & 0.27 $\pm$ 0.01 &   -   &   -   &   -   \\
 Treeversity\#6 & Only BC &  -   & 1.85 $\pm$ 0.01 & 1.38 $\pm$ 0.03 & 0.61 $\pm$ 0.02 &   -   &   -   &   -   \\
 Turkey & \spaplus (ours) &0.38 $\pm$ 0.03 & 0.14 $\pm$ 0.01 & 0.10 $\pm$ 0.01 & 0.06 $\pm$ 0.00 & 0.04 $\pm$ 0.00 & 0.03 $\pm$ 0.00 & 0.02 $\pm$ 0.00 \\
 Turkey & \spaplus (+ GT $\delta$) &0.38 $\pm$ 0.03 & 0.14 $\pm$ 0.01 & 0.10 $\pm$ 0.01 & 0.06 $\pm$ 0.00 & 0.04 $\pm$ 0.00 & 0.03 $\pm$ 0.00 & 0.02 $\pm$ 0.00 \\
 Turkey & Only CB &0.38 $\pm$ 0.03 & 0.14 $\pm$ 0.01 & 0.10 $\pm$ 0.01 & 0.06 $\pm$ 0.01 & 0.04 $\pm$ 0.00 & 0.03 $\pm$ 0.00 & 0.03 $\pm$ 0.00 \\
 Turkey & Only CB ($\mu$=0.25) &0.27 $\pm$ 0.02 & 0.14 $\pm$ 0.01 & 0.12 $\pm$ 0.01 & 0.10 $\pm$ 0.01 &   -   &   -   &   -   \\
 Turkey & Only BC &  -   & 0.43 $\pm$ 0.03 & 0.28 $\pm$ 0.02 & 0.13 $\pm$ 0.01 &   -   &   -   &   -   \\

		\end{tabular}
	}
		
\end{table*}
}

\newcommand{\tblBenchmarkAblationThirdPart}{
\begin{table*}[tb]
	\caption{
		Benchmark results ablation study,   metric: $KL$ $\pm$ STD but evaluated on a transformer backbone, columns are the number of annotations used
	}
	\label{tbl:BenchmarkAblationThirdPart}
	\resizebox{\linewidth}{!}{
		\begin{tabular}{l c c c c c c c c c c c c c c c c}
			\toprule
			
		dataset & method & 1 & 3 & 5 & 10 & 20 & 50 & 100 \\

\midrule

Benthic & \spaplus (+ GT $\delta$) &  -   & 0.50 $\pm$ 0.02 &   -   & 0.46 $\pm$ 0.01 &   -   &   -   & 0.43 $\pm$ 0.01 \\
 Benthic & Only CB &  -   & 0.47 $\pm$ 0.00 &   -   & 0.46 $\pm$ 0.02 &   -   &   -   & 0.45 $\pm$ 0.01 \\
 CIFAR10H & \spaplus (+ GT $\delta$) &  -   & 0.10 $\pm$ 0.00 &   -   & 0.10 $\pm$ 0.00 &   -   &   -   & 0.10 $\pm$ 0.00 \\
 CIFAR10H & Only CB &  -   & 0.10 $\pm$ 0.00 &   -   & 0.10 $\pm$ 0.00 &   -   &   -   & 0.10 $\pm$ 0.00 \\
 MiceBone & \spaplus (+ GT $\delta$) &  -   & 0.22 $\pm$ 0.02 &   -   & 0.19 $\pm$ 0.02 &   -   &   -   & 0.17 $\pm$ 0.02 \\
 MiceBone & Only CB &  -   & 0.21 $\pm$ 0.02 &   -   & 0.19 $\pm$ 0.01 &   -   &   -   & 0.19 $\pm$ 0.01 \\
 Pig & \spaplus (+ GT $\delta$) &  -   & 0.45 $\pm$ 0.01 &   -   & 0.38 $\pm$ 0.01 &   -   &   -   & 0.34 $\pm$ 0.01 \\
 Pig & Only CB &  -   & 0.44 $\pm$ 0.01 &   -   & 0.39 $\pm$ 0.01 &   -   &   -   & 0.36 $\pm$ 0.01 \\
 Plankton & \spaplus (+ GT $\delta$) &  -   & 0.22 $\pm$ 0.01 &   -   & 0.19 $\pm$ 0.01 &   -   &   -   & 0.16 $\pm$ 0.01 \\
 Plankton & Only CB &  -   & 0.19 $\pm$ 0.01 &   -   & 0.18 $\pm$ 0.01 &   -   &   -   & 0.18 $\pm$ 0.01 \\
 QualityMRI & \spaplus (+ GT $\delta$) &  -   & 0.05 $\pm$ 0.00 &   -   & 0.04 $\pm$ 0.00 &   -   &   -   & 0.05 $\pm$ 0.00 \\
 QualityMRI & Only CB &  -   & 0.06 $\pm$ 0.00 &   -   & 0.05 $\pm$ 0.01 &   -   &   -   & 0.05 $\pm$ 0.01 \\
 Synthetic & \spaplus (+ GT $\delta$) &  -   & 0.07 $\pm$ 0.00 &   -   & 0.06 $\pm$ 0.00 &   -   &   -   & 0.04 $\pm$ 0.00 \\
 Synthetic & Only CB &  -   & 0.08 $\pm$ 0.00 &   -   & 0.07 $\pm$ 0.00 &   -   &   -   & 0.07 $\pm$ 0.00 \\
 Treeversity\#1 & \spaplus (+ GT $\delta$) &  -   & 0.24 $\pm$ 0.01 &   -   & 0.21 $\pm$ 0.01 &   -   &   -   & 0.20 $\pm$ 0.01 \\
 Treeversity\#1 & Only CB &  -   & 0.24 $\pm$ 0.01 &   -   & 0.21 $\pm$ 0.01 &   -   &   -   & 0.21 $\pm$ 0.01 \\
 Treeversity\#6 & \spaplus (+ GT $\delta$) &  -   & 0.24 $\pm$ 0.01 &   -   & 0.18 $\pm$ 0.01 &   -   &   -   & 0.15 $\pm$ 0.00 \\
 Treeversity\#6 & Only CB &  -   & 0.24 $\pm$ 0.01 &   -   & 0.18 $\pm$ 0.01 &   -   &   -   & 0.16 $\pm$ 0.01 \\
 Turkey & \spaplus (+ GT $\delta$) &  -   & 0.18 $\pm$ 0.02 &   -   & 0.16 $\pm$ 0.01 &   -   &   -   & 0.16 $\pm$ 0.01 \\
 Turkey & Only CB &  -   & 0.18 $\pm$ 0.00 &   -   & 0.16 $\pm$ 0.01 &   -   &   -   & 0.16 $\pm$ 0.01 \\

		\end{tabular}
	}
		
\end{table*}
}

\newcommand{\tblBenchmarkAblationMedian}{
\begin{table}[tb]
	\caption{
		Benchmark results ablation study, first block of rows $KL$ result on normal benchmark, second block of rows $\hat{KL}$ on benchmark, last block of rows $KL$ results on benchmark but with ViT as backbone, all results are median aggregations across the datasets
	}
	\label{tbl:BenchmarkAblationMedian}
 \centering
	\resizebox{\linewidth}{!}{
		\begin{tabular}{l c c c c c c c c c c c c c c c c c c c c c}
			\toprule
			
		method &&  1 && 3 && 5 && 10 && 20 && 50 && 100 \\
  \midrule
\spaplus (ours) &&0.29 $\pm$ 0.02 && 0.28 $\pm$ 0.01 && 0.26 $\pm$ 0.02 && 0.25 $\pm$ 0.01 && 0.27 $\pm$ 0.02 && 0.25 $\pm$ 0.02 && 0.24 $\pm$ 0.01 \\
 \spaplus (+ GT $\delta$) &&0.30 $\pm$ 0.02 && 0.28 $\pm$ 0.01 && 0.27 $\pm$ 0.01 && 0.25 $\pm$ 0.02 && 0.24 $\pm$ 0.01 && 0.24 $\pm$ 0.01 && 0.24 $\pm$ 0.01 \\
 Only CB &&0.34 $\pm$ 0.03 && 0.28 $\pm$ 0.01 && 0.27 $\pm$ 0.01 && 0.25 $\pm$ 0.02 && 0.25 $\pm$ 0.01 && 0.25 $\pm$ 0.02 && 0.25 $\pm$ 0.02 \\
 Only CB ($\mu$=0.25) &&0.33 $\pm$ 0.02 && 0.28 $\pm$ 0.01 && 0.33 $\pm$ 0.02 && 0.29 $\pm$ 0.01 &&   -   &&   -   &&   -   \\
 Only BC &&  -   && 0.30 $\pm$ 0.02 && 0.29 $\pm$ 0.02 && 0.26 $\pm$ 0.02 &&   -   &&   -   &&   -   \\

\midrule

\spaplus (ours) &&0.68 $\pm$ 0.03 && 0.32 $\pm$ 0.01 && 0.25 $\pm$ 0.01 && 0.16 $\pm$ 0.00 && 0.11 $\pm$ 0.00 && 0.08 $\pm$ 0.00 && 0.07 $\pm$ 0.00 \\
 \spaplus (+ GT $\delta$) &&0.68 $\pm$ 0.03 && 0.41 $\pm$ 0.01 && 0.29 $\pm$ 0.01 && 0.16 $\pm$ 0.00 && 0.10 $\pm$ 0.00 && 0.05 $\pm$ 0.00 && 0.04 $\pm$ 0.00 \\
 Only CB &&0.68 $\pm$ 0.03 && 0.32 $\pm$ 0.01 && 0.25 $\pm$ 0.01 && 0.16 $\pm$ 0.01 && 0.12 $\pm$ 0.00 && 0.09 $\pm$ 0.00 && 0.08 $\pm$ 0.00 \\
 Only CB ($\mu$=0.25) &&0.55 $\pm$ 0.02 && 0.33 $\pm$ 0.01 && 0.29 $\pm$ 0.01 && 0.24 $\pm$ 0.01 &&   -   &&   -   &&   -   \\
 Only BC &&  -   && 1.22 $\pm$ 0.04 && 0.78 $\pm$ 0.02 && 0.36 $\pm$ 0.02 &&   -   &&   -   &&   -   \\

 \midrule

\spaplus (+ GT $\delta$) &&  -   && 0.22 $\pm$ 0.01 &&   -   && 0.18 $\pm$ 0.01 &&   -   &&   -   && 0.16 $\pm$ 0.01 \\
 Only CB &&  -   && 0.20 $\pm$ 0.01 &&   -   && 0.18 $\pm$ 0.01 &&   -   &&   -   && 0.17 $\pm$ 0.01 \\

		\end{tabular}
	}
		
\end{table}
}

\newcommand{\tblBenchmarkAblationMean}{
\begin{table}[tb]
	\caption{
		Benchmark results ablation study, first block of rows $KL$ result on normal benchmark, second block of rows $\hat{KL}$ on benchmark, last block of rows $KL$ results on benchmark but with ViT as backbone, all results are mean aggregations across the datasets
	}
	\label{tbl:BenchmarkAblationMean}
 \centering
	\resizebox{\linewidth}{!}{
		\begin{tabular}{l c c c c c c c c c c c c c c c c}
			\toprule
			
		method &  1 & 3 & 5 & 10 & 20 & 50 & 100 \\
  \midrule

\spaplus (ours) &0.37 $\pm$ 0.02 & 0.33 $\pm$ 0.02 & 0.32 $\pm$ 0.03 & 0.30 $\pm$ 0.02 & 0.31 $\pm$ 0.03 & 0.29 $\pm$ 0.02 & 0.29 $\pm$ 0.02 \\
 \spaplus (+ GT $\delta$) &0.36 $\pm$ 0.03 & 0.34 $\pm$ 0.02 & 0.34 $\pm$ 0.03 & 0.30 $\pm$ 0.02 & 0.30 $\pm$ 0.02 & 0.30 $\pm$ 0.02 & 0.29 $\pm$ 0.02 \\
 Only CB &0.37 $\pm$ 0.04 & 0.34 $\pm$ 0.02 & 0.32 $\pm$ 0.02 & 0.32 $\pm$ 0.03 & 0.30 $\pm$ 0.01 & 0.30 $\pm$ 0.02 & 0.30 $\pm$ 0.02 \\
 Only CB ($\mu$=0.25) &0.38 $\pm$ 0.02 & 0.35 $\pm$ 0.02 & 0.35 $\pm$ 0.03 & 0.34 $\pm$ 0.02 &   -   &   -   &   -   \\
 Only BC &  -   & 0.38 $\pm$ 0.05 & 0.35 $\pm$ 0.04 & 0.31 $\pm$ 0.03 &   -   &   -   &   -   \\

\midrule

 \spaplus (ours) &0.74 $\pm$ 0.03 & 0.32 $\pm$ 0.01 & 0.22 $\pm$ 0.01 & 0.14 $\pm$ 0.01 & 0.10 $\pm$ 0.00 & 0.07 $\pm$ 0.00 & 0.07 $\pm$ 0.00 \\
 \spaplus (+ GT $\delta$) &0.74 $\pm$ 0.03 & 0.36 $\pm$ 0.01 & 0.26 $\pm$ 0.01 & 0.14 $\pm$ 0.00 & 0.09 $\pm$ 0.00 & 0.05 $\pm$ 0.00 & 0.04 $\pm$ 0.00 \\
 Only CB &0.74 $\pm$ 0.03 & 0.32 $\pm$ 0.01 & 0.22 $\pm$ 0.01 & 0.15 $\pm$ 0.01 & 0.11 $\pm$ 0.00 & 0.09 $\pm$ 0.00 & 0.08 $\pm$ 0.00 \\
 Only CB ($\mu$=0.25) &0.54 $\pm$ 0.02 & 0.31 $\pm$ 0.01 & 0.25 $\pm$ 0.01 & 0.21 $\pm$ 0.01 &   -   &   -   &   -   \\
 Only BC &  -   & 1.15 $\pm$ 0.05 & 0.78 $\pm$ 0.03 & 0.34 $\pm$ 0.02 &   -   &   -   &   -   \\

 \midrule

\spaplus (+ GT $\delta$) &  -   & 0.23 $\pm$ 0.01 &   -   & 0.20 $\pm$ 0.01 &   -   &   -   & 0.18 $\pm$ 0.01 \\
 Only CB &  -   & 0.22 $\pm$ 0.01 &   -   & 0.20 $\pm$ 0.01 &   -   &   -   & 0.19 $\pm$ 0.01 \\

		\end{tabular}
	}
		
\end{table}
}